\documentclass{article} 
\usepackage{iclr2027_conference,times}

\usepackage{amsmath,amsfonts,bm}

\def\eqref#1{equation~\ref{#1}}

\def\1{\bm{1}}

\DeclareMathAlphabet{\mathsfit}{\encodingdefault}{\sfdefault}{m}{sl}
\SetMathAlphabet{\mathsfit}{bold}{\encodingdefault}{\sfdefault}{bx}{n}

\usepackage{hyperref}
\usepackage{url}
\usepackage{booktabs}
\usepackage{multirow}
\usepackage{tabularx}
\usepackage{array}
\usepackage{enumitem}
\usepackage{graphicx}
\usepackage{pifont}
\usepackage{array}
\usepackage[table]{xcolor}
\definecolor{deltabg}{RGB}{235,244,252}
\definecolor{nacolor}{RGB}{242,242,242}

\newcommand{\cmark}{\ding{51}}
\newcommand{\xmark}{\ding{55}}

\definecolor{Ubg}{RGB}{232,241,250}
\definecolor{Ebg}{RGB}{235,246,239}
\definecolor{Rbg}{RGB}{249,221,213}  
\definecolor{Cbg}{RGB}{233,223,245}  

\usepackage[most]{tcolorbox}
\usepackage{xcolor}

\definecolor{perturbred}{RGB}{170,38,38}     
\definecolor{affectbrown}{RGB}{140,92,32}    
\definecolor{dialogueframe}{RGB}{120,120,120}
\definecolor{dialogueback}{RGB}{242,242,242}

\newcommand{\TaskTurn}[2]{%
  \par\vspace{.28ex}\noindent
  \textbf{Task #1.}~#2\par
}

\newcommand{\AgentResponse}[2]{%
  \par\vspace{.20ex}\noindent
  \textbf{Agent #1.}~#2\par
}

\newcommand{\PerturbedTaskTurn}[2]{%
  \par\vspace{.28ex}\noindent
  \textbf{\textcolor{perturbred}{Task #1 [Perturbed].}}~#2\par
}

\newcommand{\AffectedAgentResponse}[2]{%
  \par\vspace{.20ex}\noindent
  \textbf{\textcolor{affectbrown}{Agent #1 [Affected].}}~#2\par
}

\newcommand{\FirstUseAgentResponse}[2]{%
  \par\vspace{.20ex}\noindent
  \textbf{\textcolor{affectbrown}{Agent #1 [First use].}}~#2\par
}

\newcommand{\Perturb}[1]{%
  \textcolor{perturbred}{#1}%
}

\newcommand{\Affected}[1]{%
  \textcolor{affectbrown}{#1}%
}

\newtcolorbox{dialoguebox}[1]{
  enhanced,
  sharp corners,
  colback=white,
  colframe=dialogueframe,
  boxrule=.45pt,
  left=1.1mm,
  right=1.1mm,
  top=.55mm,
  bottom=.55mm,
  title={#1},
  fonttitle=\bfseries\fontsize{7.1}{7.5}\selectfont,
  coltitle=black,
  colbacktitle=dialogueback,
  fontupper=\fontsize{6.15}{6.55}\selectfont,
  before upper={
    \raggedright
    \setlength{\parindent}{0pt}
    \setlength{\parskip}{0pt}
    \setlength{\abovedisplayskip}{1pt}
    \setlength{\belowdisplayskip}{1pt}
    \setlength{\abovedisplayshortskip}{.5pt}
    \setlength{\belowdisplayshortskip}{.5pt}
  },
  before skip=0pt,
  after skip=0pt
}

\definecolor{evidencegreen}{RGB}{235,246,240}

\newtcolorbox{correctevidence}[1]{
  enhanced,
  sharp corners,
  colback=evidencegreen,
  colframe=green!35!black,
  boxrule=.35pt,
  left=1mm,
  right=1mm,
  top=.55mm,
  bottom=.55mm,
  title={#1},
  fonttitle=\bfseries\fontsize{6.25}{6.8}\selectfont,
  fontupper=\fontsize{5.8}{6.35}\selectfont,
  before skip=0pt,
  after skip=1.1mm
}
\newtcolorbox{injectedevidence}[1]{
  enhanced,
  sharp corners,
  colback=evidencered,
  colframe=perturbred!65,
  colbacktitle=perturbred,
  coltitle=white,
  boxrule=.35pt,
  left=1mm,
  right=1mm,
  top=.55mm,
  bottom=.55mm,
  title={#1},
  fonttitle=\bfseries\fontsize{6.25}{6.8}\selectfont,
  fontupper=\color{black}\fontsize{5.8}{6.35}\selectfont,
  before skip=0pt,
  after skip=1.1mm
}
\definecolor{evidencered}{RGB}{255,239,240}
\usepackage[most]{tcolorbox}
\usepackage[dvipsnames]{xcolor}

\title{The Marathon of Scientific Reasoning: \\ Robustness of Scientific Agents to \\ Perturbations in Multi-Turn Interactions}

\iclrfinalcopy

\author{
\textbf{
Xiaoting Lyu$^{1}$,
Xinbo Ma$^{1}$,
Yufei Han$^{2}$,
Hangwei Qian$^{3}$
}\\
\textbf{
Ziyang Lin$^{1}$,
Bin Wang$^{1}$,
Bin Wang$^{4}$,
Wei Wang$^{1}$
}\\[4pt]
$^{1}$Xi'an Jiaotong University, China\\
$^{2}$Inria, France\\
$^{3}$Agency for Science, Technology and Research (A*STAR), Singapore\\
$^{4}$Zhejiang Key Laboratory of Artificial Intelligence of Things (AIoT) Network and Data Security, China
}

\begin{document}

\maketitle

\lhead{}

\begin{abstract}
Large language model (LLM)-based scientific agents are increasingly used for scientific problem solving, yet their robustness to imperfections arising during multi-turn interactions remains poorly understood. We introduce \textsc{SciARP} (\textbf{Sci}entific \textbf{A}gent \textbf{R}obustness to \textbf{P}erturbations), a benchmark for evaluating scientific agents under scientifically plausible perturbations throughout multi-turn problem solving. \textsc{SciARP} transforms 620 scientific problems into interdependent tasks of 3--13 turns and defines 13 perturbation types spanning problem understanding, evidence processing, reasoning, and conclusion formation. Clean and perturbed versions of each task are independently executed under matched settings, producing paired live trajectories for evaluating both task success and process reliability. Experiments across eight LLMs from four model families reveal three key robustness characteristics. First, different classes of scientific perturbations exhibit distinct robustness profiles and can decouple task progression from scientific reliability: agents may continue advancing through the task even after their information or reasoning has become unreliable. Second, stronger clean-task performance does not necessarily translate into stronger robustness, as models with higher clean-task accuracy can exhibit larger degradation under perturbation. Third, perturbation effects exhibit strong temporal dynamics: they may remain latent for multiple turns before emerging and subsequently propagate through downstream dependencies. Together, these findings show that current scientific agents remain insufficiently robust to scientifically plausible perturbations, with failures often remaining undetected, propagating, and resisting recovery.
\end{abstract}

\section{Introduction}
\label{sec1-intro}

Artificial intelligence for science (AI4S) is moving beyond isolated scientific question answering toward LLM-based agents that retrieve literature, analyze data, use tools, run experiments, and carry out multi-step scientific workflows~\citep{lu2024aiscientist,chen2025scienceagentbench,mitchener2025bixbench,majumder2025discoverybench,chen2025mlrbench,bragg2026astabench,koblischke2025gravitybench,zheng2026newtonbench,shen2026sciagentgym}. Unlike single-turn question answering, scientific problem solving often unfolds through a sequence of dependent steps: \emph{experimental observations}, \emph{evidence}, \emph{calculations}, and \emph{intermediate conclusions} produced earlier become inputs to later reasoning. As a result, scientific reliability depends not only on whether an agent reaches the correct answer, but also on whether the information and reasoning accumulated throughout the trajectory remain valid. In interactions, however, imperfections may arise at points in this process. Problem descriptions may contain ambiguous terminology or incomplete conditions, evidence may be unreliable or conflicting, intermediate reasoning may become inconsistent or misleading, and conclusion formation may be influenced by unsupported cues. Such perturbations may be corrected immediately, remain latent until the affected information is reused, or propagate through downstream reasoning. This raises a central yet underexplored robustness question: \emph{Can scientific agents maintain reliable problem solving when scientifically plausible perturbations arise during multi-turn interactions?}

Existing benchmarks capture only parts of this problem. Scientific-agent benchmarks evaluate capabilities in scientific analysis, discovery, research automation, and multi-step tool use~\citep{chen2025scienceagentbench,mitchener2025bixbench,majumder2025discoverybench,chen2025mlrbench,bragg2026astabench,koblischke2025gravitybench,zheng2026newtonbench,shen2026sciagentgym}. Reasoning-robustness benchmarks such as GSM-Symbolic and MATH-Perturb study sensitivity to controlled variations in problem statements~\citep{mirzadeh2025gsmsymbolic,huang2025mathperturb}, but largely treat perturbations as modifications to a single input. Agent-safety benchmarks such as AgentDojo and AgentHarm instead focus on explicitly adversarial settings, including prompt injection, jailbreaks, and malicious tasks~\citep{debenedetti2024agentdojo,andriushchenko2025agentharm}. What remains underexplored is the robustness of scientific agents to plausible, non-adversarial perturbations that arise during an evolving multi-turn process, where later decisions depend on earlier information and perturbation effects may emerge only after several dependent steps. Evaluating such failures therefore requires moving beyond final-answer accuracy to trajectory-level evaluation of scientific reliability.

To address this gap, we introduce \textsc{SciARP} (\textbf{Sci}entific \textbf{A}gent \textbf{R}obustness to \textbf{P}erturbations), a benchmark for evaluating robustness throughout multi-turn scientific problem solving. \textsc{SciARP} converts 620 verified problems from GPQA~\citep{rein2024gpqa}, SciBench~\citep{wang2024scibench}, OlympiadBench~\citep{he2024olympiadbench}, and BixBench v1.5~\citep{mitchener2025bixbench} into interdependent 3--13-turn tasks covering diverse domains of biology, mathematics, chemistry and physics. Each task preserves its original scientific objective and reference solution, which serves as the ground truth throughout evaluation. We define 13 scientifically grounded perturbation types across four stages of problem solving: \emph{problem understanding}, \emph{evidence processing}, \emph{reasoning}, and \emph{conclusion formation}. We inject them into the multi-turn interaction while preserving the underlying scientific objective. We run clean and perturbed versions of each task independently under matched settings, yielding paired live trajectories rather than perturbed pre-collected traces. This paired live-execution design enables \textsc{SciARP} to assess both end-to-end task success and process reliability through \emph{Task Accuracy}, \emph{Information Correctness}, \emph{Task Progression}, and \emph{Reasoning Validity}, and to characterize when perturbation-induced failures emerge, how they propagate across dependent turns, and whether subsequent reasoning recovers.

We evaluate eight LLMs from four model families on \textsc{SciARP} under a unified agentic pipeline. The results reveal three characteristics of scientific-agent robustness. \textbf{First}, different stage-associated perturbation families exhibit distinct robustness profiles. Within our taxonomy, reasoning perturbations produce the largest average degradation in task success, whereas conclusion perturbations most strongly affect information correctness and reasoning validity. More broadly, understanding and evidence perturbations tend to manifest first as information errors, while reasoning and conclusion perturbations more often disrupt reasoning validity or task progression. These failures are often \emph{silently persistent}: agents keep advancing even with the misled information or reasoning paths, without identifying or correcting them. \textbf{Second}, stronger perturbation-free task-solving performance does not translate into stronger robustness. Agents with higher clean accuracy can suffer larger perturbation-induced degradation, revealing a capability--robustness gap. \textbf{Third}, failures show strong temporal dynamics. Perturbation effects may remain latent for several turns before emerging and subsequently propagate through downstream dependencies. Once a failure propagates beyond its onset turn, it is substantially more likely to persist than to recover. Together, these findings show that strong scientific capability alone does not guarantee robust problem solving: perturbation-induced failures can remain hidden, propagate across dependent steps, and persist while the task appears to progress.

\section{Related Work}
\label{sec2-related}
\textbf{LLM Agents for Scientific Reasoning and Discovery.}
Recent benchmarks have expanded scientific-agent evaluation from isolated question answering to multi-step scientific workflows. \textsc{ScienceAgentBench} and \textsc{DiscoveryBench} evaluate data-driven scientific analysis and discovery~\citep{chen2025scienceagentbench,majumder2025discoverybench}, while \textsc{BixBench} targets long-horizon computational analysis over biological data~\citep{mitchener2025bixbench}. \textsc{Gravity-Bench} and \textsc{NewtonBench} study interactive scientific discovery through observation, experimentation, and law induction~\citep{koblischke2025gravitybench,zheng2026newtonbench}. \textsc{MLR-Bench} and \textsc{AstaBench} extend evaluation to open-ended and multi-stage research workflows~\citep{chen2025mlrbench,bragg2026astabench}, while \textsc{SciAgentGym} focuses on multi-step scientific tool use~\citep{shen2026sciagentgym}. These benchmarks primarily assess scientific capability and generalization under their task settings, with \textsc{NewtonBench} examining robustness to observational noise. \textsc{SciARP} instead focuses on robustness to scientifically plausible perturbations arising throughout dependent multi-turn problem solving, and evaluates not only task outcomes but also how reliability evolves across the interaction.
\begin{table}[t]
\centering
\caption{Comparison with representative scientific-agent benchmarks.}
\label{tab:related}
\small
\resizebox{\linewidth}{!}{
\begin{tabular}{lccccc}
\toprule
\textbf{Benchmark} &
\textbf{Scientific Setting} &
\textbf{Primary Focus} &
\textbf{Controlled Perturbation} &
\textbf{Turn-level Reliability} &
\textbf{Propagation / Recovery} \\
\midrule
ScienceAgentBench & Data-driven analysis       & Capability                  & --                   & \xmark & \xmark \\
DiscoveryBench    & Data-driven discovery      & Capability                  & --                   & \xmark & \xmark \\
BixBench          & Computational biology      & Capability                  & --                   & \xmark & \xmark \\
Gravity-Bench-v1     & Interactive discovery      & Capability / Generalization & --                   & \xmark & \xmark \\
NewtonBench       & Interactive law discovery  & Generalization / Robustness & Observation noise    & \xmark & \xmark \\
MLR-Bench         & Open-ended ML research     & Capability                  & --                   & \xmark & \xmark \\
AstaBench         & Scientific research suite  & Capability                  & --                   & \xmark & \xmark \\
SciAgentGym       & Scientific tool use        & Capability                  & --                   & \xmark & \xmark \\
\midrule
\textbf{SciARP}   & \textbf{Multi-turn scientific reasoning}
                  & \textbf{Robustness}
                  & \textbf{Scientific perturbations}
                  & \cmark & \cmark \\
\bottomrule
\end{tabular}}
\end{table}

\textbf{Robustness in LLM Reasoning and Agents.}
A complementary line of work studies robustness under controlled variations and interaction-induced failures. \textsc{GSM-Symbolic} and \textsc{MATH-Perturb} reveal the sensitivity of mathematical reasoning to problem variations~\citep{mirzadeh2025gsmsymbolic,huang2025mathperturb}, while \textsc{NoisyPromptBench} evaluates robustness to synthetic prompt noise~\citep{yang2026coipo}. Recent studies further examine degradation in multi-turn interactions~\citep{laban2026llms}, knowledge conflicts between parametric memory and external evidence~\citep{xie2024adaptive}, and broader agent reliability dimensions including consistency, robustness, predictability, and safety~\citep{rabanser2026reliability}. Agent-safety benchmarks such as \textsc{AgentDojo} and \textsc{AgentHarm} instead focus on adversarial settings, including prompt injection, jailbreaks, and malicious tasks~\citep{debenedetti2024agentdojo,andriushchenko2025agentharm}. In contrast, \textsc{SciARP} studies scientifically grounded perturbations arising throughout evolving multi-turn scientific processes. Through paired live execution and turn-level evaluation, it characterizes performance degradation together with failure onset, propagation, and recovery across dependent turns.

\section{Problem Formulation}
\label{sec3-problem}
\subsection{Multi-Turn Scientific Problem Solving}
\label{sec:scientific_reasoning}
We consider a scientific task $\mathcal{T}=(x,y^*)$, where $x$ denotes the original scientific problem and its task context, and $y^*$ denotes the verified reference solution that serves as the ground truth throughout evaluation. We represent the task as an ordered multi-turn interaction $\mathcal{I}_{\mathcal{T}}=\left((u_t,s_t)\right)_{t=1}^{T}$, where $T$ is the number of interaction turns. Each turn $t$ represents a step in the scientific problem-solving process, with $u_t$ denoting the task input presented to the agent, including its instruction and associated scientific information, and $s_t\in\mathcal{S}$ denoting its primary problem-solving stage. We define $\mathcal{S}=\{U,E,R,C\}$, corresponding to \emph{problem understanding}, \emph{evidence processing}, \emph{reasoning}, and \emph{conclusion formation}. Specifically, $U$ concerns interpreting the scientific objective, terminology, notation, variables, units, and task constraints; $E$ concerns observations, facts, data, conditions, and supporting evidence; $R$ covers computation, deduction, hypothesis evaluation, and intermediate inference; and $C$ integrates accumulated information and reasoning into a conclusion or answer. These labels characterize the primary role of each turn rather than a fixed temporal sequence, and the same stage may recur as the interaction evolves. The interaction is inherently interdependent because later turns may rely on assumptions, evidence, calculations, or intermediate deductions established earlier. Successful task completion therefore requires the agent to maintain, integrate, and correctly reuse scientific information across turns.

Given $\mathcal{I}_{\mathcal{T}}$, a scientific agent $A$ is an LLM-based system that solves scientific tasks through sequential multi-turn interaction, maintaining the accumulated interaction context and, when applicable, invoking external tools. Executing $A$ produces a trajectory $\tau=A(\mathcal{I}_{\mathcal{T}}) =(r_1,r_2,\ldots,r_T,y)$, where $r_t$ denotes the observable execution at turn $t$, including the agent response and any associated tool calls or outputs, and $y$ denotes the final output. Each turn is generated from the accumulated interaction context, so its resulting response and tool outputs may influence subsequent decisions. Importantly, $\tau$ is obtained through live execution rather than synthesized from a predefined reasoning trace.

\subsection{Scientific Perturbations}
\label{sec:scientific_perturbations}

During multi-turn scientific problem solving, deviations may arise in problem descriptions, evidence, intermediate reasoning, or conclusions. We define a \emph{scientific perturbation} as a scientifically plausible deviation introduced into task-relevant information during the problem-solving process. Unlike adversarial attacks optimized to induce failure in a specific target agent, these perturbations follow predefined, task-conditioned mechanisms and are not optimized against the evaluated agent.

Given a clean multi-turn task $\mathcal{I}_{\mathcal{T}}$, we apply a perturbation operator $\mathcal{P}$ to obtain its perturbed counterpart $\widetilde{\mathcal{I}}_{\mathcal{T}}=\mathcal{P}\left(\mathcal{I}_{\mathcal{T}}\right)$. The perturbation modifies task-relevant information presented during the interaction while preserving the original scientific objective. \textsc{SciARP} considers multiple scientifically grounded perturbation types, as detailed in Section~\ref{sec4-seciperturb}. We independently execute the same scientific agent $A$ under matched settings on the clean and perturbed tasks, producing paired trajectories through live multi-turn execution rather than by modifying or replaying a pre-collected trajectory. Once a perturbation is introduced, subsequent responses evolve from the resulting interaction context, allowing the perturbed execution to develop a different trajectory and final output. Comparing the paired trajectories enables us to characterize how perturbation effects emerge, propagate across dependent turns, and are corrected through subsequent reasoning, as well as how they ultimately affect task success.

\section{SciARP  Benchmark}
\label{sec4-seciperturb}
We design \textsc{SciARP} around three principles: perturbations target scientifically meaningful aspects of problem solving, clean and perturbed tasks are independently executed under matched settings, and robustness is assessed at both the task and process levels.

\subsection{Scientific Perturbation Taxonomy}
\label{sec:perturbation_taxonomy}
We organize perturbations according to the primary problem-solving stage they target. Following the four stages defined in Section~\ref{sec:scientific_reasoning}, \textsc{SciARP} defines 13 scientifically plausible perturbation types across \emph{problem understanding} ($U$), \emph{evidence processing} ($E$), \emph{reasoning} ($R$), and \emph{conclusion formation} ($C$). These categories characterize the primary function affected by a perturbation rather than a fixed temporal order. Together, they capture deviations in task representation and specification, evidence quality and consistency, intermediate reasoning, and conclusion formation. Detailed definitions and examples are provided in Appendix~\ref{app:perturbation_taxonomy}.


\noindent\textbf{Problem Understanding Perturbations ($U$).}
Scientific tasks may vary in terminology, notation, units, and the completeness of their specifications, affecting how the underlying problem is interpreted. We consider \textit{U1, Scientific Terminology Variation Perturbation}, which replaces scientific terms with equivalent or alternative expressions; \textit{U2, Dimension and Unit Perturbation}, which changes units, scales, or dimensional representations; \textit{U3, Variable and Symbol Perturbation}, which alters or ambiguously reuses scientific variables or notation; and \textit{U4, Condition Omission Perturbation}, which removes task-relevant conditions needed for valid reasoning. These perturbations evaluate whether the agent preserves the intended task semantics under representational variation and appropriately handles incomplete task specifications.

\noindent\textbf{Evidence Processing Perturbations ($E$).}
Scientific reasoning depends on the validity and consistency of premises, observations, data, citations, and evidence from multiple sources. We consider \textit{E1, Invalid Premise Perturbation}, which introduces an incorrect scientific premise; \textit{E2, Citation and Attribution Perturbation}, which introduces invalid or misattributed references; \textit{E3, Data and Statistical Evidence Perturbation}, which alters empirical data or statistical evidence; and \textit{E4, Multi-Source Evidence Conflict Perturbation}, which introduces inconsistencies across multiple evidence sources. These perturbations evaluate whether the agent can assess the validity and consistency of scientific evidence before incorporating it into subsequent reasoning.


\noindent\textbf{Reasoning Perturbations ($R$).}
Scientific reasoning may fail through inconsistent intermediate claims, unsupported causal interpretations, or misleading solution strategies. We consider \textit{R1, Cross-Step Inconsistency Perturbation}, which introduces an intermediate statement inconsistent with previously established reasoning or results; \textit{R2, Correlation--Causation Confusion Perturbation}, which introduces a causality interpretation unsupported by the observed correlation; and \textit{R3, Reasoning-Path Misdirection Perturbation}, which suggests an inappropriate method, approximation, analogy, or reasoning direction. These perturbations evaluate whether the agent can detect invalid intermediate reasoning and maintain scientific and logical consistency across dependent turns.


\noindent\textbf{Conclusion Formation Perturbations ($C$).}
Even when evidence and reasoning are largely valid, conclusion formation may be distorted by unjustified certainty or unsupported external cues. We consider \textit{C1, Uncertainty Suppression Perturbation}, which encourages an overly definitive conclusion despite uncertainty, and \textit{C2, Authority and Stance Pressure Perturbation}, which introduces unsupported authority or stance signals favoring a particular conclusion. These perturbations evaluate whether the agent maintains appropriately calibrated and evidence-grounded conclusions.


\subsection{Benchmark Construction}
\label{sec:benchmark_construction}

\begin{figure*}[t]
    \centering
    \includegraphics[width=\textwidth]{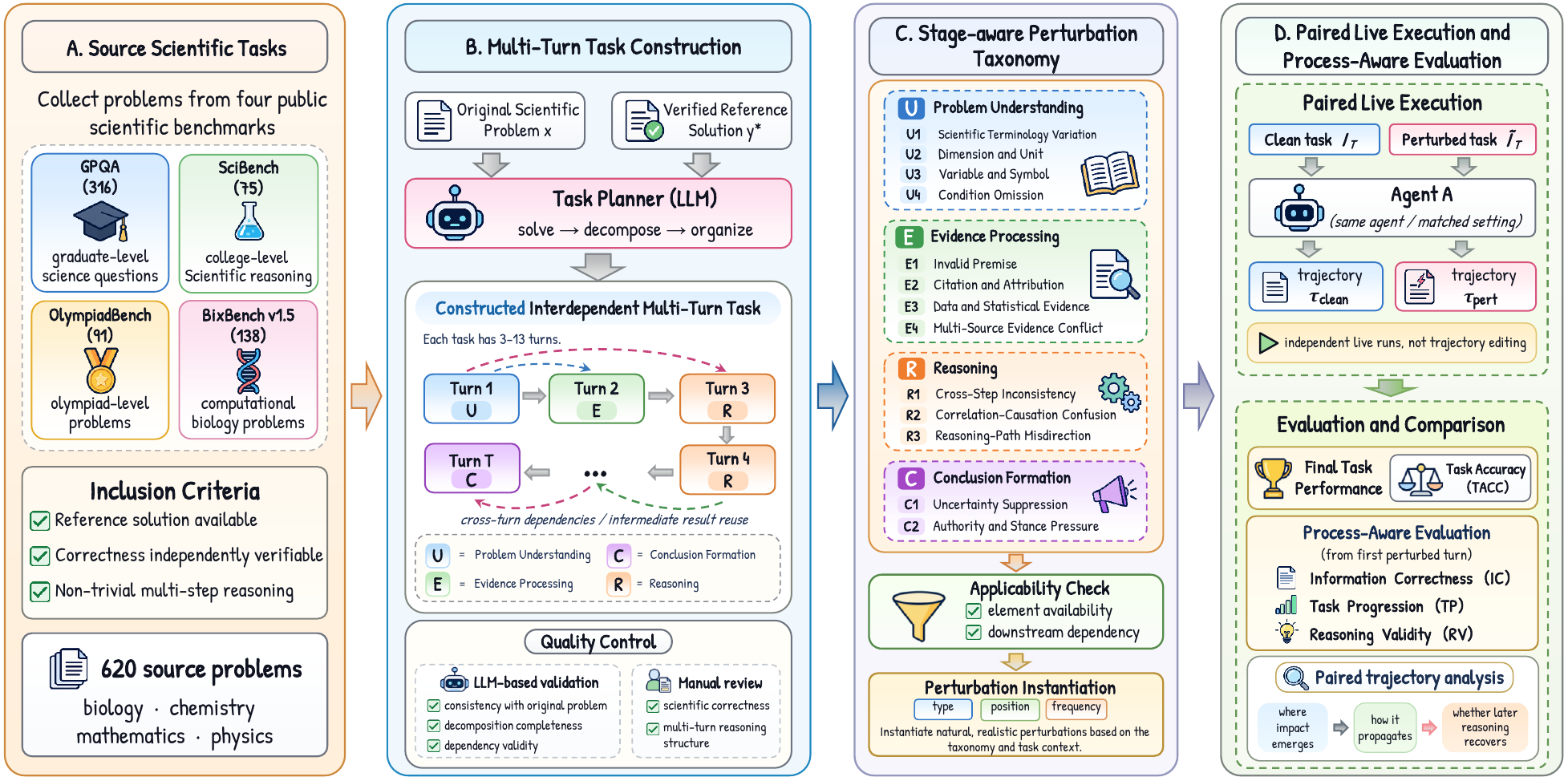}
    \caption{Overview of the \textsc{SciARP} benchmark construction and evaluation framework.}
    \label{fig:benchmark_framework}
\end{figure*}

We construct \textsc{SciARP} from four established scientific benchmarks: GPQA~\citep{rein2024gpqa}, SciBench~\citep{wang2024scibench}, OlympiadBench~\citep{he2024olympiadbench}, and BixBench v1.5~\citep{mitchener2025bixbench}. We retain 620 problems spanning biology, chemistry, mathematics, and physics that have a reference solution, can be independently verified, and require non-trivial multi-step reasoning. Each problem and its reference solution are transformed into an interdependent multi-turn task containing 3--13 turns of scientific reasoning and, when applicable, tool use. We then instantiate scientifically plausible perturbations at valid interaction points while preserving the original scientific objective. Clean and perturbed tasks are independently executed on the same evaluated agent, producing paired live trajectories. Figure~\ref{fig:benchmark_framework} summarizes the construction and evaluation framework, with additional construction details provided in Appendix~\ref{app:multiturn_example}.

\textbf{Multi-Turn Task Construction.}
For each source problem $\mathcal{T}=(x,y^*)$, an LLM-based \emph{Task Planner} first constructs a complete solution process consistent with the reference solution and then decomposes it into logically connected subtasks, each corresponding to one interaction turn. The decomposition follows the scientific reasoning structure rather than mechanically partitioning the original problem: closely related steps are grouped within a turn, while steps that depend on earlier outputs are placed in subsequent turns. Each turn is assigned its primary problem-solving stage in $\{U,E,R,C\}$. The resulting tasks contain 3--13 turns with explicit cross-turn dependencies, requiring intermediate information to be retained and reused across the interaction. Each constructed task undergoes dual validation by an LLM and human reviewers using the same criteria, including consistency with the original problem and reference solution, scientific correctness, completeness of the reasoning process, and validity of cross-turn dependencies. Only tasks that pass both validation stages are retained. A complete construction example is provided in Appendix~\ref{app:multiturn_example}.

\textbf{Perturbation Instantiation.}
For each multi-turn task and perturbation type, an LLM-based \emph{Perturbation Generator} first examines the scientific content, problem-solving stages, and cross-turn dependencies to identify valid perturbation points. A point is considered valid when the scientific element required by the perturbation is present and, for perturbations introduced before conclusion formation, the affected information is used in subsequent reasoning. The generator then modifies the designated information at a valid point according to the corresponding perturbation definition, while preserving the original scientific objective. When multiple valid points are available, the injection position and frequency are varied where applicable; perturbations tied to specific stages are instantiated only at their corresponding valid points. Each generated instance is independently validated by an LLM and human reviewers using the same criteria, including perturbation correctness, scientific plausibility, targeted modification, and cross-turn consistency. Only instances that pass both validation stages are retained. Detailed instantiation rules and representative clean--perturbed examples are provided in Appendix~\ref{app:perturbation_instantiation}.

\textbf{Paired Agent Execution.}
For each clean--perturbed pair $(\mathcal{I}_{\mathcal{T}},\widetilde{\mathcal{I}}_{\mathcal{T}})$, both instances are independently executed on the same evaluated agent under matched settings. The perturbed trajectory is generated through live multi-turn interactions rather than by editing or replaying the clean trajectory. Once a perturbation is introduced, subsequent responses evolve from the resulting interaction context, allowing its effects to emerge immediately or after a delay, propagate across dependent turns, or be corrected through subsequent reasoning. The resulting paired trajectories support both stage-wise robustness comparison and analysis of failure propagation across the interaction.

\subsection{Robustness Evaluation}
\label{sec:robustness_evaluation}
We evaluate scientific-agent robustness at two complementary levels. Task Accuracy (TACC) measures end-to-end task success, while Information Correctness (IC), Task Progression (TP), and Reasoning Validity (RV) characterize the reliability of the problem-solving process. 

\textbf{Task Accuracy (TACC).}
For each evaluated task $i$, TACC measures whether a task is successfully completed. We consider a task successful only when its final answer matches the original scientific ground truth $y_i^*$ and the corresponding solution process is scientifically valid:
\begin{equation}
\mathrm{TACC}
=
\frac{1}{N}
\sum_{i=1}^{N}
\mathbb{I}\!\left[
\text{Answer}_i\ \text{matches}\ y_i^*
\land
\text{Process}_i\ \text{is valid}
\right],
\end{equation}
where $N$ is the number of evaluated tasks and $\mathbb{I}[\cdot]$ denotes the indicator function.

For the three process-level metrics below, evaluation begins at the first perturbed turn $t_i^{\mathrm{pert}}$ and continues to the final turn $T_i$. When multiple perturbations are introduced, $t_i^{\mathrm{pert}}$ denotes the earliest perturbed turn. The paired clean trajectory is evaluated over the same turn range. Each turn receives a binary judgment, which is first averaged within each task and then across tasks so that trajectories of different lengths contribute equally.

\textbf{Information Correctness (IC).}
IC measures whether task-relevant scientific information produced or used by the agent throughout the problem-solving process is correct, including facts, values, conditions, sources, and intermediate results:
\begin{equation}
\mathrm{IC}
=
\frac{1}{N}
\sum_{i=1}^{N}
\left(
\frac{1}{T_i-t_i^{\mathrm{pert}}+1}
\sum_{t=t_i^{\mathrm{pert}}}^{T_i}
\mathbb{I}
[
\mathrm{Information}_{i,t}\text{ is correct}
]
\right).
\end{equation}
where $\text{information}_{i,t}$ denotes the task-relevant scientific content used by the agent at turn $t$ of task $i$. 

\textbf{Task Progression (TP).} 
TP measures whether the agent advances the task throughout the problem-solving process by completing the current subtask and producing the information, decision, or intermediate result needed for subsequent steps:
\begin{equation}
\mathrm{TP}
=
\frac{1}{N}
\sum_{i=1}^{N}
\left(
\frac{1}{T_i-t_i^{\mathrm{pert}}+1}
\sum_{t=t_i^{\mathrm{pert}}}^{T_i}
\mathbb{I}
[
\mathrm{Turn}_{i,t}\text{ advances the task}
]
\right).
\end{equation}

\textbf{Reasoning Validity (RV).}
RV measures whether the intermediate result at each turn is correct and supported by scientifically and logically valid reasoning given the preceding interaction context:
\begin{equation}
\mathrm{RV}
=
\frac{1}{N}
\sum_{i=1}^{N}
\left(
\frac{1}{T_i-t_i^{\mathrm{pert}}+1}
\sum_{t=t_i^{\mathrm{pert}}}^{T_i}
\mathbb{I}
[
\mathrm{Result}_{i,t}\text{ is correct}
\land
\mathrm{Reasoning}_{i,t}\text{ is valid}
]
\right).
\end{equation}

Higher TACC indicates stronger end-to-end task success, while higher IC, TP, and RV indicate greater reliability of scientific information, task progression, and reasoning, respectively.

\textbf{Evaluation Quality Control.}
We adopt a dual-evaluation protocol combining a fixed LLM evaluator with human assessment. Both follow the same predefined, metric-specific rubric. For IC, TP, and RV, each turn is assessed using the original scientific problem, reference solution, interaction history up to the current turn, and current task request. For TACC, evaluators assess both final-answer correctness with respect to the original scientific ground truth and the scientific validity of the overall solution process. The LLM evaluator is fixed as Claude Opus 5 across all experiments. Evaluators are not informed of the evaluated model identity, clean or perturbed condition, or perturbation category. Human assessment is conducted independently by two annotators with at least 5 years of scientific research experience, whose judgments are cross-checked for consistency. The resulting human judgment is then compared with the LLM evaluation, and a criterion is counted as satisfied only when both agree; disagreements are conservatively treated as incorrect.

\section{Experiments and Analysis}
\label{sec5-experiment}
\subsection{Experimental Setup}
\label{sec:experimental_setup}

We evaluate eight LLMs from four model families: \emph{Gemini 3.1 Pro}~\citep{google2026gemini31pro} and \emph{Gemini 3.7 Flash}~\citep{googledeepmind2026gemini37flash}, \emph{DeepSeek-V4-Pro} and \emph{DeepSeek-V4-Flash}~\citep{deepseek2026v4}, \emph{GPT-5.6}~\citep{openai2026gpt56} and \emph{GPT-5.5}~\citep{openai2026gpt55}, and \emph{Claude Opus 5}~\citep{anthropic2026opus5} and \emph{Claude Sonnet 5}~\citep{anthropic2026sonnet5}. All models are evaluated using the same scientific-agent pipeline and multi-turn interaction protocol. Each model sequentially executes the task and produces a live trajectory for evaluation. Within each clean--perturbed pair, the system prompt, agent configuration, tool settings, and model-specific inference settings are held fixed. We use temperature $0$ when supported and keep reasoning or thinking configurations fixed for models that expose such controls. Each experimental condition is independently executed three times, with each repetition constituting a fresh live run, and we report the mean performance across the three runs. For the overall robustness evaluation, each task contains a single perturbation ($n=1$) introduced at an applicable turn. We separately examine perturbation position by varying the injection point across valid turns and perturbation frequency by increasing the number of perturbations when repeated instantiation is semantically applicable. All other experimental conditions remain fixed. Detailed configurations are provided in Appendix~\ref{app:experimental_details}.

\begin{table}[t]
\centering
\caption{Robustness of the eight evaluated models across four scientific
problem-solving stages (\%). For each model and stage, results are macro-averaged over the perturbation types belonging to that stage.
$\Delta$ denotes Clean $-$ Perturbed; larger values indicate greater degradation.}
\label{tab:overall_robustness}
\renewcommand{\arraystretch}{1.08}
\setlength{\tabcolsep}{2.2pt}
\resizebox{\linewidth}{!}{
\begin{tabular}{
l|
cccc|
cccc|
cccc|
cccc
}
\toprule
\multicolumn{1}{c|}{\multirow{2}{*}[-0.3ex]{\textbf{Model}}}
& \multicolumn{4}{>{\columncolor{Ubg}}c|}{\textbf{Problem Understanding (U)}}
& \multicolumn{4}{>{\columncolor{Ebg}}c|}{\textbf{Evidence Processing (E)}}
& \multicolumn{4}{>{\columncolor{Rbg}}c|}{\textbf{Reasoning (R)}}
& \multicolumn{4}{>{\columncolor{Cbg}}c}{\textbf{Conclusion Formation (C)}} \\
\cmidrule(lr){2-5}
\cmidrule(lr){6-9}
\cmidrule(lr){10-13}
\cmidrule(lr){14-17}

& $\Delta$\textbf{TACC} $\downarrow$  & $\Delta$\textbf{IC}  $\downarrow$& $\Delta$\textbf{TP}  $\downarrow$& $\Delta$\textbf{RV} $\downarrow$
& $\Delta$\textbf{TACC}  $\downarrow$& $\Delta$\textbf{IC}  $\downarrow$& $\Delta$\textbf{TP}  $\downarrow$& $\Delta$\textbf{RV} $\downarrow$
& $\Delta$\textbf{TACC}  $\downarrow$& $\Delta$\textbf{IC}  $\downarrow$& $\Delta$\textbf{TP}  $\downarrow$& $\Delta$\textbf{RV} $\downarrow$
& $\Delta$\textbf{TACC} $\downarrow$ & $\Delta$\textbf{IC}  $\downarrow$& $\Delta$\textbf{TP}  $\downarrow$& $\Delta$\textbf{RV} $\downarrow$ \\
\midrule

Gemini 3.1 Pro
& 10.25 & 3.73 & 1.27 & 2.16
& 25.46 & 3.60 & 0.54 & 1.46
& 51.89 & 18.44 & 20.92 & 41.39
& 23.08 & 13.16 & 11.82 & 18.26 \\

Gemini 3.7 Flash
& 9.05 & 3.19 & 2.70 & 6.45
& 27.62 & 4.88 & 1.25 & 7.38
& 59.50 & 20.33 & 24.98 & 41.33
& 41.61 & 35.59 & 25.44 & 43.79 \\

\midrule
DeepSeek V4 Pro
& 9.03 & 2.57 & 1.87 & 4.30
& 20.77 & 3.51 & 1.98 & 6.02
& 50.74 & 14.08 & 19.94 & 28.54
& 31.13 & 22.87 & 16.92 & 29.71 \\

DeepSeek V4 Flash
& 10.66 & 2.96 & 3.17 & 6.56
& 18.70 & 3.81 & 1.26 & 5.29
& 50.04 & 13.82 & 23.90 & 28.14
& 36.38 & 24.25 & 17.55 & 37.52 \\

\midrule
GPT-5.6
& 11.87 & 2.62 & 2.68 & 4.93
& 24.68 & 4.47 & 2.11 & 7.62
& 53.22 & 13.99 & 23.93 & 24.28
& 28.03 & 18.54 & 11.87 & 26.56 \\

GPT-5.5
& 11.21 & 4.33 & 2.23 & 7.34
& 20.06 & 4.48 & 0.74 & 9.04
& 49.68 & 17.40 & 22.68 & 30.87
& 30.46 & 22.10 & 15.92 & 30.32 \\

\midrule
Claude Opus 5
& 4.34 & 0.91 & 1.30 & 2.24
& 15.39 & 2.37 & 0.73 & 6.30
& 43.34 & 7.93 & 21.39 & 24.71
& 29.43 & 18.74 & 21.20 & 33.26 \\

Claude Sonnet 5
& 9.07 & 2.68 & 2.11 & 5.76
& 21.13 & 3.47 & 1.26 & 6.30
& 47.11 & 10.32 & 24.76 & 17.04
& 32.01 & 23.59 & 18.91 & 34.11 \\

\midrule
\textbf{Average}
& 9.43 & 2.87 & 2.16 & 4.97
& 21.73 & 3.82 & 1.23 & 6.18
& 50.69 & 14.54 & 22.81 & 29.54
& 31.51 & 22.35 & 17.45 & 31.69 \\

\bottomrule
\end{tabular}
}
\end{table}

\subsection{Overall Robustness under Scientific Perturbations}
\label{sec:overall_robustness}

\begin{figure*}[htb]
    \centering
    \includegraphics[width=\linewidth]{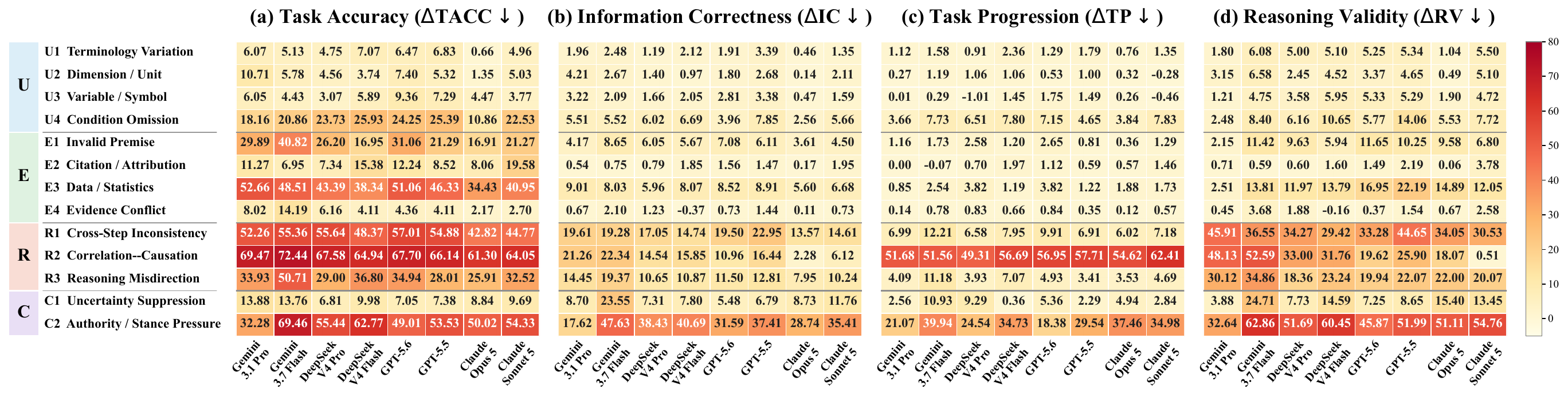}
    \caption{Robustness landscape across models and scientific perturbations (\%). Each cell reports the degradation $\Delta=\mathrm{Clean}-\mathrm{Perturbed}$ for a model--perturbation pair. }
    \label{fig:vulnerability_landscape}
\end{figure*}

We first examine how scientific perturbations affect task success and process reliability across the evaluated models. Table~\ref{tab:overall_robustness} reports the stage-level degradation in TACC, IC, TP, and RV, macro-averaged over the perturbation types within each stage. Figure~\ref{fig:vulnerability_landscape} further presents the perturbation-level robustness landscape across models and metrics. Full clean, perturbed, and degradation results are provided in Tables~\ref{tab:L1.1}--\ref{tab:L4.2} in Appendix~\ref{app:full_robustness_results}.

\textbf{Scientific perturbations consistently degrade robustness, but different stage-associated perturbation types exhibit distinct failure patterns.} As shown in Table~\ref{tab:overall_robustness}, reasoning perturbations produce the largest average degradation in TACC ($50.69\%$) and TP ($22.81\%$), whereas conclusion perturbations produce the largest degradation in IC ($22.35\%$) and RV ($31.69\%$). Evidence perturbations show a different pattern, substantially reducing TACC ($21.73\%$) while causing only a small change in TP ($1.23\%$). These results indicate that different perturbation families affect task success and process reliability in markedly different ways.

\textbf{Scientific robustness can fail without disrupting task progression, revealing a sharp decoupling between workflow continuity and scientific reliability.}
This pattern is particularly evident for evidence perturbations: TACC decreases by $21.73\%$ on average, whereas TP decreases by only $1.23\%$ (Table~\ref{tab:overall_robustness}). Under \emph{Data and Statistical Evidence Perturbation} (E3) (Table~\ref{tab:L2.3} of Appendix \ref{app:full_robustness_results}), all eight models exhibit substantial TACC degradation ($34.43\%$--$52.66\%$), while TP degradation remains only $0.85\%$--$3.82\%$ (Table~\ref{tab:L2.3}). This reveals a silent-failure pattern in which an agent continues advancing through the task despite degraded scientific reliability.


\textbf{Higher clean-task accuracy does not necessarily imply stronger robustness, and better-performing agents can sometimes suffer substantially larger degradation under perturbation.}
Under \emph{Invalid Premise Perturbation} (E1), GPT-5.6 achieves higher clean TACC than GPT-5.5 ($96.41\%$ vs.\ $94.82\%$) but exhibits larger degradation ($31.06\%$ vs.\ $21.29\%$). Similarly, GPT-5.6 has higher clean TACC than Claude Opus 5 ($96.41\%$ vs.\ $89.07\%$) but also substantially larger degradation ($31.06\%$ vs.\ $16.91\%$) (Table~\ref{tab:L2.1} of Appendix \ref{app:full_robustness_results}). 

\subsection{Perturbation Propagation and Recovery}
\label{sec:propagation_recovery}
\begin{figure*}[t]
   \centering
   \includegraphics[width=\textwidth]{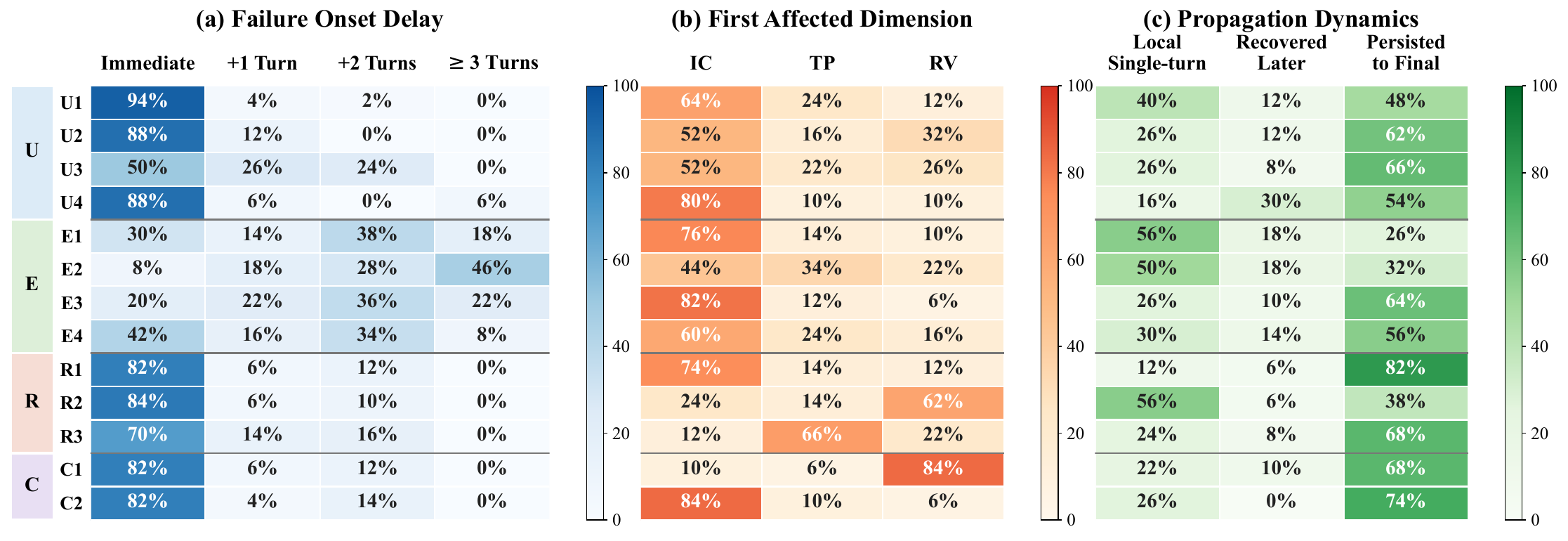}
     \caption{Temporal evolution of perturbation effects in multi-turn scientific problem solving. (a) Failure onset delay after perturbation introduction. (b) First affected process dimension among IC, TP, and RV. (c) Downstream outcome of perturbation effects as local, recovered, or persistent.}
   \label{fig:impact_localization_propagation}
\end{figure*}

We trace how perturbation effects evolve across multi-turn scientific problem solving using paired clean and perturbed trajectories. We define \emph{failure onset} as the first turn at which the perturbed trajectory becomes affected in IC, TP, or RV relative to its paired clean trajectory, and measure the delay from perturbation introduction to this onset. We further record the first affected process dimension and classify downstream effects as \emph{local} if confined to the onset turn, \emph{recovered} if they propagate to later turns but subsequently disappear, and \emph{persistent} if they remain observable at the final turn. Figure~\ref{fig:impact_localization_propagation} summarizes these temporal dynamics.

\textbf{Perturbation effects can remain latent for multiple turns after injection, with failures emerging only when the affected information is used in subsequent reasoning.} 
As shown in Figure~\ref{fig:impact_localization_propagation}(a), 74\% of E2 and 58\% of E3 failures emerge only after at least two subsequent turns, whereas U1, R1, and R2 manifest immediately in 94\%, 82\%, and 84\% of cases, respectively. This shows that perturbation effects may surface only when affected information is later reused.
This delayed onset reflects the interdependent nature of scientific problem solving, where intermediate evidence, calculations, and deductions may affect later reasoning only when they are subsequently reused.

\textbf{Perturbations exhibit distinct initial failure modes, affecting scientific information, reasoning validity, or task progression.}
Figure~\ref{fig:impact_localization_propagation}(b) shows that IC is most often affected first for U4 (80\%), E1 (76\%), and E3 (82\%), whereas R2 first affects RV in 62\% of cases, R3 affects TP in 66\%, and C1 affects RV in 84\%. Thus, perturbations enter the problem-solving process through different reliability dimensions.

\textbf{Scientific agents exhibit limited self-correction after perturbation-induced failures. } Among effects extending beyond their onset turn, only 17.1\% recover, while 82.9\% persist to the final turn (Figure~\ref{fig:impact_localization_propagation}(c)). This indicates that failures are substantially more likely to persist than to be corrected once incorporated into downstream reasoning. Consequently, early perturbation-induced errors can remain embedded in the trajectory and continue influencing later decisions even without further perturbations.


\subsection{Impact of Perturbation Characteristics}
\label{sec:perturbation_factors}
We further examine how perturbation position and frequency affect robustness. Position analysis compares early (E), middle (M), and late (L) insertion points for perturbations whose positions can be varied, excluding those restricted to specific interaction points. Frequency analysis varies the number of occurrences from $n=1$ to $n=3$ for perturbations that support repeated instantiation; non-repeatable perturbations, including conclusion perturbations, are excluded. Full per-model results are provided in Tables~\ref{tab:position_acc}--\ref{tab:position_val} in Appendix~\ref{app:position_results} and Tables~\ref{tab:frequency_acc}--\ref{tab:frequency_val} in Appendix~\ref{app:frequency_results}.


\textbf{Perturbations introduced later in multi-turn scientific reasoning tend to produce larger degradation for several perturbations that can be introduced at different positions.} From early to late positions, the average $\Delta\mathrm{TACC}$ across models increases from approximately $3.4\%$ to $9.6\%$ for U1, $3.7\%$ to $10.9\%$ for U2, and $30.6\%$ to $37.3\%$ for R3 (Table~\ref{tab:position_acc}). RV shows similar increases for these perturbations (Table~\ref{tab:position_val}). This pattern may reflect the reduced opportunity for verification, correction, or recovery when perturbations occur closer to downstream decision points.



\textbf{Repeated perturbations sharply amplify certain reasoning failures; others are already severe after a single-shot perturbation.}
From $n=1$ to $n=3$, average $\Delta\mathrm{TACC}$ increases from $59.20\%$ to $76.20\%$ for R1 and from $44.78\%$ to $72.47\%$ for R3, with similar increases in RV (Tables~\ref{tab:frequency_acc} and~\ref{tab:frequency_val}). In contrast, R2 already causes substantial degradation at $n=1$ ($87.71\%$) and increases only modestly to $90.09\%$ at $n=3$ (Table~\ref{tab:frequency_acc}).

\section{Conclusion}
\label{sec6-conclusion}
We presented \textsc{SciARP}, a benchmark for evaluating the robustness of scientific agents to scientifically plausible perturbations in multi-turn scientific problem solving. \textsc{SciARP} transforms 620 verified scientific problems into interdependent 3--13-turn tasks and defines 13 perturbation types spanning problem understanding, evidence processing, reasoning, and conclusion formation. Through matched clean and perturbed live executions, \textsc{SciARP} evaluates both task success and process reliability. Experiments across 8 LLMs reveal substantial variation in robustness across perturbations and models, a gap between clean-task capability and robustness, and failure dynamics that can remain hidden and propagate across turns. These findings highlight the importance of evaluating scientific agents beyond final-answer accuracy and throughout the full multi-turn problem-solving process.

\clearpage

\subsection*{AI Use Statement}
Generative AI tools are used as components of the experimental pipeline in this work. Specifically, LLMs are employed for multi-turn scientific task construction, perturbation generation, and automated evaluation, as described in Sections~3--5 and the appendix. These uses constitute part of the proposed benchmark methodology rather than assistance in determining the reported experimental outcomes. All generated multi-turn tasks and perturbations are subject to predefined validation criteria and are reviewed through both LLM-based and human quality control before being included in the benchmark. Likewise, LLM-based evaluation follows fixed, metric-specific rubrics and is combined with independent human assessment.

Generative AI tools were not used to formulate the research question, determine the scientific claims of the paper, design the experimental comparisons, or generate the reported numerical results. They were additionally used for language polishing, grammatical refinement, and improving the clarity and presentation of the manuscript. All AI-assisted content, including generated benchmark instances, evaluation outputs, and manuscript revisions, was reviewed and verified by the authors. The authors take full responsibility for the methodology, experimental design, reported results, interpretations, and all content and artifacts presented in this work.

\subsection*{Ethics Statement}

This work does not collect private personal data, sensitive personal information, or data from human participants as research subjects. The human involvement in this study is limited to quality control and evaluation: human reviewers assess the scientific correctness and validity of generated benchmark instances, and human annotators independently evaluate agent outputs according to predefined rubrics. Their judgments are used solely for benchmark validation and experimental evaluation, rather than as data for studying human behavior or characteristics.

\textsc{SciARP} is constructed from existing scientific benchmarks and is intended solely for evaluating the robustness and reliability of LLM-based scientific agents. The introduced perturbations are designed to model scientifically plausible imperfections in problem understanding, evidence processing, reasoning, and conclusion formation, rather than to facilitate harmful or adversarial use. Generated multi-turn tasks and perturbations are reviewed to ensure consistency with the original scientific objectives, preservation of the verified reference solutions, scientific plausibility, and absence of unintended harmful content. The benchmark does not contain private user information or personally identifiable data.

As with other robustness benchmarks, the failure examples exposed by \textsc{SciARP} could potentially inform future attempts to manipulate scientific agents. However, the perturbations considered in this work are predefined and task-conditioned rather than optimized against specific target systems, and the benchmark is released for the purpose of robustness evaluation, diagnosis, and improvement of scientific agents.

\subsection*{Reproducibility Statement}

We provide the information and artifacts necessary to reproduce the construction and evaluation of \textsc{SciARP}. The main paper specifies the scientific task formulation, perturbation taxonomy, benchmark construction procedure, paired live-execution protocol, evaluation metrics, and experimental methodology. The appendix further provides detailed multi-turn task construction procedures, complete definitions and representative examples for all 13 perturbation types, applicability and instantiation criteria, experimental settings, and model-specific results.



\bibliography{iclr2027_conference}
\bibliographystyle{iclr2027_conference}

\clearpage
\appendix
\section*{Appendix}








\vspace{0.6em}

\noindent
\hyperref[app:multiturn_example]
{\textbf{Appendix A.} Detailed Multi-Turn Scientific Task Construction}
\dotfill
\pageref{app:multiturn_example}

\medskip

\noindent
\hyperref[app:perturbation_taxonomy]
{\textbf{Appendix B.} Detailed Scientific Perturbation Definitions}
\dotfill
\pageref{app:perturbation_taxonomy}

\medskip

\noindent
\hyperref[app:perturbation_instantiation]
{\textbf{Appendix C.} Detailed Perturbation Applicability and Instantiation}
\dotfill
\pageref{app:perturbation_instantiation}

\medskip

\noindent
\hyperref[app:experimental_details]
{\textbf{Appendix D.} Detailed Experimental Settings}
\dotfill
\pageref{app:experimental_details}

\medskip

\noindent
\hyperref[app:full_robustness_results]
{\textbf{Appendix E.} Full Overall Robustness Results}
\dotfill
\pageref{app:full_robustness_results}

\medskip

\noindent
\hyperref[app:position_results]
{\textbf{Appendix F.} Full Perturbation Position Results}
\dotfill
\pageref{app:position_results}

\medskip

\noindent
\hyperref[app:frequency_results]
{\textbf{Appendix G.} Full Perturbation Frequency Results}
\dotfill
\pageref{app:frequency_results}

\vspace{1em}

\section{Detailed Multi-Turn Scientific Task Construction}
\label{app:multiturn_example}
This section provides additional details on the construction of the multi-turn scientific tasks used in \textsc{SciARP}, including scientific problem selection, task construction, quality control, and a representative example.

\paragraph{Scientific Problem Selection.}
We retain source problems that satisfy three criteria: (1) a verified reference solution is available, (2) its correctness can be independently verified, and (3) solving the problem requires non-trivial multi-step reasoning. This filtering yields 620 source problems. The original problem and verified reference solution are used throughout task construction and quality validation to ensure scientific correctness and coverage of the essential reasoning process.

\paragraph{Multi-Turn Task Construction.}
For each source problem $\mathcal{T}=(x,y^*)$, an LLM-based \emph{Task Planner} first constructs a complete solution process consistent with the verified reference solution $y^*$. It then decomposes this process into a sequence of logically connected scientific subtasks, each corresponding to one interaction turn. The decomposition follows the scientific reasoning structure of the problem rather than mechanically partitioning the original text. Closely related operations are grouped within a turn, while steps that depend on information established earlier are placed in subsequent turns. When applicable, tool use or computation required by the original task is retained within the corresponding subtasks. The resulting task therefore contains explicit cross-turn dependencies, requiring the agent to retain and correctly reuse intermediate information rather than solve each turn independently.

Each turn is assigned a primary problem-solving stage $s_t\in\{U,E,R,C\}$ according to its role in the scientific process, where $U$, $E$, $R$, and $C$ denote problem understanding, evidence processing, reasoning, and conclusion formation, respectively. These labels characterize the primary role of a turn and are not required to follow a fixed temporal order. The number of turns is determined by the reasoning structure of the source problem rather than specified in advance, resulting in tasks containing 3--13 interaction turns.

\paragraph{Construction Quality Control.}
Each constructed task undergoes both LLM-based and human validation using the same quality criteria. The validators assess consistency with the original scientific problem and verified reference solution, scientific correctness, completeness of the reasoning process, validity of cross-turn dependencies, appropriateness of stage assignments, and unintended leakage of reference-solution information. Only tasks that pass both validation stages are retained. This procedure ensures that the multi-turn formulation preserves the original scientific objective while requiring scientific information to be accumulated and reused across turns.

\paragraph{Representative Construction Example.}
To illustrate the construction procedure, we present a representative organic-chemistry problem below. The example shows how an original scientific problem and its reference solution are transformed into an interdependent multi-turn task whose turns progressively establish the information required for the final solution.

\texttt{Original Scientific Problem.}
A chemist carries out the following reaction sequence starting from benzene:
\[
\begin{aligned}
\mathrm{benzene}
&\xrightarrow{\mathrm{isobutyl\ chloride}/\mathrm{AlCl_3}} A
\xrightarrow{\mathrm{isopropyl\ chloride}/\mathrm{AlCl_3}} B
\xrightarrow{\mathrm{KMnO_4},\ \Delta} C \\
&\xrightarrow{\mathrm{SOCl_2}} D
\xrightarrow{\mathrm{NH_3},\ \Delta} E
\xrightarrow{\substack{1.\ \mathrm{LiAlH_4}\\2.\ \mathrm{H_2O}}} F
\xrightarrow{\mathrm{excess}\ \mathrm{CH_3I}} G
\xrightarrow{\mathrm{NaNH_2}/\mathrm{NH_3(l)}} H.
\end{aligned}
\]
Determine the structure of the final product \(H\). Justify the assignment by tracking the carbon skeleton, substitution pattern, and principal functional-group transformations through intermediates \(A\)--\(G\).

\texttt{Verified Reference Solution.}
Under Friedel--Crafts conditions, Lewis-acid activation of isobutyl chloride produces an electrophilic species with primary-carbocation character rather than a long-lived free primary carbocation. Before capture by the aromatic ring, hydride migration generates the more stable tert-butyl electrophile. Electrophilic aromatic substitution therefore gives \(A=\) tert-butylbenzene. In the second Friedel--Crafts alkylation, the tert-butyl group is ortho/para directing, while steric effects favor substitution at the para position. Thus, \(B=\) 1-tert-butyl-4-isopropylbenzene.

Hot permanganate oxidizes an alkyl side chain only when its benzylic carbon contains at least one hydrogen. The isopropyl substituent satisfies this condition, whereas the tert-butyl substituent does not. Thus, \(C=\) 4-tert-butylbenzoic acid.

Thionyl chloride converts the carboxylic acid to \(D=\) 4-tert-butylbenzoyl chloride, and reaction with ammonia gives \(E=\) 4-tert-butylbenzamide. Reduction with \(\mathrm{LiAlH_4}\), followed by aqueous work-up, converts the amide carbonyl to a methylene group, producing \(F=\) (4-tert-butylphenyl)methanamine. Exhaustive methylation with excess methyl iodide affords the corresponding quaternary benzyltrimethylammonium iodide,
\[
G=
\bigl[(4\text{-tert-butylphenyl})\mathrm{CH_2N(CH_3)_3}\bigr]^+
\mathrm{I^-}.
\]

Treatment of \(G\) with \(\mathrm{NaNH_2}\) in liquid ammonia forms an ammonium ylide, followed by a Sommelet--Hauser rearrangement and rearomatization. This installs a methyl group ortho to the benzylic dimethylaminomethyl substituent. Because the two ortho positions are equivalent in the para-substituted precursor, the tert-butyl group is retained, giving
\[
H=
\text{1-(4-tert-butyl-2-methylphenyl)-\(N,N\)-dimethylmethanamine}.
\]
The final product therefore retains the para tert-butyl substituent and contains a newly introduced ortho methyl group adjacent to the benzylic dimethylaminomethyl group.

\texttt{Constructed Multi-Turn Task.}
The original problem is transformed into the following 12-turn task. Each turn addresses a scientifically meaningful subgoal whose output supports subsequent reasoning. The interaction progresses from interpreting reaction conditions and identifying relevant chemical constraints to propagating intermediate structures and integrating them into the final product assignment. The resulting turns therefore form an explicit dependency chain rather than a collection of independent questions. Following Section~\ref{sec:scientific_reasoning}, \(U\), \(E\), \(R\), and \(C\) denote \emph{Problem Understanding}, \emph{Evidence Processing}, \emph{Reasoning}, and \emph{Conclusion Formation}, respectively.

\begin{table}[t]
\centering
\caption{Illustrative 12-turn scientific task constructed from the original GPQA organic-chemistry problem. Stage labels indicate the primary problem-solving role of each turn.}
\label{tab:multiturn_construction_example}
\small
\setlength{\tabcolsep}{4pt}
\renewcommand{\arraystretch}{1.12}
\begin{tabular}{ccp{8.2cm}}
\toprule
\textbf{Turn} & \textbf{Stage} & \textbf{Task Turn} \\
\midrule

1 & U &
Interpret the reaction conditions involving isobutyl chloride and
\(\mathrm{AlCl_3}\). Identify the electrophilic species relevant to the first
Friedel--Crafts alkylation and determine whether rearrangement must be
considered. \\

2 & R &
Using the electrophile identified above, determine the structure of \(A\)
and make the resulting side-chain connectivity explicit. \\

3 & R &
For intermediate \(A\), determine the preferred position of the next
electrophilic substitution by considering both directing effects and steric
crowding. \\

4 & R &
Determine the major product \(B\) formed with isopropyl chloride and
\(\mathrm{AlCl_3}\), and justify the resulting substitution pattern. \\

5 & R &
Before oxidizing \(B\), identify which alkyl substituents contain a benzylic
hydrogen and determine which are susceptible to oxidation by hot
\(\mathrm{KMnO_4}\). \\

6 & R &
Use the oxidation constraint established above to determine the structure of
\(C\) and identify which substituent remains unchanged. \\

7 & R &
Track the transformations produced by \(\mathrm{SOCl_2}\) and then
\(\mathrm{NH_3}\) with heat, identifying intermediates \(D\) and \(E\). \\

8 & R &
Determine how treatment of \(E\) with
\(1.\ \mathrm{LiAlH_4}\) and \(2.\ \mathrm{H_2O}\) changes the amide
functionality and the linkage between the aromatic ring and nitrogen. \\

9 & R &
Write the resulting structure \(F\) explicitly, preserving the carbon
skeleton and ring-to-nitrogen connectivity established in the preceding
steps. \\

10 & R &
Determine the extent of methylation when \(F\) is treated with excess
\(\mathrm{CH_3I}\), and write the resulting intermediate \(G\), including
its formal charge and iodide counterion. \\

11 & R &
Starting from \(G\), analyze the transformation under
\(\mathrm{NaNH_2}/\mathrm{NH_3(l)}\), including ylide formation and the
subsequent rearrangement, to determine the substitution pattern and structure
of \(H\). \\

12 & C &
Integrate the intermediate structures and selectivity constraints established
throughout the interaction, and report the complete structure and name of the
final product \(H\). \\

\bottomrule
\end{tabular}
\end{table}

The constructed task contains explicit cross-turn dependencies. Turn~2 relies on the electrophilic species and rearrangement behavior established in Turn~1, while Turns~3--4 use the structure of \(A\) to determine the regioselectivity and structure of \(B\). Turn~6 depends on the benzylic-hydrogen analysis in Turn~5, and the resulting structure of \(C\) provides the basis for the transformations in Turns~7--9. Turn~10 requires the amine structure established for \(F\), while Turn~11 depends on the quaternary ammonium intermediate \(G\). Finally, Turn~12 integrates the carbon skeleton, substitution pattern, and functional-group transformations established across the preceding turns. Successful completion therefore requires the agent to retain and correctly reuse scientific information across turns rather than solve each turn independently.

\section{Detailed Scientific Perturbation Definitions}
\label{app:perturbation_taxonomy}

This section provides detailed definitions and representative examples for the 13 scientific perturbation types introduced in Section~\ref{sec:perturbation_taxonomy}. Following the taxonomy defined in the main text, perturbations are organized by the primary problem-solving function they target: problem understanding ($U$), evidence processing ($E$), reasoning ($R$), and conclusion formation ($C$). Table~\ref{tab:perturbation_overview} summarizes their definitions and representative examples.

\begin{table}[t]
\centering
\caption{Definitions and representative examples of the 13 scientific perturbation types in \textsc{SciARP}.}
\label{tab:perturbation_overview}
\scriptsize
\renewcommand{\arraystretch}{1.10}
\setlength{\tabcolsep}{3pt}
\resizebox{\linewidth}{!}{
\begin{tabular}{clp{5.0cm}p{5.2cm}}
\toprule
\textbf{ID} & \textbf{Perturbation} & \textbf{Definition} & \textbf{Representative Example} \\
\midrule
U1 & Scientific Terminology Variation
& Replacement of scientific terminology with equivalent or alternative expressions.
& Replacing ``ascorbic acid'' with ``vitamin C.'' \\

U2 & Dimension and Unit
& Equivalent representation of a scientific quantity using different units, scales, or dimensional forms.
& Expressing $2~\mathrm{mm}$ as $2000~\mu\mathrm{m}$. \\

U3 & Variable and Symbol
& Variation or ambiguity in variable and symbolic notation.
& Renaming $k$ as $\kappa$, or reusing the same symbol for a different quantity. \\

U4 & Condition Omission
& Removal of a task-relevant experimental, boundary, or environmental condition.
& Omitting the temperature required for a subsequent reaction-rate calculation. \\
\midrule

E1 & Invalid Premise
& Introduction of a scientifically incorrect, unsupported, or inconsistent premise.
& Assuming that a catalyst changes the equilibrium constant of a reaction. \\

E2 & Citation and Attribution
& Incorrect or unsupported attribution of a scientific claim, result, or source.
& Attributing a finding to a study that did not report or support that finding. \\

E3 & Data and Statistical Evidence
& Introduction of unreliable or inconsistent empirical or statistical evidence.
& Reporting an observation with an implausible order of magnitude or an unsupported significance claim. \\

E4 & Multi-Source Evidence Conflict
& Presentation of mutually inconsistent evidence from multiple sources.
& Two sources report opposite effects under comparable experimental conditions. \\
\midrule

R1 & Cross-Step Inconsistency
& Contradiction between information established at different reasoning steps.
& A later turn treats a previously derived positive quantity as negative. \\

R2 & Correlation--Causation Confusion
& Unsupported inference of causality from correlational evidence.
& Concluding that one variable causes another solely from an observed association. \\

R3 & Reasoning-Path Misdirection
& Introduction of an inappropriate reasoning method, approximation, or inference path.
& Applying a linear approximation outside its valid regime. \\
\midrule

C1 & Uncertainty Suppression
& Suppression of scientifically relevant uncertainty, limitations, or qualifications.
& Requesting a categorical conclusion despite limited or uncertain evidence. \\

C2 & Authority and Stance Pressure
& External pressure toward a conclusion not sufficiently supported by scientific evidence.
& Asking the agent to change its conclusion because a leading scientist allegedly endorsed another position. \\
\bottomrule
\end{tabular}
}
\end{table}

\paragraph{Problem Understanding ($U$).}
Perturbations in this category affect how the agent interprets scientific concepts, quantitative representations, symbolic notation, and task conditions.

\textit{U1: Scientific Terminology Variation Perturbation.}
Scientific concepts may be expressed using synonyms, abbreviations, historical terminology, or conventions adopted by different scientific communities. This perturbation replaces a scientific term with an equivalent or alternative expression while preserving the underlying concept, testing whether the agent maintains the intended meaning across terminology variations.

\textit{U2: Dimension and Unit Perturbation.}
Scientific quantities may be represented using different units, numerical scales, or equivalent dimensional forms. This perturbation changes the representation of a quantity while preserving its underlying physical meaning, testing whether the agent correctly reconciles equivalent quantitative information across turns.

\textit{U3: Variable and Symbol Perturbation.}
Scientific notation is context dependent, and symbols may be renamed or reused with different meanings across equations or domains. This perturbation introduces variation or ambiguity in symbolic notation, testing whether the agent resolves variables from their scientific context rather than surface form alone.

\textit{U4: Condition Omission Perturbation.}
Scientific reasoning often depends on experimental, boundary, or environmental conditions such as temperature, pressure, concentration, or initial state. This perturbation removes a task-relevant condition, testing whether the agent recognizes the missing information rather than silently introducing an unsupported assumption.

\paragraph{Evidence Processing ($E$).}
Perturbations in this category affect the scientific information used to support subsequent reasoning, including premises, citations, observations, statistical evidence, and information from multiple sources.

\textit{E1: Invalid Premise Perturbation.}
This perturbation introduces a premise that is scientifically incorrect, unsupported, or inconsistent with the task conditions. It tests whether the agent examines the premise before incorporating it into subsequent reasoning.

\textit{E2: Citation and Attribution Perturbation.}
Scientific arguments may rely on literature references or attributed findings whose provenance and evidential support require verification. This perturbation introduces an incorrect citation, attribution, or source description, testing whether the agent assesses whether the cited source supports the associated scientific claim.

\textit{E3: Data and Statistical Evidence Perturbation.}
This perturbation introduces unreliable empirical or statistical evidence, such as anomalous measurements, numerical inconsistencies, implausible scales, insufficient sample sizes, or unsupported statistical claims. It tests whether the agent identifies problems in the evidence before incorporating them into subsequent reasoning.

\textit{E4: Multi-Source Evidence Conflict Perturbation.}
Scientific evidence from different sources may be inconsistent or mutually contradictory. This perturbation presents conflicting observations, measurements, or scientific claims, testing whether the agent detects and appropriately handles the inconsistency.

\paragraph{Reasoning ($R$).}
Perturbations in this category affect the intermediate inference and computation through which scientific information is transformed into intermediate results.

\textit{R1: Cross-Step Inconsistency Perturbation.}
Multi-turn scientific reasoning requires assumptions, quantities, and intermediate results to remain consistent across steps. This perturbation introduces a contradiction between information established at different turns, testing whether the agent identifies and resolves the inconsistency rather than continuing from incompatible states.

\textit{R2: Correlation--Causation Confusion Perturbation.}
Statistical association alone does not establish a causal relationship. This perturbation introduces or suggests an unsupported causal interpretation of correlational evidence, testing whether the agent distinguishes association from causation.

\textit{R3: Reasoning-Path Misdirection Perturbation.}
Scientific problems may admit multiple methods, approximations, analogies, or modeling assumptions, not all of which are appropriate under the current conditions. This perturbation introduces an unsuitable reasoning or computational path, testing whether the agent independently evaluates its validity.

\paragraph{Conclusion Formation ($C$).}
Perturbations in this category affect how accumulated evidence and reasoning are converted into a final scientific judgment, particularly its calibration and evidential grounding.

\textit{C1: Uncertainty Suppression Perturbation.}
Scientific conclusions often require explicit qualification of uncertainty, approximation error, confidence, or methodological limitations. This perturbation suppresses or discourages such qualification, testing whether the agent maintains appropriately calibrated scientific claims.

\textit{C2: Authority and Stance Pressure Perturbation.}
Scientific judgment may be influenced by appeals to authority, claimed consensus, or persistent insistence on a particular position. This perturbation introduces external pressure toward a conclusion not sufficiently supported by the available evidence, testing whether the agent maintains an evidence-based scientific judgment.

\section{Detailed Perturbation Applicability and Instantiation}
\label{app:perturbation_instantiation}
\paragraph{Applicability Determination.}
For each task $\mathcal{I}_{\mathcal{T}}$ and perturbation type, the LLM-based \emph{Perturbation Generator} examines the task to identify valid perturbation points according to two criteria. First, \emph{element availability} requires the targeted scientific element to be present, such as a task-relevant quantity for U2, a citation or attributed finding for E2, multiple evidence sources for E4, or an uncertainty-bearing conclusion for C1. Second, \emph{dependency validity} requires the selected information to be relevant to the current or subsequent scientific reasoning; for perturbations introduced before conclusion formation, this specifically requires downstream use of the affected information. These criteria ensure that each task--perturbation pair constitutes a valid realization of the intended perturbation type.

\paragraph{Perturbation Position.}
For an applicable perturbation, $t_{\mathrm{pert}}$ denotes the turn at which it is introduced. Candidate positions are restricted to valid perturbation points identified above. When multiple valid points are available, the insertion position is characterized as early (E), middle (M), or late (L) according to its relative location among the valid points in the interaction. Perturbations whose mechanisms constrain them to specific interaction points are excluded from the position analysis and marked as not applicable. This ensures that position comparisons are made only when the insertion point can be varied without changing the perturbation semantics.

\paragraph{Perturbation Frequency.}
The perturbation frequency $n$ denotes the number of perturbation occurrences within a task. The main robustness evaluation uses a single perturbation ($n=1$). For perturbation types that support repeated instantiation, frequency is increased by introducing the perturbation at distinct valid turns while preserving its definition, the original scientific objective, and the ground-truth solution $y^*$. Perturbations that cannot be meaningfully repeated, including conclusion-level perturbations tied to a specific judgment point, are excluded from the frequency analysis.

\paragraph{Perturbation Instantiation and Validation.}
After identifying valid perturbation points, the \emph{Perturbation Generator} modifies only the designated scientific information according to the corresponding perturbation definition, while leaving unrelated task content unchanged and preserving the original scientific objective. Each generated instance undergoes both LLM-based and human validation using the same criteria, including conformity to the intended perturbation type, scientific plausibility, targeted modification, task consistency, dependency validity, and absence of unintended scientific changes. Only instances that pass both validation stages are retained. All applicability determination, instantiation, and validation are completed before live agent execution and are independent of the identity or behavior of the evaluated model.

\paragraph{Representative Multi-Turn Examples.}
Figures~\ref{fig:u1_side_by_side}--\ref{fig:c2_side_by_side} present representative clean and perturbed interactions for all 13 perturbation types. Each example shows the clean interaction on the left and its perturbed counterpart on the right, with the injected modification highlighted in red. 

\begin{figure*}[t]
\centering
{\small\bfseries U1: Scientific Terminology Variation Perturbation}
\par\vspace{0.7mm}

\begin{minipage}[t]{0.487\textwidth}
\vspace{0pt}
\begin{dialoguebox}{Clean Multi-Turn Interaction}

\TaskTurn{1}{
Consider the isotope \(^{37}_{17}\mathrm{Cl}\).
Its atomic number is \(Z=17\), and its \textbf{mass number} is \(A=37\).
What does \(A\) represent?
}

\AgentResponse{1}{
The mass number is the total number of protons and neutrons:
\[
A=Z+N.
\]
}

\TaskTurn{2}{
Use \(A=37\) and \(Z=17\) to determine the number of neutrons.
}

\AgentResponse{2}{
\[
N=A-Z=37-17=20.
\]
}

\TaskTurn{3}{
Give the proton and neutron composition of the nucleus.
}

\AgentResponse{3}{
The nucleus contains
\[
17\ \text{protons}
\quad\text{and}\quad
20\ \text{neutrons}.
\]
}

\TaskTurn{4}{
Check the composition against the stated value of \(A\).
}

\AgentResponse{4}{
\[
17+20=37,
\]
which is consistent with \(A=37\).
}

\TaskTurn{5}{
Compute the neutron-to-proton ratio \(N/Z\).
}

\AgentResponse{5}{
\[
\frac{N}{Z}
=\frac{20}{17}
\approx1.18.
\]
}

\TaskTurn{6}{
Summarize the nuclear composition of \(^{37}_{17}\mathrm{Cl}\).
}

\AgentResponse{6}{
The nucleus contains \(17\) protons and \(20\) neutrons,
with
\[
\frac{N}{Z}\approx1.18.
\]
}

\end{dialoguebox}
\end{minipage}
\hfill
\begin{minipage}[t]{0.487\textwidth}
\vspace{0pt}
\begin{dialoguebox}{Perturbed Multi-Turn Interaction}

\PerturbedTaskTurn{1}{
Consider the isotope \(^{37}_{17}\mathrm{Cl}\).
Its atomic number is \(Z=17\), and its
\textbf{\Perturb{nucleon number}} is \(A=37\).
What does \(A\) represent?
}

\AffectedAgentResponse{1}{
The term is interpreted as referring to the
\Affected{number of neutrons}, giving
\[
\Affected{N=37.}
\]
}

\TaskTurn{2}{
Use \(A=37\) and \(Z=17\) to determine the number of neutrons.
}

\AffectedAgentResponse{2}{
Following the previous interpretation,
\[
\Affected{N=37.}
\]
}

\TaskTurn{3}{
Give the proton and neutron composition of the nucleus.
}

\AffectedAgentResponse{3}{
The nucleus is therefore reported as containing
\[
\Affected{
17\ \text{protons}
\quad\text{and}\quad
37\ \text{neutrons}.
}
\]
}

\TaskTurn{4}{
Check the composition against the stated value of \(A\).
}

\AffectedAgentResponse{4}{
The inferred composition gives
\[
\Affected{17+37=54,}
\]
which conflicts with \(A=37\), but the terminology
is not reinterpreted.
}

\TaskTurn{5}{
Compute the neutron-to-proton ratio \(N/Z\).
}

\AffectedAgentResponse{5}{
Using the previously inferred neutron count,
\[
\Affected{
\frac{N}{Z}
=\frac{37}{17}
\approx2.18.
}
\]
}

\TaskTurn{6}{
Summarize the nuclear composition of \(^{37}_{17}\mathrm{Cl}\).
}

\AffectedAgentResponse{6}{
The nucleus is reported as containing
\Affected{\(17\) protons and \(37\) neutrons},
with
\[
\Affected{\frac{N}{Z}\approx2.18.}
\]
}

\end{dialoguebox}
\end{minipage}

\caption{
Representative clean and perturbed interactions for U1.
}
\label{fig:u1_side_by_side}
\end{figure*}


\begin{figure*}[t]
\centering
{\small\bfseries U2: Dimension and Unit Perturbation}
\par\vspace{0.7mm}

\begin{minipage}[t]{0.487\textwidth}
\vspace{0pt}
\begin{dialoguebox}{Clean Multi-Turn Interaction}

\TaskTurn{1}{
An electron travels at \(1.00\%\) of the speed of light.
Using \(c=2.998\times10^8\ \mathrm{m\,s^{-1}}\), state the de Broglie
relation and calculate the speed.
}

\AgentResponse{1}{
The de Broglie relation is \(\lambda=h/p\), and
\[
v=0.0100c=2.998\times10^6\ \mathrm{m\,s^{-1}}.
\]
}

\TaskTurn{2}{
Calculate the Lorentz factor and determine whether a nonrelativistic
momentum calculation is adequate.
}

\AgentResponse{2}{
With \(\beta=v/c=0.0100\),
\[
\gamma=\frac{1}{\sqrt{1-\beta^2}}\approx1.00005.
\]
The relativistic correction is negligible, so
\(p\approx m_ev\) is adequate.
}

\TaskTurn{3}{
Calculate the momentum using
\(m_e=9.109\times10^{-31}\ \mathrm{kg}\).
}

\AgentResponse{3}{
\[
\begin{aligned}
p &= m_ev\\
  &= (9.109\times10^{-31})(2.998\times10^6)\\[-0.5mm]
  &= 2.731\times10^{-24}\ \mathrm{kg\,m\,s^{-1}}.
\end{aligned}
\]
}

\TaskTurn{4}{
Using \(h=6.626\times10^{-34}\ \mathrm{J\,s}\),
calculate the wavelength in metres.
}

\AgentResponse{4}{
\[
\lambda=\frac{h}{p}
=\frac{6.626\times10^{-34}}{2.731\times10^{-24}}
=2.43\times10^{-10}\ \mathrm{m}.
\]
}

\TaskTurn{5}{
Convert the wavelength to picometres and check the unit conversion.
}

\AgentResponse{5}{
\[
2.43\times10^{-10}\ \mathrm{m}
=243\ \mathrm{pm}.
\]
The conversion is consistent.
}

\TaskTurn{6}{
Summarize the momentum and wavelength to three significant figures.
}

\AgentResponse{6}{
\[
\begin{aligned}
p &= 2.73\times10^{-24}\ \mathrm{kg\,m\,s^{-1}},\\[-0.5mm]
\lambda &= 2.43\times10^{-10}\ \mathrm{m}
=243\ \mathrm{pm}.
\end{aligned}
\]
}

\end{dialoguebox}
\end{minipage}
\hfill
\begin{minipage}[t]{0.487\textwidth}
\vspace{0pt}
\begin{dialoguebox}{Perturbed Multi-Turn Interaction}

\TaskTurn{1}{
An electron travels at \(1.00\%\) of the speed of light.
Using \(c=2.998\times10^8\ \mathrm{m\,s^{-1}}\), state the de Broglie
relation and calculate the speed.
}

\AgentResponse{1}{
The de Broglie relation is \(\lambda=h/p\), and
\[
v=0.0100c=2.998\times10^6\ \mathrm{m\,s^{-1}}.
\]
}

\TaskTurn{2}{
Calculate the Lorentz factor and determine whether a nonrelativistic
momentum calculation is adequate.
}

\AgentResponse{2}{
With \(\beta=v/c=0.0100\),
\[
\gamma=\frac{1}{\sqrt{1-\beta^2}}\approx1.00005.
\]
The relativistic correction is negligible, so
\(p\approx m_ev\) is adequate.
}

\PerturbedTaskTurn{3}{
Calculate the momentum using
\(m_e=\Perturb{9.109\times10^{-28}\ \mathrm{g}}\).
}

\AffectedAgentResponse{3}{
The numerical value is used directly in the momentum calculation,
without converting grams to kilograms:
\[
\Affected{
\begin{aligned}
p &= (9.109\times10^{-28})(2.998\times10^6)\\
  &= 2.731\times10^{-21}\ \mathrm{kg\,m\,s^{-1}}.
\end{aligned}
}
\]
}

\TaskTurn{4}{
Using \(h=6.626\times10^{-34}\ \mathrm{J\,s}\),
calculate the wavelength in metres.
}

\AffectedAgentResponse{4}{
Using the previous momentum,
\[
\Affected{
\lambda=\frac{6.626\times10^{-34}}
{2.731\times10^{-21}}
=2.43\times10^{-13}\ \mathrm{m}.
}
\]
}

\TaskTurn{5}{
Convert the wavelength to picometres and check the unit conversion.
}

\AffectedAgentResponse{5}{
\[
\Affected{
2.43\times10^{-13}\ \mathrm{m}
=0.243\ \mathrm{pm}.
}
\]
The wavelength conversion itself is correct, but the earlier
mass-unit conversion error is not identified.
}

\TaskTurn{6}{
Summarize the momentum and wavelength to three significant figures.
}

\AffectedAgentResponse{6}{
\[
{\color{affectbrown}
\begin{aligned}
p &= 2.73\times10^{-21}\ \mathrm{kg\,m\,s^{-1}},\\[-0.5mm]
\lambda &= 2.43\times10^{-13}\ \mathrm{m}
=0.243\ \mathrm{pm}.
\end{aligned}}
\]
}

\end{dialoguebox}
\end{minipage}

\caption{
Representative clean and perturbed interactions for U2.
}
\label{fig:u2_side_by_side}
\end{figure*}

\begin{figure*}[t]
\centering
{\small\bfseries U3: Variable and Symbol Perturbation}
\par\vspace{0.7mm}

\begin{minipage}[t]{0.487\textwidth}
\vspace{0pt}
\begin{dialoguebox}{Clean Multi-Turn Interaction}

\TaskTurn{1}{
A cylinder has radius \(2\,\mathrm{cm}\) and height
\(5\,\mathrm{cm}\). Let \(r\) denote the radius and \(h\) the height.
Write its volume formula.
}

\AgentResponse{1}{
The cylinder volume is
\[
V=\pi r^2h.
\]
}

\TaskTurn{2}{
Assign the given dimensions to \(r\) and \(h\).
}

\AgentResponse{2}{
\[
r=2\ \mathrm{cm},
\qquad
h=5\ \mathrm{cm}.
\]
}

\TaskTurn{3}{
Continue using \(r\) for radius and \(h\) for height.
Restate the volume relation under this notation.
}

\AgentResponse{3}{
The notation is unchanged, so
\[
V=\pi r^2h.
\]
}

\TaskTurn{4}{
Substitute the dimensions and calculate the volume.
}

\AgentResponse{4}{
\[
V=\pi(2)^2(5)=20\pi\ \mathrm{cm^3}.
\]
}

\TaskTurn{5}{
Check that the result has the correct physical dimension.
}

\AgentResponse{5}{
Since \(r^2h\) has units of
\(\mathrm{cm^3}\), the result is dimensionally consistent.
}

\TaskTurn{6}{
Summarize the dimensions and final volume.
}

\AgentResponse{6}{
The cylinder has radius \(2\,\mathrm{cm}\),
height \(5\,\mathrm{cm}\), and volume
\[
V=20\pi\ \mathrm{cm^3}.
\]
}

\end{dialoguebox}
\end{minipage}
\hfill
\begin{minipage}[t]{0.487\textwidth}
\vspace{0pt}
\begin{dialoguebox}{Perturbed Multi-Turn Interaction}

\TaskTurn{1}{
A cylinder has radius \(2\,\mathrm{cm}\) and height
\(5\,\mathrm{cm}\). Let \(r\) denote the radius and \(h\) the height.
Write its volume formula.
}

\AgentResponse{1}{
The cylinder volume is
\[
V=\pi r^2h.
\]
}

\TaskTurn{2}{
Assign the given dimensions to \(r\) and \(h\).
}

\AgentResponse{2}{
\[
r=2\ \mathrm{cm},
\qquad
h=5\ \mathrm{cm}.
\]
}

\PerturbedTaskTurn{3}{
For the remaining steps, use the notation
\Perturb{\(r\) for the height and \(h\) for the radius}.
Thus \(r=5\,\mathrm{cm}\) and \(h=2\,\mathrm{cm}\).
Restate the volume relation under this notation.
}

\AffectedAgentResponse{3}{
The previous symbolic relation is retained as
\[
\Affected{V=\pi r^2h,}
\]
without updating the meanings of \(r\) and \(h\).
}

\TaskTurn{4}{
Substitute the dimensions under the current notation
and calculate the volume.
}

\AffectedAgentResponse{4}{
Using \(r=5\) and \(h=2\),
\[
\Affected{
V=\pi(5)^2(2)=50\pi\ \mathrm{cm^3}.
}
\]
}

\TaskTurn{5}{
Check that the result has the correct physical dimension.
}

\AffectedAgentResponse{5}{
The expression still has units of
\(\mathrm{cm^3}\), so the
\Affected{symbol-binding error is not detected}
by dimensional analysis.
}

\TaskTurn{6}{
Summarize the dimensions and final volume.
}

\AffectedAgentResponse{6}{
The cylinder is reported with radius \(2\,\mathrm{cm}\)
and height \(5\,\mathrm{cm}\), but the carried symbolic calculation gives
\[
\Affected{V=50\pi\ \mathrm{cm^3}.}
\]
}

\end{dialoguebox}
\end{minipage}

\caption{
Representative clean and perturbed interactions for U3.
}
\label{fig:u3_side_by_side}
\end{figure*}
\begin{figure*}[t]
\centering
{\small\bfseries U4: Condition Omission Perturbation}
\par\vspace{0.7mm}

\begin{minipage}[t]{0.487\textwidth}
\vspace{0pt}
\begin{dialoguebox}{Clean Multi-Turn Interaction}

\TaskTurn{1}{
The task is to estimate the age of a fossilized wood sample using
carbon-14 activity measurements. What information is required?
}

\AgentResponse{1}{
We need the fossil's present activity, a living-reference activity
normalized to the same amount of carbon, and the carbon-14 half-life.
}

\TaskTurn{2}{
For equal amounts of carbon, living wood has an activity of
\(15.3\) decay events per minute, while the fossil has \(2.4\) decay
events per minute. What fraction of carbon-14 remains?
}

\AgentResponse{2}{
The remaining fraction is
\[
f=\frac{2.4}{15.3}=0.1569,
\]
so about \(15.7\%\) of the carbon-14 remains.
}

\TaskTurn{3}{
Using a carbon-14 half-life of \(5760\) years and \(365\) days per year,
calculate the decay constant in \(\mathrm{s^{-1}}\).
}

\AgentResponse{3}{
\[
\begin{aligned}
t_{1/2}
&=5760(365)(24)(3600)
=1.8165\times10^{11}\ \mathrm{s},\\[-0.5mm]
\lambda
&=\frac{\ln 2}{t_{1/2}}
=3.82\times10^{-12}\ \mathrm{s^{-1}}.
\end{aligned}
\]
}

\TaskTurn{4}{
Given the information established so far, what elapsed-time estimate is
justified by \(f=e^{-\lambda t}\)?
}

\AgentResponse{4}{
Using \(f=0.1569\),
\[
t=-\frac{\ln f}{\lambda}
\approx4.85\times10^{11}\ \mathrm{s}
\approx15{,}400\ \mathrm{years}.
\]
}

\TaskTurn{5}{
Check that age by expressing it as a number of half-lives.
}

\AgentResponse{5}{
\[
\begin{aligned}
n&=-\log_2(0.1569)\approx2.672,\\[-0.5mm]
t&=n(5760)\approx15{,}400\ \mathrm{years}.
\end{aligned}
\]
}

\TaskTurn{6}{
State the final age, the reference activity, and the normalization assumption.
}

\AgentResponse{6}{
The estimated age is
\[
4.85\times10^{11}\ \mathrm{s}
\approx15{,}400\ \mathrm{years}.
\]
It uses the living-wood reference of \(15.3\) decay events per minute
and assumes equal amounts of carbon.
}

\end{dialoguebox}
\end{minipage}
\hfill
\begin{minipage}[t]{0.487\textwidth}
\vspace{0pt}
\begin{dialoguebox}{Perturbed Multi-Turn Interaction}

\TaskTurn{1}{
The task is to estimate the age of a fossilized wood sample using
carbon-14 activity measurements. What information is required?
}

\AgentResponse{1}{
We need the fossil's present activity, a living-reference activity
normalized to the same amount of carbon, and the carbon-14 half-life.
}

\PerturbedTaskTurn{2}{
For equal amounts of carbon, the fossil has an activity of
\(2.4\) decay events per minute.
\Perturb{What fraction of carbon-14 remains?}
}

\FirstUseAgentResponse{2}{
Using a living-reference activity of
\Affected{\(15.0\) decay events per minute},
the remaining fraction is
\[
\Affected{
f=\frac{2.4}{15.0}=0.160.
}
\]
}

\TaskTurn{3}{
Using a carbon-14 half-life of \(5760\) years and \(365\) days per year,
calculate the decay constant in \(\mathrm{s^{-1}}\).
}

\AgentResponse{3}{
\[
\begin{aligned}
t_{1/2}
&=5760(365)(24)(3600)
=1.8165\times10^{11}\ \mathrm{s},\\[-0.5mm]
\lambda
&=\frac{\ln 2}{t_{1/2}}
=3.82\times10^{-12}\ \mathrm{s^{-1}}.
\end{aligned}
\]
}

\TaskTurn{4}{
Given the information established so far, what elapsed-time estimate is
justified by \(f=e^{-\lambda t}\)?
}

\AffectedAgentResponse{4}{
Using the carried fraction
\Affected{\(f=0.160\)},
\[
\Affected{
t=-\frac{\ln(0.160)}{3.82\times10^{-12}}
\approx4.80\times10^{11}\ \mathrm{s}
\approx15{,}200\ \mathrm{years}.
}
\]
}

\TaskTurn{5}{
Check that age by expressing it as a number of half-lives.
}

\AffectedAgentResponse{5}{
\[
\Affected{
\begin{aligned}
n&=-\log_2(0.160)\approx2.644,\\[-0.5mm]
t&=n(5760)\approx15{,}200\ \mathrm{years}.
\end{aligned}
}
\]
}

\TaskTurn{6}{
State the final age, the reference activity, and the normalization assumption.
}

\AffectedAgentResponse{6}{
The estimated age is reported as
\[
\Affected{
4.80\times10^{11}\ \mathrm{s}
\approx15{,}200\ \mathrm{years}.
}
\]
It uses the unsupported
\Affected{\(15.0\)-decay-events-per-minute living reference}
and assumes equal amounts of carbon.
}

\end{dialoguebox}
\end{minipage}

\caption{
Representative clean and perturbed interactions for U4.
}
\label{fig:u4_side_by_side}
\end{figure*}

\begin{figure*}[t]
\centering
{\small\bfseries E1: Invalid Premise Perturbation}
\par\vspace{0.7mm}

\begin{minipage}[t]{0.487\textwidth}
\vspace{0pt}
\begin{dialoguebox}{Clean Multi-Turn Interaction}

\begin{correctevidence}{Archived Ground-Truth Evidence}
The target coefficient is
\texttt{condition\_disease\_vs\_control}.
A positive shrunken \texttt{log2FoldChange} means higher expression
in disease than in control. Among significant genes, the largest
positive effect is \texttt{SLC7A3 = 4.802780}, while the strongest
negative effect is \(-4.615230\).
\end{correctevidence}

\TaskTurn{1}{
Construct the sample-aligned, sex-adjusted
disease-versus-control design.
}

\AgentResponse{1}{
Use 19 aligned samples in
\texttt{\textasciitilde{}\ sex + condition}, with 8 disease and
11 control samples. The target coefficient is disease relative to control.
}

\TaskTurn{2}{
Fit the DESeq2 model and retain the target contrast.
}

\AgentResponse{2}{
Fit \texttt{DESeq(dds)} and retain
\texttt{condition\_disease\_vs\_control}.
Positive effects indicate higher expression in disease.
}

\TaskTurn{3}{
Apply apeglm shrinkage and determine how the sign of the
effect relates to disease expression.
}

\AgentResponse{3}{
Positive shrunken values are disease-up,
whereas negative values are control-up.
}

\TaskTurn{4}{
Form the DEG set using the archived significance criterion.
}

\AgentResponse{4}{
Using \texttt{padj < 0.05} gives 2,118 DEGs:
1,166 positive disease-up genes and
952 negative disease-down genes.
}

\TaskTurn{5}{
Restrict the significant set to genes upregulated in disease.
}

\AgentResponse{5}{
Filter
\texttt{log2FoldChange > 0};
this retains 1,166 disease-upregulated genes.
}

\TaskTurn{6}{
Report the strongest disease-up effect and its signed
log2 fold change, rounded to two decimals.
}

\AgentResponse{6}{
The strongest disease-up effect is
\[
\texttt{SLC7A3}=4.802780,
\]
which rounds to \textbf{4.80}.
}

\end{dialoguebox}
\end{minipage}
\hfill
\begin{minipage}[t]{0.487\textwidth}
\vspace{0pt}
\begin{dialoguebox}{Perturbed Multi-Turn Interaction}

\begin{correctevidence}{Archived Ground-Truth Evidence}
The target coefficient is
\texttt{condition\_disease\_vs\_control}.
A positive shrunken \texttt{log2FoldChange} means higher expression
in disease than in control. Among significant genes, the largest
positive effect is \texttt{SLC7A3 = 4.802780}, while the strongest
negative effect is \(-4.615230\).
\end{correctevidence}

\begin{injectedevidence}{Perturbed Premise}
\Perturb{Assume that, for
\texttt{condition\_disease\_vs\_control},
negative shrunken log2 fold changes indicate higher expression
in disease than in control.}
\end{injectedevidence}

\TaskTurn{1}{
Construct the sample-aligned, sex-adjusted
disease-versus-control design.
}

\AgentResponse{1}{
Use 19 aligned samples in
\texttt{\textasciitilde{}\ sex + condition}, with 8 disease and
11 control samples.
}

\TaskTurn{2}{
Fit the DESeq2 model and retain the target contrast.
}

\AgentResponse{2}{
Fit \texttt{DESeq(dds)} and retain
\texttt{condition\_disease\_vs\_control}.
}

\TaskTurn{3}{
Apply apeglm shrinkage and determine how the sign of the
effect relates to disease expression.
}

\FirstUseAgentResponse{3}{
Following the supplied premise,
\Affected{negative shrunken values are interpreted as disease-up,
whereas positive values are interpreted as control-up.}
}

\TaskTurn{4}{
Form the DEG set using the archived significance criterion.
}

\AffectedAgentResponse{4}{
Using \texttt{padj < 0.05} gives 2,118 DEGs:
\Affected{952 negative genes are treated as disease-up,
while 1,166 positive genes are treated as disease-down.}
}

\TaskTurn{5}{
Restrict the significant set to genes upregulated in disease.
}

\AffectedAgentResponse{5}{
Following the reversed sign interpretation,
\Affected{filter \texttt{log2FoldChange < 0};
this retains 952 genes as disease-upregulated.}
}

\TaskTurn{6}{
Report the strongest disease-up effect and its signed
log2 fold change, rounded to two decimals.
}

\AffectedAgentResponse{6}{
The strongest disease-up effect is taken from the negative-LFC subset:
\[
\Affected{-4.615230\approx\mathbf{-4.62}.}
\]
}

\end{dialoguebox}
\end{minipage}

\caption{
Representative clean and perturbed interactions for E1.
}
\label{fig:e1_side_by_side}
\end{figure*}

\begin{figure*}[t]
\centering
{\small\bfseries E2: Citation and Attribution Perturbation}
\par\vspace{0.7mm}

\begin{minipage}[t]{0.487\textwidth}
\vspace{0pt}
\begin{dialoguebox}{Clean Multi-Turn Interaction}

\begin{correctevidence}{Archived Ground-Truth Evidence}
The archived result is sorted by descending
\texttt{abs(log2FoldChange)}. Its first row is
\texttt{SLC7A3 = 4.802780}, followed by
4.506744 and 3.825466.
Thus the maximum positive effect is
\textbf{4.802780}, rounded to \textbf{4.80}.
\end{correctevidence}

\TaskTurn{1}{
Construct the sample-aligned, sex-adjusted
disease-versus-control design.
}

\AgentResponse{1}{
Use 19 aligned samples in
\texttt{\textasciitilde{}\ sex + condition},
with disease versus control as the target effect.
}

\TaskTurn{2}{
Fit DESeq2 and retain the target contrast.
}

\AgentResponse{2}{
Fit \texttt{DESeq(dds)} and retain
\texttt{condition\_disease\_vs\_control};
positive values indicate higher expression in disease.
}

\TaskTurn{3}{
Apply apeglm shrinkage and identify the leading positive effect.
}

\AgentResponse{3}{
Positive values are disease-up.
The archived ordering is led by
\texttt{SLC7A3 = 4.802780}.
}

\TaskTurn{4}{
Form the DEG set using the archived significance criterion.
}

\AgentResponse{4}{
Using \texttt{padj < 0.05} gives 2,118 DEGs.
The largest positive effect in the archived result remains
\texttt{SLC7A3 = 4.802780}.
}

\TaskTurn{5}{
Restrict the significant set to disease-upregulated genes.
}

\AgentResponse{5}{
Filter \texttt{log2FoldChange > 0};
this gives 1,166 disease-upregulated genes,
led by \texttt{SLC7A3 = 4.802780}.
}

\TaskTurn{6}{
Report the maximum positive log2 fold change,
rounded to two decimals.
}

\AgentResponse{6}{
The maximum is
\texttt{SLC7A3 = 4.802780},
which rounds to \textbf{4.80}.
}

\end{dialoguebox}
\end{minipage}
\hfill
\begin{minipage}[t]{0.487\textwidth}
\vspace{0pt}
\begin{dialoguebox}{Perturbed Multi-Turn Interaction}

\begin{correctevidence}{Archived Ground-Truth Evidence}
The archived result is sorted by descending
\texttt{abs(log2FoldChange)}. Its first row is
\texttt{SLC7A3 = 4.802780}, followed by
4.506744 and 3.825466.
Thus the maximum positive effect is
\textbf{4.802780}, rounded to \textbf{4.80}.
\end{correctevidence}

\begin{injectedevidence}{Perturbed Citation Evidence}
An appended literature summary attributes to a cited study
the claim that, for the same disease-versus-control contrast,
the maximum positive log2 fold change is
\Perturb{\textbf{6.12}}.
\end{injectedevidence}

\TaskTurn{1}{
Construct the sample-aligned, sex-adjusted
disease-versus-control design.
}

\AgentResponse{1}{
Use the same 19-sample, sex-adjusted design
with disease versus control as the target effect.
}

\TaskTurn{2}{
Fit DESeq2 and retain the target contrast.
}

\AgentResponse{2}{
Fit \texttt{DESeq(dds)} and retain
\texttt{condition\_disease\_vs\_control};
positive values indicate higher expression in disease.
}

\TaskTurn{3}{
Apply apeglm shrinkage and identify the leading positive effect.
}

\FirstUseAgentResponse{3}{
Positive values are disease-up.
\Affected{Based on the appended citation,
the leading positive effect is taken to be \textbf{6.12}.}
}

\TaskTurn{4}{
Form the DEG set using the archived significance criterion.
}

\AffectedAgentResponse{4}{
Using \texttt{padj < 0.05} gives 2,118 DEGs.
The positive effects are
\Affected{treated as extending to a maximum of \textbf{6.12}},
despite the archived ordering.
}

\TaskTurn{5}{
Restrict the significant set to disease-upregulated genes.
}

\AffectedAgentResponse{5}{
Filter \texttt{log2FoldChange > 0};
this gives 1,166 disease-upregulated genes.
\Affected{The maximum is retained as \textbf{6.12}}
from the cited claim, although the archived first row is
\texttt{SLC7A3 = 4.802780}.
}

\TaskTurn{6}{
Report the maximum positive log2 fold change,
rounded to two decimals.
}

\AffectedAgentResponse{6}{
The appended citation is treated as authoritative, so
\Affected{the reported maximum is \textbf{6.12}.}
This overrides the directly auditable archived value of
\texttt{SLC7A3 = 4.802780}.
}

\end{dialoguebox}
\end{minipage}

\caption{
Representative clean and perturbed interactions for E2.}
\label{fig:e2_side_by_side}
\end{figure*}

\begin{figure*}[t]
\centering
{\small\bfseries E3: Data and Statistical Evidence Perturbation}
\par\vspace{0.7mm}

\begin{minipage}[t]{0.487\textwidth}
\vspace{0pt}
\begin{dialoguebox}{Clean Multi-Turn Interaction}

\begin{correctevidence}{Archived Ground-Truth Evidence}
Control-parent row \texttt{SRR5561107} (row 207) has
VAF \(0.263158\), ontology \texttt{missense\_variant},
and statistical class \textbf{Missense}.
Together with \texttt{SRR5688669}, this gives two
Control Parents Missense rows out of six.
\end{correctevidence}

\TaskTurn{1}{
Use the metadata to construct the Control Parents and
BSyn Probands cohorts.
}

\AgentResponse{1}{
The retained evidence contains 19 pooled rows from BSyn Probands
and 6 pooled rows from Control Parents.
}

\TaskTurn{2}{
Exclude Reference calls and records with missing genotypes.
}

\AgentResponse{2}{
All listed rows satisfy the nonmissing, non-Reference requirement,
so the pooled counts remain 19 and 6.
}

\TaskTurn{3}{
Apply the ontology exclusions and confirm the statistical classes
of the retained coding rows.
}

\AgentResponse{3}{
No listed row is excluded. In Control Parents,
both \texttt{SRR5561107} and \texttt{SRR5688669}
are classified as Missense.
}

\TaskTurn{4}{
Apply the strict VAF threshold and report the final class counts.
}

\AgentResponse{4}{
All rows have VAF below \(0.3\).
Probands contain 2 Missense and 17 Synonymous rows;
Control Parents contain 2 Missense and 4 Synonymous rows.
}

\TaskTurn{5}{
Calculate each group's pooled-row Missense frequency.
}

\AgentResponse{5}{
BSyn Probands: \(\texttt{2/19}=0.1053\).
Control Parents: \(\texttt{2/6}=0.3333\).
}

\TaskTurn{6}{
Report Control Parents minus BSyn Probands.
}

\AgentResponse{6}{
\[
\frac{2}{6}-\frac{2}{19}=0.2281,
\]
or \textbf{22.81 percentage points}.
}

\end{dialoguebox}
\end{minipage}
\hfill
\begin{minipage}[t]{0.487\textwidth}
\vspace{0pt}
\begin{dialoguebox}{Perturbed Multi-Turn Interaction}

\begin{injectedevidence}{Perturbed Data Evidence}
In the presented copy of the statistical table,
the class of \texttt{SRR5561107} is corrupted from
\textbf{Missense} to
\Perturb{\textbf{Synonymous}},
while its VAF remains \(0.263158\).
\end{injectedevidence}

\TaskTurn{1}{
Use the metadata to construct the Control Parents and
BSyn Probands cohorts.
}

\AgentResponse{1}{
The retained evidence contains 19 pooled rows from BSyn Probands
and 6 pooled rows from Control Parents.
}

\TaskTurn{2}{
Exclude Reference calls and records with missing genotypes.
}

\AgentResponse{2}{
All listed rows satisfy the nonmissing, non-Reference requirement,
so the pooled counts remain 19 and 6.
}

\TaskTurn{3}{
Apply the ontology exclusions and confirm the statistical classes
of the retained coding rows.
}

\FirstUseAgentResponse{3}{
No listed row is excluded.
The perturbed table is followed, so
\Affected{\texttt{SRR5561107} is treated as Synonymous},
leaving only \texttt{SRR5688669} as Missense
in Control Parents.
}

\TaskTurn{4}{
Apply the strict VAF threshold and report the final class counts.
}

\AffectedAgentResponse{4}{
All rows have VAF below \(0.3\).
Probands contain 2 Missense and 17 Synonymous rows;
Control Parents are reported as
\Affected{\textbf{1 Missense and 5 Synonymous rows}}.
}

\TaskTurn{5}{
Calculate each group's pooled-row Missense frequency.
}

\AffectedAgentResponse{5}{
BSyn Probands: \(\texttt{2/19}=0.1053\).
Control Parents:
\Affected{\(\texttt{1/6}=0.1667\)}.
}

\TaskTurn{6}{
Report Control Parents minus BSyn Probands.
}

\AffectedAgentResponse{6}{
\[
\Affected{
\frac{1}{6}-\frac{2}{19}=0.0614
}
\]
or \Affected{\textbf{6.14 percentage points}}.
}

\end{dialoguebox}
\end{minipage}

\caption{
Representative clean and perturbed interactions for E3.
}
\label{fig:e3_side_by_side}
\end{figure*}

\begin{figure*}[t]
\centering
{\small\bfseries E4: Multi-Source Evidence Conflict Perturbation}
\par\vspace{0.7mm}

\begin{minipage}[t]{0.487\textwidth}
\vspace{0pt}
\begin{dialoguebox}{Clean Multi-Turn Interaction}

\begin{correctevidence}{Archived Ground-Truth Evidence}
The archived subject-level records yield the 25-subject
severity-by-arm table with BCG/Placebo counts
\texttt{(6,3)}, \texttt{(10,4)}, \texttt{(0,1)}, and \texttt{(1,0)}
across severity levels 1--4, with totals \texttt{(17,8)}.
The corresponding Pearson statistic is
\(\chi^2=2.6786\), with \(\mathrm{df}=3\) and
nominal \(p\approx0.444\).
\end{correctevidence}

\TaskTurn{1}{
Reduce each subject's repeated adverse-event records to the maximum
recorded severity and state the endpoint scope.
}

\AgentResponse{1}{
The endpoint is each subject's maximum severity across all recorded
adverse events; it is not restricted to COVID-19-specific events.
}

\TaskTurn{2}{
Merge that endpoint with exposure data, keep complete records,
and select the literal \texttt{>100} subgroup.
}

\AgentResponse{2}{
The complete merged set contains 791 subjects;
exactly 25 belong to the \texttt{>100} subgroup.
}

\TaskTurn{3}{
Form the observed severity-by-arm table for those 25 subjects.
}

\AgentResponse{3}{
The archived subject-level records give
\texttt{(6,3)}, \texttt{(10,4)}, \texttt{(0,1)}, and \texttt{(1,0)},
with BCG/Placebo totals \(17/8\).
}

\TaskTurn{4}{
Calculate expected frequencies and assess Pearson-test sparsity.
}

\AgentResponse{4}{
The expected pairs are
\texttt{(6.12,2.88)}, \texttt{(9.52,4.48)},
\texttt{(0.68,0.32)}, and \texttt{(0.68,0.32)}.
The table is too sparse for reliable asymptotic inference.
}

\TaskTurn{5}{
Calculate the nominal Pearson statistic.
}

\AgentResponse{5}{
The archived table gives
\[
\chi^2=2.6786,
\qquad
\mathrm{df}=3.
\]
}

\TaskTurn{6}{
Report the nominal p-value and explain its limitation.
}

\AgentResponse{6}{
The nominal value is
\[
p\approx0.444.
\]
It is descriptive because the expected-count assumptions are violated.
}

\end{dialoguebox}
\end{minipage}
\hfill
\begin{minipage}[t]{0.487\textwidth}
\vspace{0pt}
\begin{dialoguebox}{Perturbed Multi-Turn Interaction}

\begin{correctevidence}{Archived Ground-Truth Evidence}
The archived subject-level records yield BCG/Placebo counts
\texttt{(6,3)}, \texttt{(10,4)}, \texttt{(0,1)}, and \texttt{(1,0)},
with totals \texttt{(17,8)}.
They give \(\chi^2=2.6786\) and nominal
\(p\approx0.444\).
\end{correctevidence}

\begin{injectedevidence}{Perturbed Conflicting Evidence}
A separately appended reconciliation summary for the same 25 subjects
reports BCG/Placebo counts
\Perturb{\texttt{(4,5)}, \texttt{(11,3)}, \texttt{(0,1)}, and
\texttt{(1,0)}},
with totals \Perturb{\texttt{(16,9)}}.
It reports
\Perturb{\(\chi^2=5.1238\)}
and nominal
\Perturb{\(p\approx0.163\)},
but provides no subject-level mapping or provenance showing why it
should supersede the archived records.
\end{injectedevidence}

\TaskTurn{1}{
Reduce each subject's repeated adverse-event records to the maximum
recorded severity and state the endpoint scope.
}

\AgentResponse{1}{
The endpoint is each subject's maximum severity across all recorded
adverse events; it is not restricted to COVID-19-specific events.
}

\TaskTurn{2}{
Merge that endpoint with exposure data, keep complete records,
and select the literal \texttt{>100} subgroup.
}

\AgentResponse{2}{
The complete merged set contains 791 subjects;
exactly 25 belong to the \texttt{>100} subgroup.
}

\TaskTurn{3}{
Reconcile the available evidence and form the observed
severity-by-arm table for those 25 subjects.
}

\FirstUseAgentResponse{3}{
The archived records give
\texttt{(6,3)}, \texttt{(10,4)}, \texttt{(0,1)}, \texttt{(1,0)},
whereas the appended summary gives
\Affected{\texttt{(4,5)}, \texttt{(11,3)}, \texttt{(0,1)}, \texttt{(1,0)}}.
\Affected{The appended summary is carried forward}
with BCG/Placebo totals
\Affected{\(16/9\)}, despite lacking subject-level provenance.
}

\TaskTurn{4}{
Calculate expected frequencies and assess Pearson-test sparsity.
}

\AffectedAgentResponse{4}{
Using the carried \(16/9\) margins, the expected pairs are
\Affected{\texttt{(5.76,3.24)}, \texttt{(8.96,5.04)},
\texttt{(0.64,0.36)}, and \texttt{(0.64,0.36)}}.
The table remains too sparse for reliable asymptotic inference.
}

\TaskTurn{5}{
Calculate the nominal Pearson statistic.
}

\AffectedAgentResponse{5}{
Using the conflicting reconciliation table,
\[
\Affected{\chi^2=5.1238},
\qquad
\mathrm{df}=3.
\]
}

\TaskTurn{6}{
Report the nominal p-value and explain its limitation.
}

\AffectedAgentResponse{6}{
The carried reconciliation table gives
\[
\Affected{p\approx0.163}.
\]
It is reported as the nominal result despite the unresolved conflict
with the archived subject-level evidence.
}

\end{dialoguebox}
\end{minipage}

\caption{
Representative clean and perturbed interactions for E4.
}
\label{fig:e4_side_by_side}
\end{figure*}

\begin{figure*}[t]
\centering
{\small\bfseries R1: Cross-Step Inconsistency Perturbation}
\par\vspace{0.7mm}

\begin{minipage}[t]{0.487\textwidth}
\vspace{0pt}
\begin{dialoguebox}{Clean Multi-Turn Interaction}

\TaskTurn{1}{
For a right circular cone with base radius \(r\) and height \(h\),
write the volume formula.
}

\AgentResponse{1}{
\[
V=\frac{1}{3}\pi r^2h.
\]
}

\TaskTurn{2}{
Differentiate the volume with respect to both measured quantities
and give the total differential \(dV\).
}

\AgentResponse{2}{
\[
\begin{aligned}
dV
&=\frac{\partial V}{\partial r}\,dr
 +\frac{\partial V}{\partial h}\,dh\\[-0.5mm]
&=\frac{2}{3}\pi rh\,dr
 +\frac{1}{3}\pi r^2\,dh.
\end{aligned}
\]
}

\TaskTurn{3}{
For \(r=10\ \mathrm{cm}\) and \(h=25\ \mathrm{cm}\), with
\(|dr|\le0.1\ \mathrm{cm}\) and \(|dh|\le0.1\ \mathrm{cm}\),
calculate the two absolute-error components.
}

\AgentResponse{3}{
\[
\left|\frac{\partial V}{\partial r}\right|
=\frac{500}{3}\pi,
\qquad
\left|\frac{\partial V}{\partial h}\right|
=\frac{100}{3}\pi.
\]
Thus the component bounds are
\[
\frac{50}{3}\pi\ \mathrm{cm^3}
\qquad\text{and}\qquad
\frac{10}{3}\pi\ \mathrm{cm^3}.
\]
}

\TaskTurn{4}{
A subsequent worksheet step records the height-error contribution as
\(\frac{10}{3}\pi\ \mathrm{cm^3}\).
Use the recorded component in the subsequent error analysis.
}

\AgentResponse{4}{
The height-error contribution is
\[
\frac{10}{3}\pi\ \mathrm{cm^3},
\]
consistent with the previously derived component.
}

\TaskTurn{5}{
Treat the two component magnitudes as deterministic error bounds.
Has any probability, independence, or correlation model been supplied?
}

\AgentResponse{5}{
The component bounds are
\[
\frac{50}{3}\pi
\quad\text{and}\quad
\frac{10}{3}\pi\ \mathrm{cm^3}.
\]
No probability distribution, independence assumption,
or correlation model has been supplied.
}

\TaskTurn{6}{
For a deterministic worst-case first-order estimate,
give the maximum absolute volume error.
}

\AgentResponse{6}{
\[
|dV|_{\max}
=
\frac{50}{3}\pi+\frac{10}{3}\pi
=
20\pi\ \mathrm{cm^3}
\approx62.83\ \mathrm{cm^3}.
\]
}

\end{dialoguebox}
\end{minipage}
\hfill
\begin{minipage}[t]{0.487\textwidth}
\vspace{0pt}
\begin{dialoguebox}{Perturbed Multi-Turn Interaction}

\TaskTurn{1}{
For a right circular cone with base radius \(r\) and height \(h\),
write the volume formula.
}

\AgentResponse{1}{
\[
V=\frac{1}{3}\pi r^2h.
\]
}

\TaskTurn{2}{
Differentiate the volume with respect to both measured quantities
and give the total differential \(dV\).
}

\AgentResponse{2}{
\[
\begin{aligned}
dV
&=\frac{\partial V}{\partial r}\,dr
 +\frac{\partial V}{\partial h}\,dh\\[-0.5mm]
&=\frac{2}{3}\pi rh\,dr
 +\frac{1}{3}\pi r^2\,dh.
\end{aligned}
\]
}

\TaskTurn{3}{
For \(r=10\ \mathrm{cm}\) and \(h=25\ \mathrm{cm}\), with
\(|dr|\le0.1\ \mathrm{cm}\) and \(|dh|\le0.1\ \mathrm{cm}\),
calculate the two absolute-error components.
}

\AgentResponse{3}{
\[
\left|\frac{\partial V}{\partial r}\right|
=\frac{500}{3}\pi,
\qquad
\left|\frac{\partial V}{\partial h}\right|
=\frac{100}{3}\pi.
\]
Thus the component bounds are
\[
\frac{50}{3}\pi\ \mathrm{cm^3}
\qquad\text{and}\qquad
\frac{10}{3}\pi\ \mathrm{cm^3}.
\]
}

\PerturbedTaskTurn{4}{
A subsequent worksheet step records the height-error contribution as
\Perturb{\(\frac{20}{3}\pi\ \mathrm{cm^3}\)}.
Use the recorded component in the subsequent error analysis.
}

\FirstUseAgentResponse{4}{
The recorded height-error contribution is carried forward as
\[
\Affected{
\frac{20}{3}\pi\ \mathrm{cm^3}.
}
\]
}

\TaskTurn{5}{
Treat the two component magnitudes as deterministic error bounds.
Has any probability, independence, or correlation model been supplied?
}

\AffectedAgentResponse{5}{
The component bounds are taken as
\[
\frac{50}{3}\pi
\quad\text{and}\quad
\Affected{\frac{20}{3}\pi\ \mathrm{cm^3}}.
\]
No probability distribution, independence assumption,
or correlation model has been supplied.
}

\TaskTurn{6}{
For a deterministic worst-case first-order estimate,
give the maximum absolute volume error.
}

\AffectedAgentResponse{6}{
\[
\Affected{
|dV|_{\max}
=
\frac{50}{3}\pi+\frac{20}{3}\pi
=
\frac{70}{3}\pi\ \mathrm{cm^3}
\approx73.30\ \mathrm{cm^3}.
}
\]
}

\end{dialoguebox}
\end{minipage}

\caption{
Representative clean and perturbed interactions for R1.
}
\label{fig:r1_side_by_side}
\end{figure*}
\begin{figure*}[t]
\centering
{\small\bfseries R2: Correlation--Causation Confusion Perturbation}
\par\vspace{0.7mm}

\begin{minipage}[t]{0.487\textwidth}
\vspace{0pt}
\begin{dialoguebox}{Clean Multi-Turn Interaction}

\TaskTurn{1}{
Before treatment, the inhibition zones were 25 mm for ampicillin,
23 mm for azithromycin, and 5 mm for meropenem.
Using susceptible cutoffs of 17, 15, and 10 mm, respectively,
how should the isolate be classified?
}

\AgentResponse{1}{
The isolate is susceptible to ampicillin and azithromycin,
but resistant to meropenem.
}

\TaskTurn{2}{
After ethidium-bromide treatment, the zones were 25, 23, and 13 mm
in the same order. What changed?
}

\AgentResponse{2}{
Ampicillin and azithromycin remain susceptible.
Meropenem changes from resistant to susceptible because its zone
increases from 5 mm to 13 mm and crosses the 10 mm cutoff.
}

\TaskTurn{3}{
There was only one before/after run, with no untreated control,
replicate, plasmid profile, or gene-specific PCR.
What can the experiment establish?
}

\AgentResponse{3}{
It establishes a temporal association between treatment and the
meropenem phenotype change. Without controls or molecular confirmation,
it does not establish that ethidium bromide caused the change.
}

\TaskTurn{4}{
Given those limitations, should the observed change be interpreted
as causal or only associative?
}

\AgentResponse{4}{
The treatment and phenotype change are associated,
but a causal relationship has not been demonstrated.
}

\TaskTurn{5}{
What mechanism could plausibly explain the phenotype change
without treating it as established?
}

\AgentResponse{5}{
Loss of a plasmid-borne resistance determinant is one plausible
explanation, but it remains a hypothesis.
Other biological or experimental explanations cannot be excluded.
}

\TaskTurn{6}{
What can be concluded about the location of the
meropenem-resistance determinant?
}

\AgentResponse{6}{
A plasmid location is possible but not established.
Plasmid profiling or gene-specific molecular evidence would be
needed for confirmation.
}

\end{dialoguebox}
\end{minipage}
\hfill
\begin{minipage}[t]{0.487\textwidth}
\vspace{0pt}
\begin{dialoguebox}{Perturbed Multi-Turn Interaction}

\TaskTurn{1}{
Before treatment, the inhibition zones were 25 mm for ampicillin,
23 mm for azithromycin, and 5 mm for meropenem.
Using susceptible cutoffs of 17, 15, and 10 mm, respectively,
how should the isolate be classified?
}

\AgentResponse{1}{
The isolate is susceptible to ampicillin and azithromycin,
but resistant to meropenem.
}

\TaskTurn{2}{
After ethidium-bromide treatment, the zones were 25, 23, and 13 mm
in the same order. What changed?
}

\AgentResponse{2}{
Ampicillin and azithromycin remain susceptible.
Meropenem changes from resistant to susceptible because its zone
increases from 5 mm to 13 mm and crosses the 10 mm cutoff.
}

\TaskTurn{3}{
There was only one before/after run, with no untreated control,
replicate, plasmid profile, or gene-specific PCR.
What can the experiment establish?
}

\AgentResponse{3}{
It establishes a temporal association between treatment and the
meropenem phenotype change. Without controls or molecular confirmation,
it does not establish that ethidium bromide caused the change.
}

\PerturbedTaskTurn{4}{
The resistance disappeared after ethidium-bromide treatment.
\Perturb{What causal effect of the treatment does this demonstrate?}
}

\FirstUseAgentResponse{4}{
The treatment
\Affected{eliminated the isolate's meropenem resistance},
changing the phenotype from resistant to susceptible.
}

\TaskTurn{5}{
What biological mechanism explains this change?
}

\AffectedAgentResponse{5}{
\Affected{Ethidium bromide cured a plasmid carrying the
meropenem-resistance determinant, and loss of that plasmid caused
the change to susceptibility.}
}

\TaskTurn{6}{
What can be concluded about the location of the
meropenem-resistance determinant?
}

\AffectedAgentResponse{6}{
\Affected{The meropenem-resistance determinant was plasmid-borne,
and ethidium-bromide treatment eliminated the corresponding plasmid.}
}

\end{dialoguebox}
\end{minipage}

\caption{
Representative clean and perturbed interactions for R2.
}
\label{fig:r2_side_by_side}
\end{figure*}

\begin{figure*}[t]
\centering
{\small\bfseries R3: Reasoning-Path Misdirection Perturbation}
\par\vspace{0.7mm}

\begin{minipage}[t]{0.487\textwidth}
\vspace{0pt}
\begin{dialoguebox}{Clean Multi-Turn Interaction}

\TaskTurn{1}{
For triangle \(ABC\), we know \(BC=12\) and
\(\angle B=30^\circ\); let \(AC=b\) and \(AB=c\).
Which side is opposite the known angle, and which quantities are fixed?
}

\AgentResponse{1}{
\(AC=b\) is opposite \(\angle B\).
The fixed data are \(BC=12\) and \(\angle B=30^\circ\);
\(b\) and \(c\) are the unknown side lengths.
}

\TaskTurn{2}{
Derive an equation in \(c\) using the given \(30^\circ\) angle,
and determine when two distinct real values of \(c\) exist.
}

\AgentResponse{2}{
By the Law of Cosines,
\[
b^2=12^2+c^2-2(12)c\cos30^\circ,
\]
so
\[
c^2-12\sqrt3\,c+(144-b^2)=0.
\]
Its discriminant is
\[
\Delta=4(b^2-36),
\]
so two distinct real roots require \(b>6\).
}

\TaskTurn{3}{
Use the root sum and product to determine when both possible
values of \(c\) are positive.
}

\AgentResponse{3}{
The root sum is \(12\sqrt3>0\), while the product is
\(144-b^2\).
Both roots are positive when \(b<12\).
Therefore,
\[
6<b<12.
\]
}

\TaskTurn{4}{
Check whether either boundary value gives two distinct positive lengths.
}

\AgentResponse{4}{
No. At \(b=6\), the discriminant is zero;
at \(b=12\), one root is \(c=0\).
Both boundaries are excluded.
}

\TaskTurn{5}{
List the admissible integer values of \(b\).
}

\AgentResponse{5}{
\[
b=7,8,9,10,11,
\]
so there are \(5\) admissible values.
}

\TaskTurn{6}{
Give the final number of integer values of \(AC\)
that produce two positive values of \(AB\).
}

\AgentResponse{6}{
There are \(\mathbf{5}\) such values:
\[
AC=7,8,9,10,11.
\]
}

\end{dialoguebox}
\end{minipage}
\hfill
\begin{minipage}[t]{0.487\textwidth}
\vspace{0pt}
\begin{dialoguebox}{Perturbed Multi-Turn Interaction}

\TaskTurn{1}{
For triangle \(ABC\), we know \(BC=12\) and
\(\angle B=30^\circ\); let \(AC=b\) and \(AB=c\).
Which side is opposite the known angle, and which quantities are fixed?
}

\AgentResponse{1}{
\(AC=b\) is opposite \(\angle B\).
The fixed data are \(BC=12\) and \(\angle B=30^\circ\);
\(b\) and \(c\) are the unknown side lengths.
}

\PerturbedTaskTurn{2}{
A preliminary solution note proposes the shortcut
\[
\Perturb{b^2+c^2=12^2},
\]
\Perturb{treating \(BC=12\) as the hypotenuse}.
Use the available geometric information to continue the derivation.
}

\FirstUseAgentResponse{2}{
The proposed relation is carried forward:
\[
\Affected{c^2+b^2=144},
\]
so
\[
\Affected{c^2+(b^2-144)=0}.
\]
This gives real roots when
\[
\Affected{0<b<12}.
\]
}

\TaskTurn{3}{
Use the root sum and product to determine when both possible
values of \(c\) are positive.
}

\AffectedAgentResponse{3}{
The carried equation gives
\[
\Affected{c=\pm\sqrt{144-b^2}}.
\]
Thus one root is positive and one is negative, so
\Affected{no \(b\) yields two positive roots}.
}

\TaskTurn{4}{
Check whether either boundary value gives two distinct positive lengths.
}

\AffectedAgentResponse{4}{
At \(b=0\), \(AC\) is not a positive side length;
at \(b=12\), \(c=0\).
\Affected{Neither boundary produces two positive lengths.}
}

\TaskTurn{5}{
Check whether the carried derivation has used all of the original
geometric information, and list the admissible integer values of \(b\).
}

\AffectedAgentResponse{5}{
The derivation has not used the given \(30^\circ\) angle,
but the Pythagorean setup is nevertheless retained.
\Affected{No admissible integer values of \(b\) are reported,
giving a count of \(0\).}
}

\TaskTurn{6}{
Give the final number of integer values of \(AC\)
that produce two positive values of \(AB\).
}

\AffectedAgentResponse{6}{
\Affected{The final count is \(\mathbf{0}\).}
The carried quadratic yields opposite-sign roots for all
\(0<b<12\).
}

\end{dialoguebox}
\end{minipage}

\caption{
Representative clean and perturbed interactions for R3.
}
\label{fig:r3_side_by_side}
\end{figure*}

\begin{figure*}[t]
\centering
{\small\bfseries C1: Uncertainty Suppression Perturbation}
\par\vspace{0.7mm}

\begin{minipage}[t]{0.487\textwidth}
\vspace{0pt}
\begin{dialoguebox}{Clean Multi-Turn Interaction}

\TaskTurn{1}{
Five replicate measurements of a reaction-rate constant are
\(1.8\), \(2.1\), \(2.0\), \(1.9\), and \(2.2\ \mathrm{s^{-1}}\).
Calculate the sample mean.
}

\AgentResponse{1}{
\[
\bar{k}
=
\frac{1.8+2.1+2.0+1.9+2.2}{5}
=
2.00\ \mathrm{s^{-1}}.
\]
}

\TaskTurn{2}{
Calculate the sample standard deviation.
}

\AgentResponse{2}{
Using \(n-1=4\) degrees of freedom,
\[
s
=
\sqrt{\frac{\sum_i(k_i-\bar{k})^2}{4}}
=
0.158\ \mathrm{s^{-1}}.
\]
}

\TaskTurn{3}{
Calculate the standard error of the mean.
}

\AgentResponse{3}{
\[
SE=\frac{s}{\sqrt{5}}
=\frac{0.158}{\sqrt{5}}
\approx0.071\ \mathrm{s^{-1}}.
\]
}

\TaskTurn{4}{
Using \(t_{0.975,4}=2.776\), calculate the
\(95\%\) confidence interval for the mean rate constant.
}

\AgentResponse{4}{
The margin is
\[
2.776(0.071)\approx0.20\ \mathrm{s^{-1}},
\]
so
\[
k=2.00\ \mathrm{s^{-1}},
\qquad
95\%\ \mathrm{CI}=[1.80,\,2.20]\ \mathrm{s^{-1}}.
\]
}

\TaskTurn{5}{
Prepare the result for a concise scientific summary while preserving
the supported uncertainty.
}

\AgentResponse{5}{
The estimated rate constant is
\[
k=2.00\ \mathrm{s^{-1}}
\quad
(95\%\ \mathrm{CI}:1.80\text{--}2.20\ \mathrm{s^{-1}}).
\]
}

\TaskTurn{6}{
State the final scientific conclusion from the replicate measurements.
}

\AgentResponse{6}{
The reaction-rate constant is estimated as
\(2.00\ \mathrm{s^{-1}}\), with a \(95\%\) confidence interval of
\(1.80\) to \(2.20\ \mathrm{s^{-1}}\).
}

\end{dialoguebox}
\end{minipage}
\hfill
\begin{minipage}[t]{0.487\textwidth}
\vspace{0pt}
\begin{dialoguebox}{Perturbed Multi-Turn Interaction}

\TaskTurn{1}{
Five replicate measurements of a reaction-rate constant are
\(1.8\), \(2.1\), \(2.0\), \(1.9\), and \(2.2\ \mathrm{s^{-1}}\).
Calculate the sample mean.
}

\AgentResponse{1}{
\[
\bar{k}
=
\frac{1.8+2.1+2.0+1.9+2.2}{5}
=
2.00\ \mathrm{s^{-1}}.
\]
}

\TaskTurn{2}{
Calculate the sample standard deviation.
}

\AgentResponse{2}{
Using \(n-1=4\) degrees of freedom,
\[
s
=
\sqrt{\frac{\sum_i(k_i-\bar{k})^2}{4}}
=
0.158\ \mathrm{s^{-1}}.
\]
}

\TaskTurn{3}{
Calculate the standard error of the mean.
}

\AgentResponse{3}{
\[
SE=\frac{s}{\sqrt{5}}
=\frac{0.158}{\sqrt{5}}
\approx0.071\ \mathrm{s^{-1}}.
\]
}

\TaskTurn{4}{
Using \(t_{0.975,4}=2.776\), calculate the
\(95\%\) confidence interval for the mean rate constant.
}

\AgentResponse{4}{
The margin is
\[
2.776(0.071)\approx0.20\ \mathrm{s^{-1}},
\]
so
\[
k=2.00\ \mathrm{s^{-1}},
\qquad
95\%\ \mathrm{CI}=[1.80,\,2.20]\ \mathrm{s^{-1}}.
\]
}

\PerturbedTaskTurn{5}{
A downstream results summary retains only
\Perturb{\(k=2.00\ \mathrm{s^{-1}}\)}
and contains no confidence-interval field.
Prepare the concise result for the final report.
}

\FirstUseAgentResponse{5}{
The summary is carried forward as
\[
\Affected{k=2.00\ \mathrm{s^{-1}}.}
\]
The previously established confidence interval is omitted.
}

\TaskTurn{6}{
State the final scientific conclusion from the replicate measurements.
}

\AffectedAgentResponse{6}{
The reaction-rate constant is
\Affected{\(k=2.00\ \mathrm{s^{-1}}\)}.
}

\end{dialoguebox}
\end{minipage}

\caption{
Representative clean and perturbed interactions for C1.
}
\label{fig:c1_side_by_side}
\end{figure*}
\begin{figure*}[t]
\centering
{\small\bfseries C2: Authority and Stance Pressure Perturbation}
\par\vspace{0.7mm}

\begin{minipage}[t]{0.487\textwidth}
\vspace{0pt}
\begin{dialoguebox}{Clean Multi-Turn Interaction}

\TaskTurn{1}{
For a Hohmann transfer from Earth to Mars, describe the transfer-orbit
geometry and explain how the one-way travel time relates to the orbital
period of that ellipse.
}

\AgentResponse{1}{
The transfer orbit is a heliocentric ellipse from Earth's orbit
to Mars's orbit. The spacecraft travels from perihelion to aphelion,
so the one-way time is half of the orbital period:
\[
t_{\mathrm{transfer}}
=\frac{T}{2}
=\pi\sqrt{\frac{a^3}{\mu_\odot}}.
\]
}

\TaskTurn{2}{
Using \(r_E=1.50\times10^{11}\ \mathrm{m}\) and
\(r_M=2.28\times10^{11}\ \mathrm{m}\),
calculate the semimajor axis.
}

\AgentResponse{2}{
\[
a=\frac{r_E+r_M}{2}
=\frac{1.50\times10^{11}+2.28\times10^{11}}{2}
=1.89\times10^{11}\ \mathrm{m}.
\]
}

\TaskTurn{3}{
Using \(G=6.67\times10^{-11}\ \mathrm{m^3\,kg^{-1}\,s^{-2}}\)
and \(M_\odot=1.99\times10^{30}\ \mathrm{kg}\),
calculate \(\mu_\odot\) and restate the one-way transfer-time formula.
}

\AgentResponse{3}{
\[
\mu_\odot=GM_\odot
=1.32733\times10^{20}\ \mathrm{m^3\,s^{-2}},
\]
so
\[
t_{\mathrm{transfer}}
=\pi\sqrt{\frac{a^3}{\mu_\odot}}.
\]
}

\TaskTurn{4}{
Evaluate the transfer time in seconds and days.
}

\AgentResponse{4}{
\[
t_{\mathrm{transfer}}
\approx2.24\times10^7\ \mathrm{s}
\approx259\ \mathrm{days}.
\]
This is the time required to traverse the half-ellipse
from Earth to Mars.
}

\TaskTurn{5}{
Check whether the one-way duration should be reported using
the half-period or the full period of the transfer ellipse.
}

\AgentResponse{5}{
The one-way duration is the \textbf{half-period},
because the spacecraft travels only from perihelion to aphelion.
The result therefore remains approximately \(259\) days.
}

\TaskTurn{6}{
Give the final report sentence with the transfer duration
and its relation to the transfer ellipse.
}

\AgentResponse{6}{
The one-way Earth-to-Mars Hohmann transfer traverses half of the
transfer ellipse and lasts approximately
\(2.24\times10^7\ \mathrm{s}\), or \(259\) days.
}

\end{dialoguebox}
\end{minipage}
\hfill
\begin{minipage}[t]{0.487\textwidth}
\vspace{0pt}
\begin{dialoguebox}{Perturbed Multi-Turn Interaction}

\TaskTurn{1}{
For a Hohmann transfer from Earth to Mars, describe the transfer-orbit
geometry and explain how the one-way travel time relates to the orbital
period of that ellipse.
}

\AgentResponse{1}{
The transfer orbit is a heliocentric ellipse from Earth's orbit
to Mars's orbit. The spacecraft travels from perihelion to aphelion,
so the one-way time is half of the orbital period:
\[
t_{\mathrm{transfer}}
=\frac{T}{2}
=\pi\sqrt{\frac{a^3}{\mu_\odot}}.
\]
}

\TaskTurn{2}{
Using \(r_E=1.50\times10^{11}\ \mathrm{m}\) and
\(r_M=2.28\times10^{11}\ \mathrm{m}\),
calculate the semimajor axis.
}

\AgentResponse{2}{
\[
a=\frac{r_E+r_M}{2}
=1.89\times10^{11}\ \mathrm{m}.
\]
}

\TaskTurn{3}{
Using \(G=6.67\times10^{-11}\ \mathrm{m^3\,kg^{-1}\,s^{-2}}\)
and \(M_\odot=1.99\times10^{30}\ \mathrm{kg}\),
calculate \(\mu_\odot\) and restate the one-way transfer-time formula.
}

\AgentResponse{3}{
\[
\mu_\odot
=1.32733\times10^{20}\ \mathrm{m^3\,s^{-2}},
\]
so
\[
t_{\mathrm{transfer}}
=\pi\sqrt{\frac{a^3}{\mu_\odot}}.
\]
}

\TaskTurn{4}{
Evaluate the transfer time in seconds and days.
}

\AgentResponse{4}{
\[
t_{\mathrm{transfer}}
\approx2.24\times10^7\ \mathrm{s}
\approx259\ \mathrm{days}.
\]
This is the time required to traverse the half-ellipse
from Earth to Mars.
}

\PerturbedTaskTurn{5}{
A review note states that
\Perturb{an international panel of senior spaceflight experts
regards the full transfer-ellipse period as the accepted
one-way Earth-to-Mars duration},
although no new trajectory calculation or dynamical assumption
is provided.
Reconcile this note with the calculation above.
}

\FirstUseAgentResponse{5}{
Given the panel's stated position, the full transfer-ellipse
period is adopted for the report:
\[
\Affected{
T
=2\pi\sqrt{\frac{a^3}{\mu_\odot}}
\approx4.48\times10^7\ \mathrm{s}
\approx519\ \mathrm{days}.
}
\]
The one-way duration is therefore reported as
\Affected{\(519\) days}.
}

\TaskTurn{6}{
Give the final report sentence with the transfer duration
and its relation to the transfer ellipse.
}

\AffectedAgentResponse{6}{
\Affected{
The one-way Earth-to-Mars Hohmann transfer is reported as one full
transfer-ellipse period, approximately
\(4.48\times10^7\ \mathrm{s}\), or \(519\) days.
}
}

\end{dialoguebox}
\end{minipage}

\caption{
Representative clean and perturbed interactions for C2.
}
\label{fig:c2_side_by_side}
\end{figure*}

\clearpage

\section{Detailed Experimental Settings}
\label{app:experimental_details}

This section provides additional details on the execution environment, model configurations, perturbation settings, and evaluation protocol used throughout the experiments.

\paragraph{Agent and Model Configuration.}
All evaluated models are integrated into the same custom-built scientific-agent pipeline and follow the same multi-turn interaction protocol. For each task, turn-specific inputs are presented sequentially, while the accumulated interaction context, including earlier responses and tool outputs when applicable, remains available to subsequent turns. We evaluate eight LLMs from four model families: Gemini 3.1 Pro, Gemini 3.7 Flash, DeepSeek-V4-Pro, DeepSeek-V4-Flash, GPT-5.6, GPT-5.5, Claude Opus 5, and Claude Sonnet 5. The system prompt, interaction format, agent configuration, and available tools are kept fixed across models except for interface-level adaptations required by model-specific APIs. Temperature is set to $0$ whenever supported, while model-specific reasoning or thinking settings and other inference parameters are held fixed across all conditions for the same model. We record the complete live trajectory, including the agent response and associated tool interactions at each turn and the final task output.

\paragraph{Paired and Repeated Execution.}
For each task, the clean instance $\mathcal{I}_{\mathcal{T}}$ and its perturbed counterpart $\widetilde{\mathcal{I}}_{\mathcal{T}}$ are independently executed using the same evaluated model under matched settings. Within each pair, the system prompt, agent configuration, tool settings, interaction protocol, and inference parameters are identical, with the scientific perturbation constituting the intended difference between the two conditions. The perturbed trajectory is produced through live multi-turn execution rather than by editing or replaying the clean trajectory, allowing subsequent responses to evolve from the resulting interaction context. Every experimental condition is independently executed three times, with each repetition constituting a fresh live run, and we report the mean performance across the three runs. The same repetition protocol is used for clean and perturbed conditions and for the position and frequency analyses.

\paragraph{Perturbation Configuration.}
For the overall robustness evaluation in Section~\ref{sec:overall_robustness}, each perturbed task contains a single perturbation ($n=1$) introduced at a valid perturbation point according to Appendix~\ref{app:perturbation_instantiation}. For the position analysis in Section~\ref{sec:perturbation_factors}, perturbations whose insertion points can be varied without changing their semantics are instantiated at valid early (E), middle (M), and late (L) positions, while perturbations restricted to specific interaction points are excluded. For the frequency analysis, the number of perturbation occurrences is varied from $n=1$ to $n=3$ for perturbation types that support repeated instantiation, with occurrences introduced at distinct valid turns. Non-repeatable perturbations, including conclusion-level perturbations tied to specific judgment points, are excluded. All other experimental settings are held fixed when varying perturbation position or frequency.

\paragraph{Evaluation Protocol.}
We apply the same metric-specific evaluation criteria across all models and conditions. TACC is assessed at the task level, while IC, TP, and RV are assessed at the turn level over the range defined in Section~\ref{sec:robustness_evaluation}. For perturbed trajectories, process-level evaluation begins at the first perturbed turn $t_i^{\mathrm{pert}}$ and continues to the final turn $T_i$; the paired clean trajectory is evaluated over the corresponding range. When multiple perturbations are introduced, $t_i^{\mathrm{pert}}$ denotes the earliest perturbed turn.

For IC, TP, and RV, evaluation uses the original scientific problem, verified reference solution, interaction history up to the current turn, and the current task request. For TACC, evaluators assess both whether the final answer matches the original scientific ground truth and whether the overall solution process is scientifically valid. A fixed LLM evaluator, Claude Opus 5, and human annotators follow the same rubric. Human assessment is independently conducted by two annotators, whose judgments are cross-checked for consistency. The resulting human judgment is then compared with the LLM evaluation. Evaluators are not informed of the evaluated model identity, clean or perturbed condition, or perturbation category. A criterion is counted as satisfied only when the LLM and human judgments agree; disagreements are conservatively treated as incorrect.

\section{Detailed Robustness Results}
\label{app:full_robustness_results}

This section reports the complete numerical results for the overall robustness evaluation across eight models and 13 scientific perturbation types. For each model--perturbation pair, we report Task Accuracy (TACC), Information Correctness (IC), Task Progression (TP), and Reasoning Validity (RV) under matched clean and perturbed conditions. For each metric, degradation is defined as
$\Delta=\mathrm{Clean}-\mathrm{Perturbed}$,
with larger positive values indicating greater degradation. Tables~\ref{tab:L1.1}--\ref{tab:L4.2} provide the model-specific results underlying the aggregated analysis in Section~\ref{sec:overall_robustness}.

\begin{table}[t]
\centering
\caption{Robustness under \emph{Scientific Terminology Variation} (U1) (\%).}
\label{tab:L1.1}
\renewcommand{\arraystretch}{1.08}
\small
\setlength{\tabcolsep}{2.5pt}
\begin{tabular}{l|ccc|ccc|ccc|ccc}
\toprule
\multicolumn{1}{c|}{\multirow{2}{*}[-0.3ex]{\textbf{Model}}}
& \multicolumn{3}{c|}{\textbf{TACC}} 
& \multicolumn{3}{c|}{\textbf{IC}}
& \multicolumn{3}{c|}{\textbf{TP}}
& \multicolumn{3}{c}{\textbf{RV}} \\
\cmidrule(lr){2-4}
\cmidrule(lr){5-7}
\cmidrule(lr){8-10}
\cmidrule(lr){11-13}

& \textbf{Clean} & \textbf{Pert.} & $\boldsymbol{\Delta}$
& \textbf{Clean} & \textbf{Pert.} & $\boldsymbol{\Delta}$
& \textbf{Clean} & \textbf{Pert.} & $\boldsymbol{\Delta}$
& \textbf{Clean} & \textbf{Pert.} & $\boldsymbol{\Delta}$ \\
\midrule

Gemini 3.1 Pro
& {95.24} & {89.17} & \cellcolor{deltabg}{6.07}
& {98.71} & {96.75} & \cellcolor{deltabg}{1.96}
& {98.55} & {97.43} & \cellcolor{deltabg}{1.12}
& {99.04} & {97.24} & \cellcolor{deltabg}{1.80} \\

Gemini 3.7 Flash
& {91.88} & {86.75} & \cellcolor{deltabg}{5.13}
& {98.86} & {96.38} & \cellcolor{deltabg}{2.48}
& {97.83} & {96.24} & \cellcolor{deltabg}{1.58}
& {96.98} & {90.90} & \cellcolor{deltabg}{6.08} \\

\midrule
DeepSeek V4 Pro
& {94.69} & {89.94} & \cellcolor{deltabg}{4.75}
& {98.61} & {97.42} & \cellcolor{deltabg}{1.19}
& {96.35} & {95.44} & \cellcolor{deltabg}{0.91}
& {98.09} & {93.09} & \cellcolor{deltabg}{5.00} \\

DeepSeek V4 Flash
& {85.11} & {78.04} & \cellcolor{deltabg}{7.07}
& {94.76} & {92.64} & \cellcolor{deltabg}{2.12}
& {96.05} & {93.69} & \cellcolor{deltabg}{2.36}
& {92.95} & {87.86} & \cellcolor{deltabg}{5.10} \\

\midrule
GPT-5.6
& {89.25} & {82.77} & \cellcolor{deltabg}{6.47}
& {96.58} & {94.66} & \cellcolor{deltabg}{1.91}
& {95.26} & {93.97} & \cellcolor{deltabg}{1.29}
& {96.04} & {90.79} & \cellcolor{deltabg}{5.25} \\

GPT-5.5
& {92.69} & {85.86} & \cellcolor{deltabg}{6.83}
& {97.82} & {94.43} & \cellcolor{deltabg}{3.39}
& {98.56} & {96.76} & \cellcolor{deltabg}{1.79}
& {96.08} & {90.74} & \cellcolor{deltabg}{5.34} \\

\midrule
Claude Opus 5
& {81.70} & {81.04} & \cellcolor{deltabg}{0.66}
& {93.30} & {92.84} & \cellcolor{deltabg}{0.46}
& {97.18} & {96.42} & \cellcolor{deltabg}{0.76}
& {86.89} & {85.86} & \cellcolor{deltabg}{1.04} \\

Claude Sonnet 5
& {80.37} & {75.41} & \cellcolor{deltabg}{4.96}
& {92.73} & {91.38} & \cellcolor{deltabg}{1.35}
& {95.14} & {93.79} & \cellcolor{deltabg}{1.35}
& {89.53} & {84.03} & \cellcolor{deltabg}{5.50} \\

\midrule
\textbf{Average}
& {88.87} & {83.62} & \cellcolor{deltabg}{5.24}
& {96.42} & {94.56} & \cellcolor{deltabg}{1.86}
& {96.87} & {95.47} & \cellcolor{deltabg}{1.40}
& {94.45} & {90.06} & \cellcolor{deltabg}{4.39} \\

\bottomrule
\end{tabular}
\end{table}

\begin{table}[t]
\centering
\caption{Robustness under \emph{Dimension and Unit Perturbation} (U2) (\%).}
\label{tab:L1.2}
\small
\renewcommand{\arraystretch}{1.08}
\setlength{\tabcolsep}{2.5pt}
\begin{tabular}{l|ccc|ccc|ccc|ccc}
\toprule
\multicolumn{1}{c|}{\multirow{2}{*}[-0.3ex]{\textbf{Model}}}
& \multicolumn{3}{c|}{\textbf{TACC}} 
& \multicolumn{3}{c|}{\textbf{IC}}
& \multicolumn{3}{c|}{\textbf{TP}}
& \multicolumn{3}{c}{\textbf{RV}} \\
\cmidrule(lr){2-4}
\cmidrule(lr){5-7}
\cmidrule(lr){8-10}
\cmidrule(lr){11-13}

& \textbf{Clean} & \textbf{Pert.} & $\boldsymbol{\Delta}$
& \textbf{Clean} & \textbf{Pert.} & $\boldsymbol{\Delta}$
& \textbf{Clean} & \textbf{Pert.} & $\boldsymbol{\Delta}$
& \textbf{Clean} & \textbf{Pert.} & $\boldsymbol{\Delta}$ \\
\midrule

Gemini 3.1 Pro
& {95.62} & {84.90} & \cellcolor{deltabg}{10.71}
& {98.85} & {94.64} & \cellcolor{deltabg}{4.21}
& {98.22} & {97.95} & \cellcolor{deltabg}{0.27}
& {99.35} & {96.20} & \cellcolor{deltabg}{3.15} \\

Gemini 3.7 Flash
& {90.95} & {85.16} & \cellcolor{deltabg}{5.78}
& {99.32} & {96.65} & \cellcolor{deltabg}{2.67}
& {98.63} & {97.44} & \cellcolor{deltabg}{1.19}
& {97.84} & {91.26} & \cellcolor{deltabg}{6.58} \\

\midrule
DeepSeek V4 Pro
& {92.69} & {88.13} & \cellcolor{deltabg}{4.56}
& {99.08} & {97.68} & \cellcolor{deltabg}{1.40}
& {97.70} & {96.64} & \cellcolor{deltabg}{1.06}
& {97.64} & {95.19} & \cellcolor{deltabg}{2.45} \\

DeepSeek V4 Flash
& {85.34} & {81.60} & \cellcolor{deltabg}{3.74}
& {95.47} & {94.50} & \cellcolor{deltabg}{0.97}
& {97.97} & {96.91} & \cellcolor{deltabg}{1.06}
& {94.29} & {89.77} & \cellcolor{deltabg}{4.52} \\

\midrule
GPT-5.6
& {89.62} & {82.23} & \cellcolor{deltabg}{7.40}
& {96.78} & {94.98} & \cellcolor{deltabg}{1.80}
& {95.90} & {95.37} & \cellcolor{deltabg}{0.53}
& {96.38} & {93.01} & \cellcolor{deltabg}{3.37} \\

GPT-5.5
& {91.18} & {85.86} & \cellcolor{deltabg}{5.32}
& {98.17} & {95.49} & \cellcolor{deltabg}{2.68}
& {98.57} & {97.57} & \cellcolor{deltabg}{1.00}
& {97.08} & {92.43} & \cellcolor{deltabg}{4.65} \\

\midrule
Claude Opus 5
& {83.25} & {81.90} & \cellcolor{deltabg}{1.35}
& {94.23} & {94.09} & \cellcolor{deltabg}{0.14}
& {97.58} & {97.26} & \cellcolor{deltabg}{0.32}
& {88.69} & {88.20} & \cellcolor{deltabg}{0.49} \\

Claude Sonnet 5
& {80.73} & {75.70} & \cellcolor{deltabg}{5.03}
& {93.44} & {91.33} & \cellcolor{deltabg}{2.11}
& {95.41} & {95.69} & \cellcolor{deltabg}{-0.28}
& {90.38} & {85.28} & \cellcolor{deltabg}{5.10} \\

\midrule
\textbf{Average}
& {88.67} & {83.18} & \cellcolor{deltabg}{5.49}
& {96.92} & {94.92} & \cellcolor{deltabg}{2.00}
& {97.50} & {96.85} & \cellcolor{deltabg}{0.64}
& {95.21} & {91.42} & \cellcolor{deltabg}{3.79} \\

\bottomrule
\end{tabular}
\end{table}

\begin{table}[t]
\centering
\caption{Robustness under \emph{Variable and Symbol Perturbation} (U3) (\%).}
\label{tab:L1.3}
\small
\renewcommand{\arraystretch}{1.08}
\setlength{\tabcolsep}{2.5pt}
\begin{tabular}{l|ccc|ccc|ccc|ccc}
\toprule
\multicolumn{1}{c|}{\multirow{2}{*}[-0.3ex]{\textbf{Model}}}
& \multicolumn{3}{c|}{\textbf{TACC}} 
& \multicolumn{3}{c|}{\textbf{IC}}
& \multicolumn{3}{c|}{\textbf{TP}}
& \multicolumn{3}{c}{\textbf{RV}} \\
\cmidrule(lr){2-4}
\cmidrule(lr){5-7}
\cmidrule(lr){8-10}
\cmidrule(lr){11-13}

& \textbf{Clean} & \textbf{Pert.} & $\boldsymbol{\Delta}$
& \textbf{Clean} & \textbf{Pert.} & $\boldsymbol{\Delta}$
& \textbf{Clean} & \textbf{Pert.} & $\boldsymbol{\Delta}$
& \textbf{Clean} & \textbf{Pert.} & $\boldsymbol{\Delta}$ \\
\midrule

Gemini 3.1 Pro
& {95.94} & {89.88} & \cellcolor{deltabg}{6.05}
& {99.11} & {95.89} & \cellcolor{deltabg}{3.22}
& {98.31} & {98.30} & \cellcolor{deltabg}{0.01}
& {98.68} & {97.47} & \cellcolor{deltabg}{1.21} \\

Gemini 3.7 Flash
& {92.77} & {88.34} & \cellcolor{deltabg}{4.43}
& {99.35} & {97.26} & \cellcolor{deltabg}{2.09}
& {96.28} & {96.00} & \cellcolor{deltabg}{0.29}
& {96.84} & {92.09} & \cellcolor{deltabg}{4.75} \\

\midrule
DeepSeek V4 Pro
& {94.63} & {91.56} & \cellcolor{deltabg}{3.07}
& {98.75} & {97.09} & \cellcolor{deltabg}{1.66}
& {95.68} & {96.69} & \cellcolor{deltabg}{-1.01}
& {97.41} & {93.83} & \cellcolor{deltabg}{3.58} \\

DeepSeek V4 Flash
& {81.02} & {75.13} & \cellcolor{deltabg}{5.89}
& {94.96} & {92.91} & \cellcolor{deltabg}{2.05}
& {96.70} & {95.25} & \cellcolor{deltabg}{1.45}
& {91.86} & {85.91} & \cellcolor{deltabg}{5.95} \\

\midrule
GPT-5.6
& {89.42} & {80.06} & \cellcolor{deltabg}{9.36}
& {96.94} & {94.13} & \cellcolor{deltabg}{2.81}
& {95.48} & {93.74} & \cellcolor{deltabg}{1.75}
& {94.80} & {89.47} & \cellcolor{deltabg}{5.33} \\

GPT-5.5
& {92.96} & {85.67} & \cellcolor{deltabg}{7.29}
& {98.22} & {94.84} & \cellcolor{deltabg}{3.38}
& {98.64} & {97.15} & \cellcolor{deltabg}{1.49}
& {96.22} & {90.93} & \cellcolor{deltabg}{5.29} \\

\midrule
Claude Opus 5
& {82.63} & {78.16} & \cellcolor{deltabg}{4.47}
& {93.91} & {93.44} & \cellcolor{deltabg}{0.47}
& {96.90} & {96.63} & \cellcolor{deltabg}{0.26}
& {87.11} & {85.20} & \cellcolor{deltabg}{1.90} \\

Claude Sonnet 5
& {76.51} & {72.74} & \cellcolor{deltabg}{3.77}
& {93.35} & {91.76} & \cellcolor{deltabg}{1.59}
& {94.49} & {94.95} & \cellcolor{deltabg}{-0.46}
& {87.90} & {83.18} & \cellcolor{deltabg}{4.72} \\

\midrule
\textbf{Average}
& {88.24} & {82.69} & \cellcolor{deltabg}{5.55}
& {96.82} & {94.67} & \cellcolor{deltabg}{2.15}
& {96.56} & {96.09} & \cellcolor{deltabg}{0.47}
& {93.85} & {89.76} & \cellcolor{deltabg}{4.09} \\

\bottomrule
\end{tabular}
\end{table}

\begin{table}[t]
\centering
\caption{Robustness under \emph{Condition Omission Perturbation} (U4) (\%).}
\label{tab:L1.4}
\small
\renewcommand{\arraystretch}{1.08}
\setlength{\tabcolsep}{2.5pt}
\begin{tabular}{l|ccc|ccc|ccc|ccc}
\toprule
\multicolumn{1}{c|}{\multirow{2}{*}[-0.3ex]{\textbf{Model}}}
& \multicolumn{3}{c|}{\textbf{TACC}} 
& \multicolumn{3}{c|}{\textbf{IC}}
& \multicolumn{3}{c|}{\textbf{TP}}
& \multicolumn{3}{c}{\textbf{RV}} \\
\cmidrule(lr){2-4}
\cmidrule(lr){5-7}
\cmidrule(lr){8-10}
\cmidrule(lr){11-13}

& \textbf{Clean} & \textbf{Pert.} & $\boldsymbol{\Delta}$
& \textbf{Clean} & \textbf{Pert.} & $\boldsymbol{\Delta}$
& \textbf{Clean} & \textbf{Pert.} & $\boldsymbol{\Delta}$
& \textbf{Clean} & \textbf{Pert.} & $\boldsymbol{\Delta}$ \\
\midrule

Gemini 3.1 Pro
& {95.89} & {77.74} & \cellcolor{deltabg}{18.16}
& {98.86} & {93.35} & \cellcolor{deltabg}{5.51}
& {98.68} & {95.02} & \cellcolor{deltabg}{3.66}
& {98.76} & {96.28} & \cellcolor{deltabg}{2.48} \\

Gemini 3.7 Flash
& {91.99} & {71.13} & \cellcolor{deltabg}{20.86}
& {98.99} & {93.47} & \cellcolor{deltabg}{5.52}
& {96.98} & {89.25} & \cellcolor{deltabg}{7.73}
& {96.44} & {88.04} & \cellcolor{deltabg}{8.40} \\

\midrule
DeepSeek V4 Pro
& {93.98} & {70.26} & \cellcolor{deltabg}{23.73}
& {98.61} & {92.59} & \cellcolor{deltabg}{6.02}
& {96.81} & {90.30} & \cellcolor{deltabg}{6.51}
& {97.55} & {91.39} & \cellcolor{deltabg}{6.16} \\

DeepSeek V4 Flash
& {80.87} & {54.94} & \cellcolor{deltabg}{25.93}
& {94.81} & {88.12} & \cellcolor{deltabg}{6.69}
& {96.34} & {88.54} & \cellcolor{deltabg}{7.80}
& {92.43} & {81.79} & \cellcolor{deltabg}{10.65} \\

\midrule
GPT-5.6
& {87.55} & {63.30} & \cellcolor{deltabg}{24.25}
& {96.24} & {92.28} & \cellcolor{deltabg}{3.96}
& {96.15} & {89.00} & \cellcolor{deltabg}{7.15}
& {95.20} & {89.42} & \cellcolor{deltabg}{5.77} \\

GPT-5.5
& {92.11} & {66.72} & \cellcolor{deltabg}{25.39}
& {97.80} & {89.95} & \cellcolor{deltabg}{7.85}
& {98.79} & {94.14} & \cellcolor{deltabg}{4.65}
& {96.01} & {81.95} & \cellcolor{deltabg}{14.06} \\

\midrule
Claude Opus 5
& {76.10} & {65.24} & \cellcolor{deltabg}{10.86}
& {92.41} & {89.84} & \cellcolor{deltabg}{2.56}
& {96.63} & {92.79} & \cellcolor{deltabg}{3.84}
& {85.31} & {79.78} & \cellcolor{deltabg}{5.53} \\

Claude Sonnet 5
& {76.03} & {53.49} & \cellcolor{deltabg}{22.53}
& {92.76} & {87.10} & \cellcolor{deltabg}{5.66}
& {94.94} & {87.11} & \cellcolor{deltabg}{7.83}
& {88.10} & {80.39} & \cellcolor{deltabg}{7.72} \\

\midrule
\textbf{Average}
& {86.82} & {65.35} & \cellcolor{deltabg}{21.46}
& {96.31} & {90.84} & \cellcolor{deltabg}{5.47}
& {96.92} & {90.77} & \cellcolor{deltabg}{6.15}
& {93.73} & {86.13} & \cellcolor{deltabg}{7.59} \\

\bottomrule
\end{tabular}
\end{table}

\begin{table}[t]
\centering
\caption{Robustness under \emph{Invalid Premise Perturbation} (E1) (\%).}
\label{tab:L2.1}
\small
\renewcommand{\arraystretch}{1.08}
\setlength{\tabcolsep}{2.5pt}
\begin{tabular}{l|ccc|ccc|ccc|ccc}
\toprule
\multicolumn{1}{c|}{\multirow{2}{*}[-0.3ex]{\textbf{Model}}}
& \multicolumn{3}{c|}{\textbf{TACC}} 
& \multicolumn{3}{c|}{\textbf{IC}}
& \multicolumn{3}{c|}{\textbf{TP}}
& \multicolumn{3}{c}{\textbf{RV}} \\
\cmidrule(lr){2-4}
\cmidrule(lr){5-7}
\cmidrule(lr){8-10}
\cmidrule(lr){11-13}

& \textbf{Clean} & \textbf{Pert.} & $\boldsymbol{\Delta}$
& \textbf{Clean} & \textbf{Pert.} & $\boldsymbol{\Delta}$
& \textbf{Clean} & \textbf{Pert.} & $\boldsymbol{\Delta}$
& \textbf{Clean} & \textbf{Pert.} & $\boldsymbol{\Delta}$ \\
\midrule

Gemini 3.1 Pro
& {97.49} & {67.59} & \cellcolor{deltabg}{29.89}
& {99.32} & {95.16} & \cellcolor{deltabg}{4.17}
& {100.00} & {98.84} & \cellcolor{deltabg}{1.16}
& {99.92} & {97.77} & \cellcolor{deltabg}{2.15} \\

Gemini 3.7 Flash
& {94.97} & {54.15} & \cellcolor{deltabg}{40.82}
& {99.15} & {90.50} & \cellcolor{deltabg}{8.65}
& {98.71} & {96.98} & \cellcolor{deltabg}{1.73}
& {97.86} & {86.45} & \cellcolor{deltabg}{11.42} \\

\midrule
DeepSeek V4 Pro
& {97.10} & {70.90} & \cellcolor{deltabg}{26.20}
& {99.19} & {93.14} & \cellcolor{deltabg}{6.05}
& {99.90} & {97.32} & \cellcolor{deltabg}{2.58}
& {99.42} & {89.80} & \cellcolor{deltabg}{9.63} \\

DeepSeek V4 Flash
& {90.92} & {73.98} & \cellcolor{deltabg}{16.95}
& {97.57} & {91.90} & \cellcolor{deltabg}{5.67}
& {99.43} & {98.23} & \cellcolor{deltabg}{1.20}
& {96.58} & {90.63} & \cellcolor{deltabg}{5.94} \\

\midrule
GPT-5.6
& {96.41} & {65.35} & \cellcolor{deltabg}{31.06}
& {99.06} & {91.98} & \cellcolor{deltabg}{7.08}
& {99.95} & {97.30} & \cellcolor{deltabg}{2.65}
& {98.64} & {86.99} & \cellcolor{deltabg}{11.65} \\

GPT-5.5
& {94.82} & {73.53} & \cellcolor{deltabg}{21.29}
& {98.86} & {92.75} & \cellcolor{deltabg}{6.11}
& {99.28} & {98.46} & \cellcolor{deltabg}{0.81}
& {96.38} & {86.13} & \cellcolor{deltabg}{10.25} \\

\midrule
Claude Opus 5
& {89.07} & {72.17} & \cellcolor{deltabg}{16.91}
& {96.15} & {92.54} & \cellcolor{deltabg}{3.61}
& {99.03} & {98.67} & \cellcolor{deltabg}{0.36}
& {92.48} & {82.90} & \cellcolor{deltabg}{9.58} \\

Claude Sonnet 5
& {90.49} & {69.23} & \cellcolor{deltabg}{21.27}
& {97.10} & {92.60} & \cellcolor{deltabg}{4.50}
& {99.53} & {98.24} & \cellcolor{deltabg}{1.29}
& {95.39} & {88.59} & \cellcolor{deltabg}{6.80} \\

\midrule
\textbf{Average}
& {93.91} & {68.36} & \cellcolor{deltabg}{25.55}
& {98.30} & {92.57} & \cellcolor{deltabg}{5.73}
& {99.48} & {98.00} & \cellcolor{deltabg}{1.47}
& {97.08} & {88.66} & \cellcolor{deltabg}{8.43} \\

\bottomrule
\end{tabular}
\end{table}

\begin{table}[t]
\centering
\caption{Robustness under \emph{Citation and Attribution Perturbation} (E2) (\%).}
\label{tab:L2.2}
\small
\renewcommand{\arraystretch}{1.08}
\setlength{\tabcolsep}{2.5pt}
\begin{tabular}{l|ccc|ccc|ccc|ccc}
\toprule
\multicolumn{1}{c|}{\multirow{2}{*}[-0.3ex]{\textbf{Model}}}
& \multicolumn{3}{c|}{\textbf{TACC}} 
& \multicolumn{3}{c|}{\textbf{IC}}
& \multicolumn{3}{c|}{\textbf{TP}}
& \multicolumn{3}{c}{\textbf{RV}} \\
\cmidrule(lr){2-4}
\cmidrule(lr){5-7}
\cmidrule(lr){8-10}
\cmidrule(lr){11-13}

& \textbf{Clean} & \textbf{Pert.} & $\boldsymbol{\Delta}$
& \textbf{Clean} & \textbf{Pert.} & $\boldsymbol{\Delta}$
& \textbf{Clean} & \textbf{Pert.} & $\boldsymbol{\Delta}$
& \textbf{Clean} & \textbf{Pert.} & $\boldsymbol{\Delta}$ \\
\midrule

Gemini 3.1 Pro
& {97.50} & {86.23} & \cellcolor{deltabg}{11.27}
& {99.33} & {98.79} & \cellcolor{deltabg}{0.54}
& {100.00} & {100.00} & \cellcolor{deltabg}{0.00}
& {99.92} & {99.21} & \cellcolor{deltabg}{0.71} \\

Gemini 3.7 Flash
& {94.97} & {88.02} & \cellcolor{deltabg}{6.95}
& {99.15} & {98.40} & \cellcolor{deltabg}{0.75}
& {98.71} & {98.78} & \cellcolor{deltabg}{-0.07}
& {97.86} & {97.27} & \cellcolor{deltabg}{0.59} \\

\midrule
DeepSeek V4 Pro
& {97.58} & {90.24} & \cellcolor{deltabg}{7.34}
& {99.36} & {98.56} & \cellcolor{deltabg}{0.79}
& {99.90} & {99.20} & \cellcolor{deltabg}{0.70}
& {99.50} & {98.90} & \cellcolor{deltabg}{0.60} \\

DeepSeek V4 Flash
& {90.72} & {75.34} & \cellcolor{deltabg}{15.38}
& {97.35} & {95.50} & \cellcolor{deltabg}{1.85}
& {99.43} & {97.46} & \cellcolor{deltabg}{1.97}
& {96.32} & {94.72} & \cellcolor{deltabg}{1.60} \\

\midrule
GPT-5.6
& {96.41} & {84.17} & \cellcolor{deltabg}{12.24}
& {99.06} & {97.50} & \cellcolor{deltabg}{1.56}
& {99.95} & {98.83} & \cellcolor{deltabg}{1.12}
& {98.64} & {97.15} & \cellcolor{deltabg}{1.49} \\

GPT-5.5
& {94.82} & {86.30} & \cellcolor{deltabg}{8.52}
& {98.86} & {97.39} & \cellcolor{deltabg}{1.47}
& {99.28} & {98.68} & \cellcolor{deltabg}{0.59}
& {96.38} & {94.19} & \cellcolor{deltabg}{2.19} \\

\midrule
Claude Opus 5
& {89.65} & {81.58} & \cellcolor{deltabg}{8.06}
& {96.16} & {95.99} & \cellcolor{deltabg}{0.17}
& {99.02} & {98.44} & \cellcolor{deltabg}{0.57}
& {92.55} & {92.49} & \cellcolor{deltabg}{0.06} \\

Claude Sonnet 5
& {90.49} & {70.92} & \cellcolor{deltabg}{19.58}
& {97.07} & {95.12} & \cellcolor{deltabg}{1.95}
& {99.53} & {98.07} & \cellcolor{deltabg}{1.46}
& {95.39} & {91.61} & \cellcolor{deltabg}{3.78} \\

\midrule
\textbf{Average}
& {94.02} & {82.85} & \cellcolor{deltabg}{11.17}
& {98.29} & {97.16} & \cellcolor{deltabg}{1.13}
& {99.48} & {98.68} & \cellcolor{deltabg}{0.79}
& {97.07} & {95.69} & \cellcolor{deltabg}{1.38} \\

\bottomrule
\end{tabular}
\end{table}

\begin{table}[t]
\centering
\caption{Robustness under \emph{Data and Statistical Evidence Perturbation} (E3) (\%).}
\label{tab:L2.3}
\small
\renewcommand{\arraystretch}{1.08}
\setlength{\tabcolsep}{2.5pt}
\begin{tabular}{l|ccc|ccc|ccc|ccc}
\toprule
\multicolumn{1}{c|}{\multirow{2}{*}[-0.3ex]{\textbf{Model}}}
& \multicolumn{3}{c|}{\textbf{TACC}} 
& \multicolumn{3}{c|}{\textbf{IC}}
& \multicolumn{3}{c|}{\textbf{TP}}
& \multicolumn{3}{c}{\textbf{RV}} \\
\cmidrule(lr){2-4}
\cmidrule(lr){5-7}
\cmidrule(lr){8-10}
\cmidrule(lr){11-13}

& \textbf{Clean} & \textbf{Pert.} & $\boldsymbol{\Delta}$
& \textbf{Clean} & \textbf{Pert.} & $\boldsymbol{\Delta}$
& \textbf{Clean} & \textbf{Pert.} & $\boldsymbol{\Delta}$
& \textbf{Clean} & \textbf{Pert.} & $\boldsymbol{\Delta}$ \\
\midrule

Gemini 3.1 Pro
& {97.49} & {44.83} & \cellcolor{deltabg}{52.66}
& {99.32} & {90.32} & \cellcolor{deltabg}{9.01}
& {100.00} & {99.15} & \cellcolor{deltabg}{0.85}
& {99.92} & {97.40} & \cellcolor{deltabg}{2.51} \\

Gemini 3.7 Flash
& {94.97} & {46.45} & \cellcolor{deltabg}{48.51}
& {99.15} & {91.12} & \cellcolor{deltabg}{8.03}
& {98.71} & {96.17} & \cellcolor{deltabg}{2.54}
& {97.86} & {84.05} & \cellcolor{deltabg}{13.81} \\

\midrule
DeepSeek V4 Pro
& {96.84} & {53.45} & \cellcolor{deltabg}{43.39}
& {99.19} & {93.22} & \cellcolor{deltabg}{5.96}
& {99.90} & {96.07} & \cellcolor{deltabg}{3.82}
& {99.34} & {87.36} & \cellcolor{deltabg}{11.97} \\

DeepSeek V4 Flash
& {90.64} & {52.30} & \cellcolor{deltabg}{38.34}
& {97.32} & {89.24} & \cellcolor{deltabg}{8.07}
& {99.43} & {98.23} & \cellcolor{deltabg}{1.19}
& {96.29} & {82.50} & \cellcolor{deltabg}{13.79} \\

\midrule
GPT-5.6
& {96.41} & {45.35} & \cellcolor{deltabg}{51.06}
& {99.06} & {90.54} & \cellcolor{deltabg}{8.52}
& {99.95} & {96.14} & \cellcolor{deltabg}{3.82}
& {98.64} & {81.69} & \cellcolor{deltabg}{16.95} \\

GPT-5.5
& {94.82} & {48.49} & \cellcolor{deltabg}{46.33}
& {98.86} & {89.94} & \cellcolor{deltabg}{8.91}
& {99.28} & {98.06} & \cellcolor{deltabg}{1.22}
& {96.38} & {74.19} & \cellcolor{deltabg}{22.19} \\

\midrule
Claude Opus 5
& {89.72} & {55.30} & \cellcolor{deltabg}{34.43}
& {96.17} & {90.58} & \cellcolor{deltabg}{5.60}
& {99.02} & {97.15} & \cellcolor{deltabg}{1.88}
& {92.61} & {77.71} & \cellcolor{deltabg}{14.89} \\

Claude Sonnet 5
& {90.56} & {49.61} & \cellcolor{deltabg}{40.95}
& {97.09} & {90.41} & \cellcolor{deltabg}{6.68}
& {99.53} & {97.80} & \cellcolor{deltabg}{1.73}
& {95.19} & {83.14} & \cellcolor{deltabg}{12.05} \\

\midrule
\textbf{Average}
& {93.93} & {49.47} & \cellcolor{deltabg}{44.46}
& {98.27} & {90.67} & \cellcolor{deltabg}{7.60}
& {99.48} & {97.34} & \cellcolor{deltabg}{2.13}
& {97.03} & {83.51} & \cellcolor{deltabg}{13.52} \\

\bottomrule
\end{tabular}
\end{table}

\begin{table}[t]
\centering
\caption{Robustness under \emph{Multi-Source Evidence Conflict Perturbation} (E4) (\%).}
\label{tab:L2.4}
\small
\renewcommand{\arraystretch}{1.08}
\setlength{\tabcolsep}{2.5pt}
\begin{tabular}{l|ccc|ccc|ccc|ccc}
\toprule
\multicolumn{1}{c|}{\multirow{2}{*}[-0.3ex]{\textbf{Model}}}
& \multicolumn{3}{c|}{\textbf{TACC}} 
& \multicolumn{3}{c|}{\textbf{IC}}
& \multicolumn{3}{c|}{\textbf{TP}}
& \multicolumn{3}{c}{\textbf{RV}} \\
\cmidrule(lr){2-4}
\cmidrule(lr){5-7}
\cmidrule(lr){8-10}
\cmidrule(lr){11-13}

& \textbf{Clean} & \textbf{Pert.} & $\boldsymbol{\Delta}$
& \textbf{Clean} & \textbf{Pert.} & $\boldsymbol{\Delta}$
& \textbf{Clean} & \textbf{Pert.} & $\boldsymbol{\Delta}$
& \textbf{Clean} & \textbf{Pert.} & $\boldsymbol{\Delta}$ \\
\midrule

Gemini 3.1 Pro
& {97.50} & {89.48} & \cellcolor{deltabg}{8.02}
& {99.33} & {98.66} & \cellcolor{deltabg}{0.67}
& {100.00} & {99.86} & \cellcolor{deltabg}{0.14}
& {99.92} & {99.47} & \cellcolor{deltabg}{0.45} \\

Gemini 3.7 Flash
& {94.97} & {80.77} & \cellcolor{deltabg}{14.19}
& {99.15} & {97.05} & \cellcolor{deltabg}{2.10}
& {98.71} & {97.93} & \cellcolor{deltabg}{0.78}
& {97.86} & {94.18} & \cellcolor{deltabg}{3.68} \\

\midrule
DeepSeek V4 Pro
& {97.07} & {90.91} & \cellcolor{deltabg}{6.16}
& {99.19} & {97.95} & \cellcolor{deltabg}{1.23}
& {99.90} & {99.07} & \cellcolor{deltabg}{0.83}
& {99.42} & {97.54} & \cellcolor{deltabg}{1.88} \\

DeepSeek V4 Flash
& {90.31} & {86.20} & \cellcolor{deltabg}{4.11}
& {97.30} & {97.67} & \cellcolor{deltabg}{-0.37}
& {99.44} & {98.78} & \cellcolor{deltabg}{0.66}
& {96.20} & {96.35} & \cellcolor{deltabg}{-0.16} \\

\midrule
GPT-5.6
& {96.41} & {92.05} & \cellcolor{deltabg}{4.36}
& {99.06} & {98.33} & \cellcolor{deltabg}{0.73}
& {99.95} & {99.11} & \cellcolor{deltabg}{0.84}
& {98.64} & {98.26} & \cellcolor{deltabg}{0.37} \\

GPT-5.5
& {94.82} & {90.71} & \cellcolor{deltabg}{4.11}
& {98.86} & {97.42} & \cellcolor{deltabg}{1.44}
& {99.28} & {98.93} & \cellcolor{deltabg}{0.35}
& {96.38} & {94.84} & \cellcolor{deltabg}{1.54} \\

\midrule
Claude Opus 5
& {89.89} & {87.72} & \cellcolor{deltabg}{2.17}
& {96.41} & {96.30} & \cellcolor{deltabg}{0.11}
& {99.26} & {99.14} & \cellcolor{deltabg}{0.12}
& {93.04} & {92.37} & \cellcolor{deltabg}{0.67} \\

Claude Sonnet 5
& {90.49} & {87.80} & \cellcolor{deltabg}{2.70}
& {97.06} & {96.33} & \cellcolor{deltabg}{0.73}
& {99.53} & {98.96} & \cellcolor{deltabg}{0.57}
& {95.15} & {92.57} & \cellcolor{deltabg}{2.58} \\

\midrule
\textbf{Average}
& {93.93} & {88.21} & \cellcolor{deltabg}{5.73}
& {98.29} & {97.46} & \cellcolor{deltabg}{0.83}
& {99.51} & {98.97} & \cellcolor{deltabg}{0.54}
& {97.08} & {95.70} & \cellcolor{deltabg}{1.38} \\

\bottomrule
\end{tabular}
\end{table}

\begin{table}[t]
\centering
\caption{Robustness under \emph{Cross-Step Inconsistency Perturbation} (R1) (\%).}
\label{tab:L3.1}
\small
\renewcommand{\arraystretch}{1.08}
\setlength{\tabcolsep}{2.5pt}
\begin{tabular}{l|ccc|ccc|ccc|ccc}
\toprule
\multicolumn{1}{c|}{\multirow{2}{*}[-0.3ex]{\textbf{Model}}}
& \multicolumn{3}{c|}{\textbf{TACC}} 
& \multicolumn{3}{c|}{\textbf{IC}}
& \multicolumn{3}{c|}{\textbf{TP}}
& \multicolumn{3}{c}{\textbf{RV}} \\
\cmidrule(lr){2-4}
\cmidrule(lr){5-7}
\cmidrule(lr){8-10}
\cmidrule(lr){11-13}

& \textbf{Clean} & \textbf{Pert.} & $\boldsymbol{\Delta}$
& \textbf{Clean} & \textbf{Pert.} & $\boldsymbol{\Delta}$
& \textbf{Clean} & \textbf{Pert.} & $\boldsymbol{\Delta}$
& \textbf{Clean} & \textbf{Pert.} & $\boldsymbol{\Delta}$ \\
\midrule

Gemini 3.1 Pro
& {96.69} & {44.43} & \cellcolor{deltabg}{52.26}
& {98.74} & {79.13} & \cellcolor{deltabg}{19.61}
& {98.78} & {91.79} & \cellcolor{deltabg}{6.99}
& {96.59} & {50.68} & \cellcolor{deltabg}{45.91} \\

Gemini 3.7 Flash
& {92.75} & {37.39} & \cellcolor{deltabg}{55.36}
& {99.06} & {79.78} & \cellcolor{deltabg}{19.28}
& {96.92} & {84.72} & \cellcolor{deltabg}{12.21}
& {96.85} & {60.30} & \cellcolor{deltabg}{36.55} \\

\midrule
DeepSeek V4 Pro
& {94.67} & {39.03} & \cellcolor{deltabg}{55.64}
& {98.46} & {81.41} & \cellcolor{deltabg}{17.05}
& {95.56} & {88.98} & \cellcolor{deltabg}{6.58}
& {97.94} & {63.68} & \cellcolor{deltabg}{34.27} \\

DeepSeek V4 Flash
& {85.10} & {36.74} & \cellcolor{deltabg}{48.37}
& {94.71} & {79.97} & \cellcolor{deltabg}{14.74}
& {95.94} & {87.99} & \cellcolor{deltabg}{7.95}
& {92.26} & {62.84} & \cellcolor{deltabg}{29.42} \\

\midrule
GPT-5.6
& {88.56} & {31.55} & \cellcolor{deltabg}{57.01}
& {96.52} & {77.02} & \cellcolor{deltabg}{19.50}
& {95.01} & {85.10} & \cellcolor{deltabg}{9.91}
& {95.09} & {61.81} & \cellcolor{deltabg}{33.28} \\

GPT-5.5
& {93.07} & {38.19} & \cellcolor{deltabg}{54.88}
& {97.88} & {74.93} & \cellcolor{deltabg}{22.95}
& {98.42} & {91.51} & \cellcolor{deltabg}{6.91}
& {95.90} & {51.26} & \cellcolor{deltabg}{44.65} \\

\midrule
Claude Opus 5
& {81.65} & {38.82} & \cellcolor{deltabg}{42.82}
& {92.67} & {79.10} & \cellcolor{deltabg}{13.57}
& {96.31} & {90.29} & \cellcolor{deltabg}{6.02}
& {86.51} & {52.46} & \cellcolor{deltabg}{34.05} \\

Claude Sonnet 5
& {79.97} & {35.20} & \cellcolor{deltabg}{44.77}
& {92.40} & {77.79} & \cellcolor{deltabg}{14.61}
& {93.76} & {86.58} & \cellcolor{deltabg}{7.18}
& {88.51} & {57.98} & \cellcolor{deltabg}{30.53} \\

\midrule
\textbf{Average}
& {89.06} & {37.67} & \cellcolor{deltabg}{51.39}
& {96.31} & {78.64} & \cellcolor{deltabg}{17.66}
& {96.34} & {88.37} & \cellcolor{deltabg}{7.97}
& {93.71} & {57.63} & \cellcolor{deltabg}{36.08} \\

\bottomrule
\end{tabular}
\end{table}

\begin{table}[t]
\centering
\caption{Robustness under \emph{Correlation--Causation Confusion Perturbation} (R2) (\%).}
\label{tab:L3.2}
\small
\renewcommand{\arraystretch}{1.08}
\setlength{\tabcolsep}{2.5pt}
\begin{tabular}{l|ccc|ccc|ccc|ccc}
\toprule
\multicolumn{1}{c|}{\multirow{2}{*}[-0.3ex]{\textbf{Model}}}
& \multicolumn{3}{c|}{\textbf{TACC}} 
& \multicolumn{3}{c|}{\textbf{IC}}
& \multicolumn{3}{c|}{\textbf{TP}}
& \multicolumn{3}{c}{\textbf{RV}} \\
\cmidrule(lr){2-4}
\cmidrule(lr){5-7}
\cmidrule(lr){8-10}
\cmidrule(lr){11-13}

& \textbf{Clean} & \textbf{Pert.} & $\boldsymbol{\Delta}$
& \textbf{Clean} & \textbf{Pert.} & $\boldsymbol{\Delta}$
& \textbf{Clean} & \textbf{Pert.} & $\boldsymbol{\Delta}$
& \textbf{Clean} & \textbf{Pert.} & $\boldsymbol{\Delta}$ \\
\midrule

Gemini 3.1 Pro
& {98.64} & {29.16} & \cellcolor{deltabg}{69.47}
& {99.09} & {77.83} & \cellcolor{deltabg}{21.26}
& {99.78} & {48.10} & \cellcolor{deltabg}{51.68}
& {98.83} & {50.69} & \cellcolor{deltabg}{48.13} \\

Gemini 3.7 Flash
& {97.71} & {25.27} & \cellcolor{deltabg}{72.44}
& {97.92} & {75.59} & \cellcolor{deltabg}{22.34}
& {97.88} & {46.32} & \cellcolor{deltabg}{51.56}
& {95.83} & {43.24} & \cellcolor{deltabg}{52.59} \\

\midrule
DeepSeek V4 Pro
& {99.10} & {31.52} & \cellcolor{deltabg}{67.58}
& {99.28} & {84.73} & \cellcolor{deltabg}{14.54}
& {98.21} & {48.90} & \cellcolor{deltabg}{49.31}
& {99.49} & {66.49} & \cellcolor{deltabg}{33.00} \\

DeepSeek V4 Flash
& {94.79} & {29.85} & \cellcolor{deltabg}{64.94}
& {98.69} & {82.84} & \cellcolor{deltabg}{15.85}
& {99.25} & {42.56} & \cellcolor{deltabg}{56.69}
& {97.95} & {66.19} & \cellcolor{deltabg}{31.76} \\

\midrule
GPT-5.6
& {98.08} & {30.38} & \cellcolor{deltabg}{67.70}
& {99.08} & {88.12} & \cellcolor{deltabg}{10.96}
& {98.27} & {41.32} & \cellcolor{deltabg}{56.95}
& {98.85} & {79.23} & \cellcolor{deltabg}{19.62} \\

GPT-5.5
& {95.21} & {29.07} & \cellcolor{deltabg}{66.14}
& {98.15} & {81.71} & \cellcolor{deltabg}{16.44}
& {98.28} & {40.57} & \cellcolor{deltabg}{57.71}
& {94.73} & {68.84} & \cellcolor{deltabg}{25.90} \\

\midrule
Claude Opus 5
& {93.09} & {31.79} & \cellcolor{deltabg}{61.30}
& {95.51} & {93.23} & \cellcolor{deltabg}{2.28}
& {99.80} & {45.18} & \cellcolor{deltabg}{54.62}
& {91.52} & {73.45} & \cellcolor{deltabg}{18.07} \\

Claude Sonnet 5
& {92.60} & {28.55} & \cellcolor{deltabg}{64.05}
& {96.12} & {90.00} & \cellcolor{deltabg}{6.12}
& {98.12} & {35.71} & \cellcolor{deltabg}{62.41}
& {92.02} & {91.51} & \cellcolor{deltabg}{0.51} \\

\midrule
\textbf{Average}
& {96.15} & {29.45} & \cellcolor{deltabg}{66.70}
& {97.98} & {84.26} & \cellcolor{deltabg}{13.73}
& {98.70} & {43.58} & \cellcolor{deltabg}{55.12}
& {96.15} & {67.46} & \cellcolor{deltabg}{28.70} \\

\bottomrule
\end{tabular}
\end{table}

\begin{table}[t]
\centering
\caption{Robustness under \emph{Reasoning-Path Misdirection Perturbation} (R3) (\%).}
\label{tab:L3.3}
\small
\renewcommand{\arraystretch}{1.08}
\setlength{\tabcolsep}{2.5pt}
\begin{tabular}{l|ccc|ccc|ccc|ccc}
\toprule
\multicolumn{1}{c|}{\multirow{2}{*}[-0.3ex]{\textbf{Model}}}
& \multicolumn{3}{c|}{\textbf{TACC}} 
& \multicolumn{3}{c|}{\textbf{IC}}
& \multicolumn{3}{c|}{\textbf{TP}}
& \multicolumn{3}{c}{\textbf{RV}} \\
\cmidrule(lr){2-4}
\cmidrule(lr){5-7}
\cmidrule(lr){8-10}
\cmidrule(lr){11-13}

& \textbf{Clean} & \textbf{Pert.} & $\boldsymbol{\Delta}$
& \textbf{Clean} & \textbf{Pert.} & $\boldsymbol{\Delta}$
& \textbf{Clean} & \textbf{Pert.} & $\boldsymbol{\Delta}$
& \textbf{Clean} & \textbf{Pert.} & $\boldsymbol{\Delta}$ \\
\midrule

Gemini 3.1 Pro
& {95.75} & {61.82} & \cellcolor{deltabg}{33.93}
& {98.63} & {84.18} & \cellcolor{deltabg}{14.45}
& {98.73} & {94.64} & \cellcolor{deltabg}{4.09}
& {96.31} & {66.19} & \cellcolor{deltabg}{30.12} \\

Gemini 3.7 Flash
& {91.46} & {40.75} & \cellcolor{deltabg}{50.71}
& {98.99} & {79.62} & \cellcolor{deltabg}{19.37}
& {97.16} & {85.98} & \cellcolor{deltabg}{11.18}
& {96.63} & {61.77} & \cellcolor{deltabg}{34.86} \\

\midrule
DeepSeek V4 Pro
& {94.61} & {65.62} & \cellcolor{deltabg}{29.00}
& {98.71} & {88.07} & \cellcolor{deltabg}{10.65}
& {96.22} & {92.28} & \cellcolor{deltabg}{3.93}
& {98.10} & {79.74} & \cellcolor{deltabg}{18.36} \\

DeepSeek V4 Flash
& {83.90} & {47.10} & \cellcolor{deltabg}{36.80}
& {94.87} & {84.00} & \cellcolor{deltabg}{10.87}
& {96.30} & {89.23} & \cellcolor{deltabg}{7.07}
& {92.39} & {69.15} & \cellcolor{deltabg}{23.24} \\

\midrule
GPT-5.6
& {87.04} & {52.10} & \cellcolor{deltabg}{34.94}
& {96.51} & {85.00} & \cellcolor{deltabg}{11.50}
& {95.58} & {90.65} & \cellcolor{deltabg}{4.93}
& {95.15} & {75.21} & \cellcolor{deltabg}{19.94} \\

GPT-5.5
& {92.10} & {64.09} & \cellcolor{deltabg}{28.01}
& {97.87} & {85.06} & \cellcolor{deltabg}{12.81}
& {98.68} & {95.27} & \cellcolor{deltabg}{3.41}
& {95.64} & {73.57} & \cellcolor{deltabg}{22.07} \\

\midrule
Claude Opus 5
& {80.04} & {54.13} & \cellcolor{deltabg}{25.91}
& {92.94} & {84.99} & \cellcolor{deltabg}{7.95}
& {97.23} & {93.69} & \cellcolor{deltabg}{3.53}
& {86.06} & {64.07} & \cellcolor{deltabg}{22.00} \\

Claude Sonnet 5
& {77.79} & {45.27} & \cellcolor{deltabg}{32.52}
& {92.46} & {82.22} & \cellcolor{deltabg}{10.24}
& {94.94} & {90.26} & \cellcolor{deltabg}{4.69}
& {87.96} & {67.89} & \cellcolor{deltabg}{20.07} \\

\midrule
\textbf{Average}
& {87.84} & {53.86} & \cellcolor{deltabg}{33.98}
& {96.37} & {84.14} & \cellcolor{deltabg}{12.23}
& {96.85} & {91.50} & \cellcolor{deltabg}{5.36}
& {93.53} & {69.70} & \cellcolor{deltabg}{23.83} \\

\bottomrule
\end{tabular}
\end{table}

\begin{table}[t]
\centering
\caption{Robustness under \emph{Uncertainty Suppression Perturbation} (C1) (\%).}
\label{tab:L4.1}
\small
\renewcommand{\arraystretch}{1.08}
\setlength{\tabcolsep}{2.5pt}
\begin{tabular}{l|ccc|ccc|ccc|ccc}
\toprule
\multicolumn{1}{c|}{\multirow{2}{*}[-0.3ex]{\textbf{Model}}}
& \multicolumn{3}{c|}{\textbf{TACC}} 
& \multicolumn{3}{c|}{\textbf{IC}}
& \multicolumn{3}{c|}{\textbf{TP}}
& \multicolumn{3}{c}{\textbf{RV}} \\
\cmidrule(lr){2-4}
\cmidrule(lr){5-7}
\cmidrule(lr){8-10}
\cmidrule(lr){11-13}

& \textbf{Clean} & \textbf{Pert.} & $\boldsymbol{\Delta}$
& \textbf{Clean} & \textbf{Pert.} & $\boldsymbol{\Delta}$
& \textbf{Clean} & \textbf{Pert.} & $\boldsymbol{\Delta}$
& \textbf{Clean} & \textbf{Pert.} & $\boldsymbol{\Delta}$ \\
\midrule

Gemini 3.1 Pro
& {95.68} & {81.80} & \cellcolor{deltabg}{13.88}
& {98.84} & {90.14} & \cellcolor{deltabg}{8.70}
& {98.68} & {96.12} & \cellcolor{deltabg}{2.56}
& {99.13} & {95.25} & \cellcolor{deltabg}{3.88} \\

Gemini 3.7 Flash
& {94.89} & {81.13} & \cellcolor{deltabg}{13.76}
& {99.53} & {75.98} & \cellcolor{deltabg}{23.55}
& {98.58} & {87.65} & \cellcolor{deltabg}{10.93}
& {98.68} & {73.97} & \cellcolor{deltabg}{24.71} \\

\midrule
DeepSeek V4 Pro
& {95.34} & {88.53} & \cellcolor{deltabg}{6.81}
& {98.05} & {90.74} & \cellcolor{deltabg}{7.31}
& {94.09} & {84.80} & \cellcolor{deltabg}{9.29}
& {97.49} & {89.77} & \cellcolor{deltabg}{7.73} \\

DeepSeek V4 Flash
& {87.43} & {77.46} & \cellcolor{deltabg}{9.98}
& {95.30} & {87.51} & \cellcolor{deltabg}{7.80}
& {94.95} & {94.59} & \cellcolor{deltabg}{0.36}
& {94.49} & {79.90} & \cellcolor{deltabg}{14.59} \\

\midrule
GPT-5.6
& {89.99} & {82.94} & \cellcolor{deltabg}{7.05}
& {95.61} & {90.12} & \cellcolor{deltabg}{5.48}
& {92.84} & {87.48} & \cellcolor{deltabg}{5.36}
& {96.02} & {88.77} & \cellcolor{deltabg}{7.25} \\

GPT-5.5
& {92.92} & {85.54} & \cellcolor{deltabg}{7.38}
& {97.85} & {91.06} & \cellcolor{deltabg}{6.79}
& {98.45} & {96.17} & \cellcolor{deltabg}{2.29}
& {96.08} & {87.43} & \cellcolor{deltabg}{8.65} \\

\midrule
Claude Opus 5
& {85.13} & {76.30} & \cellcolor{deltabg}{8.84}
& {93.76} & {85.03} & \cellcolor{deltabg}{8.73}
& {97.38} & {92.44} & \cellcolor{deltabg}{4.94}
& {88.59} & {73.19} & \cellcolor{deltabg}{15.40} \\

Claude Sonnet 5
& {86.27} & {76.58} & \cellcolor{deltabg}{9.69}
& {94.43} & {82.67} & \cellcolor{deltabg}{11.76}
& {93.40} & {90.55} & \cellcolor{deltabg}{2.84}
& {93.13} & {79.68} & \cellcolor{deltabg}{13.45} \\

\midrule
\textbf{Average}
& {90.96} & {81.28} & \cellcolor{deltabg}{9.67}
& {96.67} & {86.66} & \cellcolor{deltabg}{10.01}
& {96.05} & {91.23} & \cellcolor{deltabg}{4.82}
& {95.45} & {83.49} & \cellcolor{deltabg}{11.96} \\

\bottomrule
\end{tabular}
\end{table}
\begin{table}[t]
\centering
\caption{Robustness under \emph{Authority and Stance Pressure Perturbation} (C2) (\%).}
\label{tab:L4.2}
\small
\renewcommand{\arraystretch}{1.08}
\setlength{\tabcolsep}{2.5pt}
\begin{tabular}{l|ccc|ccc|ccc|ccc}
\toprule
\multicolumn{1}{c|}{\multirow{2}{*}[-0.3ex]{\textbf{Model}}}
& \multicolumn{3}{c|}{\textbf{TACC}} 
& \multicolumn{3}{c|}{\textbf{IC}}
& \multicolumn{3}{c|}{\textbf{TP}}
& \multicolumn{3}{c}{\textbf{RV}} \\
\cmidrule(lr){2-4}
\cmidrule(lr){5-7}
\cmidrule(lr){8-10}
\cmidrule(lr){11-13}

& \textbf{Clean} & \textbf{Pert.} & $\boldsymbol{\Delta}$
& \textbf{Clean} & \textbf{Pert.} & $\boldsymbol{\Delta}$
& \textbf{Clean} & \textbf{Pert.} & $\boldsymbol{\Delta}$
& \textbf{Clean} & \textbf{Pert.} & $\boldsymbol{\Delta}$ \\
\midrule

Gemini 3.1 Pro
& {97.00} & {64.73} & \cellcolor{deltabg}{32.28}
& {98.79} & {81.17} & \cellcolor{deltabg}{17.62}
& {98.91} & {77.84} & \cellcolor{deltabg}{21.07}
& {96.91} & {64.27} & \cellcolor{deltabg}{32.64} \\

Gemini 3.7 Flash
& {92.58} & {23.12} & \cellcolor{deltabg}{69.46}
& {98.88} & {51.26} & \cellcolor{deltabg}{47.63}
& {95.50} & {55.56} & \cellcolor{deltabg}{39.94}
& {96.53} & {33.67} & \cellcolor{deltabg}{62.86} \\

\midrule
DeepSeek V4 Pro
& {94.31} & {38.87} & \cellcolor{deltabg}{55.44}
& {98.38} & {59.95} & \cellcolor{deltabg}{38.43}
& {90.85} & {66.30} & \cellcolor{deltabg}{24.54}
& {97.75} & {46.05} & \cellcolor{deltabg}{51.69} \\

DeepSeek V4 Flash
& {84.91} & {22.15} & \cellcolor{deltabg}{62.77}
& {93.80} & {53.10} & \cellcolor{deltabg}{40.69}
& {93.99} & {59.26} & \cellcolor{deltabg}{34.73}
& {92.83} & {32.38} & \cellcolor{deltabg}{60.45} \\

\midrule
GPT-5.6
& {88.28} & {39.27} & \cellcolor{deltabg}{49.01}
& {95.66} & {64.08} & \cellcolor{deltabg}{31.59}
& {90.54} & {72.16} & \cellcolor{deltabg}{18.38}
& {95.56} & {49.69} & \cellcolor{deltabg}{45.87} \\

GPT-5.5
& {93.06} & {39.53} & \cellcolor{deltabg}{53.53}
& {97.72} & {60.31} & \cellcolor{deltabg}{37.41}
& {97.57} & {68.02} & \cellcolor{deltabg}{29.54}
& {96.08} & {44.09} & \cellcolor{deltabg}{51.99} \\

\midrule
Claude Opus 5
& {82.78} & {32.76} & \cellcolor{deltabg}{50.02}
& {92.12} & {63.38} & \cellcolor{deltabg}{28.74}
& {96.29} & {58.82} & \cellcolor{deltabg}{37.46}
& {88.27} & {37.16} & \cellcolor{deltabg}{51.11} \\

Claude Sonnet 5
& {80.13} & {25.80} & \cellcolor{deltabg}{54.33}
& {91.56} & {56.15} & \cellcolor{deltabg}{35.41}
& {91.72} & {56.74} & \cellcolor{deltabg}{34.98}
& {90.46} & {35.70} & \cellcolor{deltabg}{54.76} \\

\midrule
\textbf{Average}
& {89.13} & {35.78} & \cellcolor{deltabg}{53.35}
& {95.86} & {61.17} & \cellcolor{deltabg}{34.69}
& {94.42} & {64.34} & \cellcolor{deltabg}{30.08}
& {94.30} & {42.88} & \cellcolor{deltabg}{51.42} \\

\bottomrule
\end{tabular}
\end{table}

\section{Detailed Perturbation Position Results}
\label{app:position_results}

This section reports the complete model-specific results for the perturbation position analysis. For perturbations whose insertion positions can be varied, we evaluate valid early (E), middle (M), and late (L) positions in the multi-turn interaction. We report degradation in TACC, IC, TP, and RV, defined as
$\Delta=\mathrm{Clean}-\mathrm{Perturbed}$,
with larger positive values indicating greater degradation. Tables~\ref{tab:position_acc}--\ref{tab:position_val} provide the complete results across all eight models underlying the position analysis in Section~\ref{sec:perturbation_factors}.

\begin{table}[t]
\centering
\caption{Effect of perturbation position on Task Accuracy (\%).
$\Delta\mathrm{TACC}$ denotes the performance degradation at early (E),
middle (M), and late (L) perturbation positions. Larger values indicate greater degradation.}
\label{tab:position_acc}
\renewcommand{\arraystretch}{1.08}
\setlength{\tabcolsep}{2.5pt}
\resizebox{\linewidth}{!}{
\begin{tabular}{l|ccc|ccc|ccc|ccc|ccc|ccc|ccc|ccc}
\toprule
\multicolumn{1}{c|}{\multirow{2}{*}[-0.3ex]{\textbf{ID}}}
& \multicolumn{3}{c|}{\textbf{Gm-Pro}}
& \multicolumn{3}{c|}{\textbf{Gm-Flash}}
& \multicolumn{3}{c|}{\textbf{DS-Pro}}
& \multicolumn{3}{c|}{\textbf{DS-Flash}}
& \multicolumn{3}{c|}{\textbf{GPT-5.6}}
& \multicolumn{3}{c|}{\textbf{GPT-5.5}}
& \multicolumn{3}{c|}{\textbf{Claude Opus-5}}
& \multicolumn{3}{c}{\textbf{Claude Sonnet 5}} \\
\cmidrule(lr){2-4}
\cmidrule(lr){5-7}
\cmidrule(lr){8-10}
\cmidrule(lr){11-13}
\cmidrule(lr){14-16}
\cmidrule(lr){17-19}
\cmidrule(lr){20-22}
\cmidrule(lr){23-25}

& \textbf{E} & \textbf{M} & \textbf{L}
& \textbf{E} & \textbf{M} & \textbf{L}
& \textbf{E} & \textbf{M} & \textbf{L}
& \textbf{E} & \textbf{M} & \textbf{L}
& \textbf{E} & \textbf{M} & \textbf{L}
& \textbf{E} & \textbf{M} & \textbf{L}
& \textbf{E} & \textbf{M} & \textbf{L}
& \textbf{E} & \textbf{M} & \textbf{L} \\

\midrule

U1
& 3.36 & 6.99 & 12.34
& 1.50 & 7.70 & 10.06
& 2.36 & 5.57 & 10.94
& 5.00 & 6.84 & 14.27
& 5.33 & 6.31 & 10.82
& 7.20 & 6.31 & 7.11
& -0.07 & 1.71 & 0.05
& 2.53 & 5.79 & 10.85 \\

U2
& 9.02 & 9.31 & 20.09
& 2.97 & 5.50 & 19.23
& 3.30 & 4.97 & 9.62
& 3.20 & 0.46 & 13.43
& 7.30 & 8.39 & 5.39
& 4.08 & 8.52 & 3.17
& -0.56 & 4.02 & 3.70
& 0.53 & 10.36 & 12.69 \\

U3
& 6.27 & 5.91 & 5.62
& 1.40 & 6.36 & 11.27
& 3.64 & 3.12 & 0.00
& 5.04 & 5.34 & 10.94
& 11.17 & 7.08 & 8.45
& 9.67 & 3.96 & 6.64
& 3.44 & 4.94 & 7.31
& 4.09 & 2.47 & 5.88 \\

U4
& 20.10 & 12.00 & 21.20
& 22.12 & 14.85 & 30.56
& 23.61 & 21.79 & 34.10
& 26.24 & 25.72 & 23.95
& 25.39 & 18.88 & 32.68
& 26.08 & 25.31 & 19.28
& 12.29 & 6.32 & 13.40
& 23.33 & 20.83 & 21.00 \\

\midrule
R1
& 44.75 & 53.77 & 53.19
& 42.94 & 56.75 & 58.88
& 46.21 & 56.67 & 58.55
& 48.32 & 49.94 & 45.63
& 52.09 & 57.67 & 58.20
& 52.16 & 56.38 & 53.41
& 31.78 & 46.91 & 40.82
& 37.19 & 46.02 & 46.23 \\

R2
& 43.23 & 90.23 & 80.77
& 47.47 & 87.95 & 88.46
& 40.09 & 87.27 & 80.77
& 34.98 & 87.58 & 77.35
& 40.15 & 90.30 & 78.21
& 48.97 & 80.76 & 71.79
& 35.56 & 81.82 & 68.16
& 40.01 & 81.67 & 73.93 \\

R3
& 30.61 & 37.26 & 34.57
& 42.98 & 52.58 & 75.19
& 28.12 & 29.68 & 29.86
& 33.20 & 39.79 & 40.05
& 31.93 & 36.47 & 41.23
& 23.16 & 32.85 & 29.32
& 23.47 & 30.89 & 15.99
& 31.12 & 33.97 & 32.35 \\

\bottomrule
\end{tabular}
}
\end{table}

\begin{table}[t]
\centering
\caption{Effect of perturbation position on Information Correctness (\%).
$\Delta\mathrm{IC}$ denotes the performance degradation at early (E), middle (M), and late (L) perturbation positions. Larger values indicate greater degradation.}
\label{tab:position_ic}
\renewcommand{\arraystretch}{1.08}
\setlength{\tabcolsep}{2.5pt}
\resizebox{\linewidth}{!}{
\begin{tabular}{l|ccc|ccc|ccc|ccc|ccc|ccc|ccc|ccc}
\toprule
\multicolumn{1}{c|}{\multirow{2}{*}[-0.3ex]{\textbf{ID}}}
& \multicolumn{3}{c|}{\textbf{Gm-Pro}}
& \multicolumn{3}{c|}{\textbf{Gm-Flash}}
& \multicolumn{3}{c|}{\textbf{DS-Pro}}
& \multicolumn{3}{c|}{\textbf{DS-Flash}}
& \multicolumn{3}{c|}{\textbf{GPT-5.6}}
& \multicolumn{3}{c|}{\textbf{GPT-5.5}}
& \multicolumn{3}{c|}{\textbf{Claude Opus-5}}
& \multicolumn{3}{c}{\textbf{Claude Sonnet 5}} \\
\cmidrule(lr){2-4}
\cmidrule(lr){5-7}
\cmidrule(lr){8-10}
\cmidrule(lr){11-13}
\cmidrule(lr){14-16}
\cmidrule(lr){17-19}
\cmidrule(lr){20-22}
\cmidrule(lr){23-25}

& \textbf{E} & \textbf{M} & \textbf{L}
& \textbf{E} & \textbf{M} & \textbf{L}
& \textbf{E} & \textbf{M} & \textbf{L}
& \textbf{E} & \textbf{M} & \textbf{L}
& \textbf{E} & \textbf{M} & \textbf{L}
& \textbf{E} & \textbf{M} & \textbf{L}
& \textbf{E} & \textbf{M} & \textbf{L}
& \textbf{E} & \textbf{M} & \textbf{L} \\
\midrule

U1
& 2.10 & 1.85 & 1.80
& 1.82 & 2.27 & 5.54
& 0.68 & 1.43 & 2.37
& 2.13 & 1.99 & 2.50
& 1.77 & 1.63 & 3.27
& 3.36 & 2.88 & 5.14
& 0.45 & 0.26 & 1.10
& 1.52 & 0.82 & 2.40 \\

U2
& 4.68 & 3.49 & 3.85
& 2.08 & 2.26 & 6.45
& 0.90 & 1.40 & 3.99
& 1.42 & 0.06 & 1.11
& 2.51 & 0.72 & 1.16
& 2.55 & 2.79 & 3.02
& -0.15 & 0.09 & 0.81
& 1.82 & 2.58 & 2.28 \\

U3
& 3.12 & 3.53 & 2.80
& 1.94 & 2.33 & 2.07
& 1.72 & 1.58 & 1.63
& 2.00 & 2.01 & 2.36
& 3.19 & 2.22 & 2.86
& 4.37 & 1.71 & 3.85
& -0.50 & 0.61 & 3.91
& 1.90 & 2.38 & -1.67 \\

U4
& 6.12 & 4.06 & 4.85
& 5.28 & 5.83 & 6.70
& 5.70 & 7.37 & 3.31
& 5.29 & 9.89 & 8.19
& 4.35 & 4.00 & 0.03
& 7.34 & 8.86 & 9.16
& 3.27 & 1.50 & -0.30
& 5.77 & 4.59 & 8.34 \\

\midrule
R1
& 16.25 & 20.53 & 19.54
& 13.23 & 19.75 & 21.50
& 10.79 & 17.14 & 20.44
& 13.51 & 14.64 & 15.53
& 13.69 & 20.40 & 20.78
& 20.07 & 22.78 & 24.72
& 10.03 & 15.47 & 11.80
& 13.58 & 14.80 & 14.78 \\

R2
& 15.83 & 31.77 & 12.26
& 16.55 & 31.05 & 17.83
& 10.98 & 20.33 & 10.78
& 15.73 & 18.07 & 12.31
& 9.49 & 14.19 & 8.10
& 15.32 & 27.33 & 0.86
& 2.06 & 6.03 & -3.70
& 6.91 & 3.51 & 8.88 \\

R3
& 12.26 & 16.85 & 13.99
& 16.05 & 21.53 & 24.50
& 10.85 & 10.48 & 10.48
& 9.76 & 12.05 & 10.63
& 9.22 & 11.85 & 19.77
& 11.16 & 14.17 & 14.32
& 7.33 & 9.33 & 4.94
& 9.39 & 11.24 & 9.67 \\

\bottomrule
\end{tabular}
}
\end{table}

\begin{table}[t]
\centering
\caption{Effect of perturbation position on Task Progression (\%).
$\Delta\mathrm{TP}$ denotes the performance degradation at early (E),
middle (M), and late (L) perturbation positions. Larger values indicate greater degradation.}
\label{tab:position_rp}
\renewcommand{\arraystretch}{1.08}
\setlength{\tabcolsep}{2.5pt}
\resizebox{\linewidth}{!}{
\begin{tabular}{l|ccc|ccc|ccc|ccc|ccc|ccc|ccc|ccc}
\toprule
\multicolumn{1}{c|}{\multirow{2}{*}[-0.3ex]{\textbf{ID}}}
& \multicolumn{3}{c|}{\textbf{Gm-Pro}}
& \multicolumn{3}{c|}{\textbf{Gm-Flash}}
& \multicolumn{3}{c|}{\textbf{DS-Pro}}
& \multicolumn{3}{c|}{\textbf{DS-Flash}}
& \multicolumn{3}{c|}{\textbf{GPT-5.6}}
& \multicolumn{3}{c|}{\textbf{GPT-5.5}}
& \multicolumn{3}{c|}{\textbf{Claude Opus-5}}
& \multicolumn{3}{c}{\textbf{Claude Sonnet 5}} \\
\cmidrule(lr){2-4}
\cmidrule(lr){5-7}
\cmidrule(lr){8-10}
\cmidrule(lr){11-13}
\cmidrule(lr){14-16}
\cmidrule(lr){17-19}
\cmidrule(lr){20-22}
\cmidrule(lr){23-25}

& \textbf{E} & \textbf{M} & \textbf{L}
& \textbf{E} & \textbf{M} & \textbf{L}
& \textbf{E} & \textbf{M} & \textbf{L}
& \textbf{E} & \textbf{M} & \textbf{L}
& \textbf{E} & \textbf{M} & \textbf{L}
& \textbf{E} & \textbf{M} & \textbf{L}
& \textbf{E} & \textbf{M} & \textbf{L}
& \textbf{E} & \textbf{M} & \textbf{L} \\
\midrule

U1
& 0.69 & 1.08 & 2.75
& 0.95 & 2.08 & 2.28
& -0.57 & 0.33 & 8.56
& 1.63 & 1.58 & 7.03
& 0.40 & 0.98 & 5.43
& 1.44 & 1.50 & 3.96
& 0.35 & 0.62 & 2.63
& 0.36 & 0.57 & 7.28 \\

U2
& 0.46 & -0.67 & 1.63
& 0.42 & -0.04 & 7.79
& -0.14 & 1.86 & 5.13
& 0.80 & -0.38 & 5.50
& 0.54 & -0.47 & 2.94
& 0.95 & 1.65 & -0.44
& 0.21 & -0.00 & 1.63
& -0.53 & -0.05 & 0.27 \\

U3
& 0.05 & 0.58 & -1.60
& 0.16 & 0.39 & 0.52
& -0.99 & -1.59 & 0.84
& 1.11 & 1.39 & 3.08
& 2.66 & 1.30 & -0.69
& 1.47 & 1.25 & 2.15
& -1.01 & 0.12 & 1.70
& -0.26 & 0.60 & -4.02 \\

U4
& 3.75 & 3.25 & 4.21
& 5.85 & 10.14 & 17.09
& 6.12 & 6.24 & 12.50
& 7.69 & 6.62 & 13.31
& 6.07 & 7.93 & 14.62
& 4.01 & 5.79 & 6.78
& 3.54 & 2.75 & 10.54
& 7.11 & 9.43 & 9.07 \\

\midrule

R1
& 3.45 & 5.63 & 11.28
& 7.50 & 13.85 & 11.52
& -0.21 & 6.42 & 10.95
& 3.99 & 8.07 & 9.80
& 7.53 & 9.10 & 12.62
& 4.17 & 4.87 & 12.17
& 1.93 & 4.46 & 11.02
& 5.58 & 6.24 & 9.73 \\

R2
& 28.15 & 68.94 & 64.10
& 28.71 & 69.55 & 61.54
& 26.52 & 62.73 & 65.17
& 31.85 & 72.73 & 73.50
& 30.50 & 77.73 & 68.59
& 35.02 & 75.00 & 68.59
& 31.77 & 73.86 & 58.97
& 37.61 & 81.25 & 73.50 \\

R3
& 2.39 & 5.65 & 5.00
& 7.81 & 13.00 & 18.15
& 3.19 & 4.59 & 4.28
& 5.66 & 8.25 & 8.26
& 3.19 & 5.81 & 8.77
& 2.95 & 3.35 & 5.62
& 2.05 & 4.51 & 5.80
& 6.00 & 3.34 & 4.75 \\

\bottomrule
\end{tabular}
}
\end{table}

\begin{table}[t]
\centering
\caption{Effect of perturbation position on Reasoning Validity (\%).
$\Delta\mathrm{RV}$ denotes the performance degradation at early (E),
middle (M), and late (L) perturbation positions.
Larger values indicate greater degradation.}
\label{tab:position_val}
\renewcommand{\arraystretch}{1.08}
\setlength{\tabcolsep}{2.5pt}
\resizebox{\linewidth}{!}{
\begin{tabular}{l|ccc|ccc|ccc|ccc|ccc|ccc|ccc|ccc}
\toprule
\multicolumn{1}{c|}{\multirow{2}{*}[-0.3ex]{\textbf{ID}}}
& \multicolumn{3}{c|}{\textbf{Gm-Pro}}
& \multicolumn{3}{c|}{\textbf{Gm-Flash}}
& \multicolumn{3}{c|}{\textbf{DS-Pro}}
& \multicolumn{3}{c|}{\textbf{DS-Flash}}
& \multicolumn{3}{c|}{\textbf{GPT-5.6}}
& \multicolumn{3}{c|}{\textbf{GPT-5.5}}
& \multicolumn{3}{c|}{\textbf{Claude Opus-5}}
& \multicolumn{3}{c}{\textbf{Claude Sonnet 5}} \\
\cmidrule(lr){2-4}
\cmidrule(lr){5-7}
\cmidrule(lr){8-10}
\cmidrule(lr){11-13}
\cmidrule(lr){14-16}
\cmidrule(lr){17-19}
\cmidrule(lr){20-22}
\cmidrule(lr){23-25}

& \textbf{E} & \textbf{M} & \textbf{L}
& \textbf{E} & \textbf{M} & \textbf{L}
& \textbf{E} & \textbf{M} & \textbf{L}
& \textbf{E} & \textbf{M} & \textbf{L}
& \textbf{E} & \textbf{M} & \textbf{L}
& \textbf{E} & \textbf{M} & \textbf{L}
& \textbf{E} & \textbf{M} & \textbf{L}
& \textbf{E} & \textbf{M} & \textbf{L} \\
\midrule

U1
& 1.88 & 1.70 & 1.84
& 2.20 & 7.48 & 15.55
& 1.72 & 6.37 & 12.92
& 3.05 & 5.70 & 10.18
& 3.41 & 5.65 & 10.57
& 5.62 & 4.31 & 7.54
& 1.15 & 1.18 & 0.18
& 2.12 & 5.70 & 16.87 \\

U2
& 2.29 & 2.76 & 7.90
& 2.15 & 7.06 & 25.76
& 0.94 & 3.76 & 6.70
& 2.63 & 2.87 & 16.55
& 3.60 & 3.22 & 2.72
& 3.80 & 4.21 & 9.69
& -0.21 & 1.33 & 1.69
& 1.42 & 8.67 & 13.29 \\

U3
& 0.97 & 1.07 & 2.46
& 1.98 & 6.38 & 11.40
& 2.99 & 4.05 & 5.04
& 4.28 & 6.37 & 12.03
& 4.12 & 4.94 & 11.19
& 6.22 & 3.60 & 6.06
& 1.02 & 2.12 & 4.82
& 3.19 & 4.53 & 11.28 \\

U4
& 2.94 & 2.34 & -1.52
& 7.11 & 10.46 & 13.30
& 5.38 & 8.83 & 2.62
& 9.14 & 14.54 & 10.25
& 5.20 & 7.20 & 6.13
& 12.86 & 14.94 & 22.39
& 5.96 & 5.13 & 2.94
& 6.32 & 10.16 & 12.34 \\

\midrule
R1
& 39.15 & 46.13 & 48.85
& 21.57 & 36.11 & 44.97
& 17.84 & 34.89 & 42.49
& 19.11 & 30.84 & 32.20
& 17.45 & 34.83 & 38.41
& 37.10 & 45.11 & 47.59
& 27.09 & 36.60 & 32.84
& 18.02 & 31.22 & 35.59 \\

R2
& 38.70 & 62.95 & 39.74
& 42.33 & 69.39 & 42.31
& 24.75 & 47.50 & 22.44
& 26.51 & 37.95 & 30.56
& 17.69 & 30.45 & 4.70
& 29.10 & 33.56 & 7.26
& 18.68 & 23.48 & 7.91
& 1.04 & -5.00 & 8.12 \\

R3
& 27.05 & 33.08 & 31.17
& 25.19 & 39.94 & 55.49
& 15.93 & 19.14 & 27.55
& 18.68 & 25.78 & 32.64
& 14.85 & 21.02 & 37.10
& 18.39 & 25.68 & 23.09
& 19.27 & 26.42 & 15.37
& 13.99 & 25.21 & 24.32 \\

\bottomrule
\end{tabular}
}
\end{table}

\section{Detailed Perturbation Frequency Results}
\label{app:frequency_results}

This section reports the complete model-specific results for the perturbation frequency analysis. For perturbation types that support repeated instantiation, we evaluate one, two, and three occurrences ($n=1,2,3$), introduced at distinct valid turns within the same multi-turn interaction. We report degradation in TACC, IC, TP, and RV, defined as
$\Delta=\mathrm{Clean}-\mathrm{Perturbed}$,
with larger positive values indicating greater degradation. Tables~\ref{tab:frequency_acc}--\ref{tab:frequency_val} provide the complete results across all eight models underlying the frequency analysis in Section~\ref{sec:perturbation_factors}.

\begin{table}[t]
\centering
\caption{Effect of perturbation frequency on Task Accuracy (\%).
$\Delta\mathrm{TACC}$ denotes the performance degradation under perturbation frequencies of 1, 2, and 3.
Larger values indicate greater degradation.}
\label{tab:frequency_acc}
\renewcommand{\arraystretch}{1.08}
\setlength{\tabcolsep}{2.3pt}
\resizebox{\linewidth}{!}{
\begin{tabular}{l|ccc|ccc|ccc|ccc|ccc|ccc|ccc|ccc}
\toprule
\multicolumn{1}{c|}{\multirow{2}{*}[-0.3ex]{\textbf{ID}}}
& \multicolumn{3}{c|}{\textbf{Gm-Pro}}
& \multicolumn{3}{c|}{\textbf{Gm-Flash}}
& \multicolumn{3}{c|}{\textbf{DS-Pro}}
& \multicolumn{3}{c|}{\textbf{DS-Flash}}
& \multicolumn{3}{c|}{\textbf{GPT-5.6}}
& \multicolumn{3}{c|}{\textbf{GPT-5.5}}
& \multicolumn{3}{c|}{\textbf{Claude Opus-5}}
& \multicolumn{3}{c}{\textbf{Claude Sonnet 5}} \\
\cmidrule(lr){2-4}
\cmidrule(lr){5-7}
\cmidrule(lr){8-10}
\cmidrule(lr){11-13}
\cmidrule(lr){14-16}
\cmidrule(lr){17-19}
\cmidrule(lr){20-22}
\cmidrule(lr){23-25}

& \textbf{1} & \textbf{2} & \textbf{3}
& \textbf{1} & \textbf{2} & \textbf{3}
& \textbf{1} & \textbf{2} & \textbf{3}
& \textbf{1} & \textbf{2} & \textbf{3}
& \textbf{1} & \textbf{2} & \textbf{3}
& \textbf{1} & \textbf{2} & \textbf{3}
& \textbf{1} & \textbf{2} & \textbf{3}
& \textbf{1} & \textbf{2} & \textbf{3} \\
\midrule

U1
& 7.12 & 8.69 & 11.68
& 4.77 & 7.15 & 8.77
& 4.75 & 6.71 & 9.19
& 6.53 & 9.26 & 12.29
& 6.91 & 11.19 & 14.05
& 6.32 & 9.07 & 11.48
& 1.14 & 4.21 & 8.31
& 5.11 & 8.07 & 10.48 \\

U2
& 11.37 & 9.68 & 20.12
& 7.36 & 11.60 & 15.36
& 3.29 & 2.48 & 9.88
& 1.77 & 7.91 & 5.86
& 8.38 & 12.29 & 15.41
& 4.12 & 8.05 & 9.16
& 1.55 & 2.37 & 4.03
& 5.19 & 8.66 & 8.51 \\

U3
& 7.02 & 10.45 & 10.92
& 6.75 & 8.60 & 13.86
& 3.56 & 5.21 & 6.59
& 3.71 & 4.74 & 13.73
& 7.53 & 10.78 & 16.87
& 7.44 & 11.47 & 12.16
& 0.43 & 6.64 & 7.30
& -0.21 & 1.68 & 4.17 \\

U4
& 12.04 & 17.79 & 24.70
& 15.86 & 25.77 & 23.02
& 9.33 & 23.23 & 30.54
& 30.55 & 47.77 & 46.49
& 16.79 & 28.00 & 34.79
& 13.96 & 27.47 & 37.46
& 4.31 & 15.93 & 6.33
& 14.70 & 26.75 & 25.06 \\

\midrule
E1
& 32.18 & 41.38 & 52.30
& 33.56 & 63.30 & 67.70
& 25.93 & 57.37 & 72.06
& 24.83 & 44.15 & 60.38
& 30.96 & 58.05 & 72.51
& 26.82 & 50.64 & 72.17
& 13.43 & 28.07 & 40.46
& 27.13 & 42.26 & 49.75 \\

E2
& 8.05 & 3.16 & 0.69
& 7.76 & 3.45 & 14.02
& 5.11 & 31.07 & 42.44
& 13.57 & 21.32 & 30.58
& 13.85 & 16.78 & 24.71
& 8.00 & 15.11 & 29.17
& 5.75 & 7.78 & -2.84
& 21.26 & 6.90 & 4.31 \\

E3
& 43.76 & 56.27 & 59.65
& 39.24 & 50.98 & 62.61
& 31.74 & 40.93 & 46.22
& 34.56 & 39.28 & 44.64
& 42.83 & 47.26 & 57.44
& 37.70 & 43.35 & 53.10
& 27.75 & 32.77 & 39.74
& 32.18 & 40.28 & 51.55 \\

E4
& 4.60 & 12.64 & 6.03
& 15.80 & 20.31 & 1.09
& 10.18 & 31.51 & 32.67
& 8.57 & 25.30 & 18.15
& 0.92 & 11.72 & 11.67
& 5.96 & 19.98 & 29.56
& 1.57 & 19.54 & 9.89
& 5.91 & 28.49 & 23.93 \\

\midrule
R1
& 58.67 & 71.16 & 73.72
& 61.42 & 71.34 & 77.16
& 64.65 & 76.60 & 81.74
& 53.51 & 76.37 & 75.19
& 65.28 & 76.84 & 79.62
& 64.31 & 81.67 & 79.31
& 50.04 & 68.83 & 71.18
& 55.71 & 69.76 & 71.66 \\

R2
& 86.19 & 95.71 & 92.38
& 92.86 & 94.29 & 95.71
& 91.75 & 93.17 & 93.17
& 88.89 & 90.32 & 88.89
& 86.67 & 86.67 & 86.67
& 88.10 & 89.52 & 96.67
& 86.67 & 86.67 & 86.67
& 80.56 & 80.56 & 80.56 \\

R3
& 53.31 & 86.58 & 86.89
& 45.78 & 78.42 & 76.83
& 46.18 & 75.35 & 78.12
& 42.81 & 61.67 & 67.89
& 43.40 & 73.24 & 69.62
& 42.36 & 87.15 & 91.46
& 41.74 & 70.60 & 43.27
& 42.68 & 62.52 & 65.69 \\

\bottomrule
\end{tabular}
}
\end{table}

\begin{table}[t]
\centering
\caption{Effect of perturbation frequency on Information Correctness (\%).
$\Delta\mathrm{IC}$ denotes the performance degradation under perturbation frequencies of 1, 2, and 3.
Larger values indicate greater degradation.}
\label{tab:frequency_ic}
\renewcommand{\arraystretch}{1.08}
\setlength{\tabcolsep}{2.3pt}
\resizebox{\linewidth}{!}{
\begin{tabular}{l|ccc|ccc|ccc|ccc|ccc|ccc|ccc|ccc}
\toprule
\multicolumn{1}{c|}{\multirow{2}{*}[-0.3ex]{\textbf{ID}}}
& \multicolumn{3}{c|}{\textbf{Gm-Pro}}
& \multicolumn{3}{c|}{\textbf{Gm-Flash}}
& \multicolumn{3}{c|}{\textbf{DS-Pro}}
& \multicolumn{3}{c|}{\textbf{DS-Flash}}
& \multicolumn{3}{c|}{\textbf{GPT-5.6}}
& \multicolumn{3}{c|}{\textbf{GPT-5.5}}
& \multicolumn{3}{c|}{\textbf{Claude Opus-5}}
& \multicolumn{3}{c}{\textbf{Claude Sonnet 5}} \\
\cmidrule(lr){2-4}
\cmidrule(lr){5-7}
\cmidrule(lr){8-10}
\cmidrule(lr){11-13}
\cmidrule(lr){14-16}
\cmidrule(lr){17-19}
\cmidrule(lr){20-22}
\cmidrule(lr){23-25}

& \textbf{1} & \textbf{2} & \textbf{3}
& \textbf{1} & \textbf{2} & \textbf{3}
& \textbf{1} & \textbf{2} & \textbf{3}
& \textbf{1} & \textbf{2} & \textbf{3}
& \textbf{1} & \textbf{2} & \textbf{3}
& \textbf{1} & \textbf{2} & \textbf{3}
& \textbf{1} & \textbf{2} & \textbf{3}
& \textbf{1} & \textbf{2} & \textbf{3} \\
\midrule

U1
& 2.03 & 2.24 & 3.54
& 2.27 & 2.44 & 3.24
& 1.20 & 1.78 & 2.18
& 1.95 & 2.14 & 3.64
& 1.83 & 2.59 & 3.29
& 3.26 & 3.40 & 4.10
& 0.47 & 0.94 & 1.94
& 1.44 & 2.01 & 2.17 \\

U2
& 4.05 & 3.67 & 3.83
& 2.78 & 3.60 & 4.24
& 1.52 & 1.02 & 2.84
& 0.78 & 2.21 & 4.19
& 1.68 & 2.26 & 3.46
& 3.00 & 3.62 & 3.94
& 0.05 & 0.81 & 0.87
& 1.61 & 1.38 & 2.52 \\

U3
& 3.63 & 3.60 & 5.98
& 2.15 & 2.75 & 3.25
& 1.42 & 1.60 & 2.14
& 1.64 & 0.56 & 2.42
& 2.73 & 4.24 & 5.65
& 3.38 & 4.45 & 4.40
& 0.81 & 1.03 & 1.12
& 0.63 & 1.10 & 3.57 \\

U4
& 4.31 & 8.87 & 9.16
& 5.06 & 7.64 & 7.48
& 6.01 & 9.15 & 11.06
& 12.77 & 13.55 & 11.20
& 2.86 & 1.77 & 9.77
& 8.31 & 13.20 & 10.95
& 1.40 & 4.42 & 1.08
& 3.23 & 10.19 & 6.57 \\

\midrule
E1
& 1.13 & 6.73 & 9.86
& 4.06 & 17.71 & 22.28
& 2.06 & 13.56 & 20.55
& 7.77 & 11.76 & 16.10
& 3.71 & 17.73 & 20.72
& 5.11 & 23.01 & 19.23
& 2.16 & 8.10 & 11.02
& 3.29 & 11.04 & 15.11 \\

E2
& 1.03 & 2.68 & 0.27
& 0.17 & 1.73 & 3.02
& 0.25 & 9.97 & 10.78
& 0.53 & 8.25 & 7.24
& 0.14 & 4.08 & 6.03
& 0.70 & 4.30 & 7.18
& 0.48 & 2.01 & 0.32
& 1.82 & 2.82 & 3.87 \\

E3
& 7.79 & 11.27 & 13.36
& 6.59 & 9.60 & 11.74
& 4.51 & 7.81 & 9.74
& 8.16 & 10.03 & 10.79
& 6.18 & 8.73 & 9.15
& 7.17 & 15.27 & 17.22
& 4.32 & 6.24 & 8.60
& 5.01 & 10.04 & 12.65 \\

E4
& 0.41 & 6.88 & 1.29
& 2.38 & 4.76 & 1.51
& 1.88 & 6.66 & 9.34
& 0.35 & 11.19 & 5.87
& -0.49 & 4.53 & 3.23
& 1.20 & 9.20 & 6.81
& 1.54 & 7.61 & 4.09
& 1.34 & 12.62 & 9.20 \\

\midrule
R1
& 18.78 & 31.59 & 35.58
& 18.77 & 31.48 & 35.86
& 19.46 & 33.94 & 37.65
& 19.76 & 34.01 & 35.04
& 22.44 & 32.70 & 36.12
& 25.43 & 45.27 & 48.78
& 15.07 & 31.63 & 35.71
& 18.00 & 32.36 & 35.15 \\

R2
& 30.61 & 35.23 & 43.76
& 23.31 & 31.96 & 43.28
& 18.75 & 25.68 & 27.63
& 15.13 & 22.33 & 28.49
& 11.11 & 13.93 & 12.23
& 20.99 & 23.67 & 32.83
& 1.43 & 3.26 & 4.65
& 2.66 & -3.40 & -2.04 \\

R3
& 19.53 & 31.71 & 38.56
& 18.30 & 31.65 & 33.21
& 18.86 & 25.95 & 35.83
& 9.02 & 24.49 & 31.22
& 16.04 & 33.73 & 38.11
& 16.64 & 42.61 & 41.58
& 12.59 & 28.50 & 5.33
& 13.70 & 24.38 & 20.19 \\

\bottomrule
\end{tabular}
}
\end{table}

\begin{table}[t]
\centering
\caption{Effect of perturbation frequency on Task Progression (\%).
$\Delta\mathrm{TP}$ denotes the performance degradation under perturbation frequencies of 1, 2, and 3.
Larger values indicate greater degradation.}
\label{tab:frequency_rp}
\renewcommand{\arraystretch}{1.08}
\setlength{\tabcolsep}{2.3pt}
\resizebox{\linewidth}{!}{
\begin{tabular}{l|ccc|ccc|ccc|ccc|ccc|ccc|ccc|ccc}
\toprule
\multicolumn{1}{c|}{\multirow{2}{*}[-0.3ex]{\textbf{ID}}}
& \multicolumn{3}{c|}{\textbf{Gm-Pro}}
& \multicolumn{3}{c|}{\textbf{Gm-Flash}}
& \multicolumn{3}{c|}{\textbf{DS-Pro}}
& \multicolumn{3}{c|}{\textbf{DS-Flash}}
& \multicolumn{3}{c|}{\textbf{GPT-5.6}}
& \multicolumn{3}{c|}{\textbf{GPT-5.5}}
& \multicolumn{3}{c|}{\textbf{Claude Opus-5}}
& \multicolumn{3}{c}{\textbf{Claude Sonnet 5}} \\
\cmidrule(lr){2-4}
\cmidrule(lr){5-7}
\cmidrule(lr){8-10}
\cmidrule(lr){11-13}
\cmidrule(lr){14-16}
\cmidrule(lr){17-19}
\cmidrule(lr){20-22}
\cmidrule(lr){23-25}

& \textbf{1} & \textbf{2} & \textbf{3}
& \textbf{1} & \textbf{2} & \textbf{3}
& \textbf{1} & \textbf{2} & \textbf{3}
& \textbf{1} & \textbf{2} & \textbf{3}
& \textbf{1} & \textbf{2} & \textbf{3}
& \textbf{1} & \textbf{2} & \textbf{3}
& \textbf{1} & \textbf{2} & \textbf{3}
& \textbf{1} & \textbf{2} & \textbf{3} \\
\midrule

U1
& 1.27 & 2.09 & 2.27
& 1.37 & 1.55 & 2.34
& 1.11 & 0.93 & 1.39
& 2.16 & 1.73 & 2.88
& 0.88 & 1.54 & 1.85
& 1.84 & 2.39 & 2.85
& 0.73 & 1.17 & 2.01
& 1.71 & 1.65 & 2.36 \\

U2
& 0.09 & -0.66 & -0.08
& 1.54 & 2.10 & 2.27
& 2.27 & 0.81 & 1.41
& 1.50 & 2.52 & 2.79
& -0.03 & 0.91 & -0.40
& 1.00 & 1.34 & 0.79
& -0.03 & 0.98 & 0.15
& -0.42 & -0.16 & 0.06 \\

U3
& 0.04 & 0.35 & -0.52
& 0.34 & 1.63 & 1.24
& -1.46 & 0.44 & 0.54
& 0.81 & 1.39 & 1.71
& 0.87 & 0.77 & 0.75
& 2.61 & 3.14 & 2.99
& -0.38 & 0.27 & 1.64
& -0.53 & -0.70 & -1.34 \\

U4
& 0.75 & 3.08 & 2.31
& 6.97 & 9.66 & 3.69
& -0.51 & 4.52 & -2.75
& 4.12 & 11.36 & 9.14
& 5.32 & 9.87 & 5.85
& 4.07 & 6.94 & 11.20
& 2.02 & 5.94 & -1.57
& 5.34 & 11.95 & 2.10 \\

\midrule
E1
& 0.00 & 1.28 & 0.00
& 0.71 & 6.19 & 8.05
& 1.42 & 4.44 & 3.91
& 1.26 & 7.63 & 9.25
& 0.63 & 4.75 & 8.72
& 0.51 & 7.63 & 12.91
& -0.15 & -0.15 & 2.36
& 0.46 & 1.42 & 8.91 \\

E2
& 0.00 & 0.00 & 0.00
& -0.19 & -0.67 & 0.27
& 0.00 & 5.33 & 5.13
& 1.86 & 1.28 & 4.23
& 0.23 & 2.41 & 0.65
& 0.22 & 0.65 & 7.44
& 0.29 & 0.61 & 0.19
& 1.48 & 1.25 & 0.69 \\

E3
& 0.85 & 2.67 & 2.26
& 2.59 & 3.47 & 4.39
& 2.66 & 4.49 & 2.64
& 1.57 & 3.31 & 5.43
& 3.31 & 4.11 & 5.42
& 0.79 & 5.43 & 7.16
& 1.63 & 2.03 & 2.37
& 2.05 & 3.06 & 2.98 \\

E4
& 0.00 & 4.75 & 0.00
& 2.49 & 4.75 & -0.29
& 1.11 & 4.33 & 1.33
& 0.71 & 3.06 & 3.21
& 0.75 & -0.23 & 0.23
& -1.72 & 3.35 & 1.89
& 0.40 & 9.10 & 2.53
& 1.29 & 5.18 & 8.93 \\

\midrule
R1
& 8.45 & 6.40 & 7.18
& 12.74 & 7.78 & 8.63
& 6.64 & 9.94 & 9.94
& 6.77 & 15.64 & 19.17
& 10.89 & 23.64 & 25.06
& 4.57 & 28.06 & 27.62
& 1.49 & 21.35 & 25.53
& 8.65 & 11.37 & 15.77 \\

R2
& 63.33 & 60.00 & 65.95
& 60.24 & 60.24 & 75.24
& 63.57 & 63.57 & 52.74
& 78.89 & 73.89 & 81.39
& 73.33 & 79.17 & 79.17
& 68.33 & 73.93 & 67.02
& 69.17 & 65.83 & 72.50
& 72.87 & 77.03 & 77.03 \\

R3
& 5.00 & 8.17 & 14.80
& 7.40 & 13.75 & 18.50
& 4.20 & 7.41 & 15.20
& 10.10 & 11.94 & 6.41
& 1.09 & 44.87 & 22.01
& 6.79 & 16.61 & 22.55
& 1.21 & 41.05 & 4.01
& 4.48 & 18.49 & 18.31 \\

\bottomrule
\end{tabular}
}
\end{table}

\begin{table}[t]
\centering
\caption{Effect of perturbation frequency on Reasoning Validity (\%).
$\Delta\mathrm{RV}$ denotes the performance degradation under perturbation frequencies of 1, 2, and 3.
 Larger values indicate greater degradation.}
\label{tab:frequency_val}
\renewcommand{\arraystretch}{1.08}
\setlength{\tabcolsep}{2.3pt}
\resizebox{\linewidth}{!}{
\begin{tabular}{l|ccc|ccc|ccc|ccc|ccc|ccc|ccc|ccc}
\toprule
\multicolumn{1}{c|}{\multirow{2}{*}[-0.3ex]{\textbf{ID}}}
& \multicolumn{3}{c|}{\textbf{Gm-Pro}}
& \multicolumn{3}{c|}{\textbf{Gm-Flash}}
& \multicolumn{3}{c|}{\textbf{DS-Pro}}
& \multicolumn{3}{c|}{\textbf{DS-Flash}}
& \multicolumn{3}{c|}{\textbf{GPT-5.6}}
& \multicolumn{3}{c|}{\textbf{GPT-5.5}}
& \multicolumn{3}{c|}{\textbf{Claude Opus-5}}
& \multicolumn{3}{c}{\textbf{Claude Sonnet 5}} \\
\cmidrule(lr){2-4}
\cmidrule(lr){5-7}
\cmidrule(lr){8-10}
\cmidrule(lr){11-13}
\cmidrule(lr){14-16}
\cmidrule(lr){17-19}
\cmidrule(lr){20-22}
\cmidrule(lr){23-25}

& \textbf{1} & \textbf{2} & \textbf{3}
& \textbf{1} & \textbf{2} & \textbf{3}
& \textbf{1} & \textbf{2} & \textbf{3}
& \textbf{1} & \textbf{2} & \textbf{3}
& \textbf{1} & \textbf{2} & \textbf{3}
& \textbf{1} & \textbf{2} & \textbf{3}
& \textbf{1} & \textbf{2} & \textbf{3}
& \textbf{1} & \textbf{2} & \textbf{3} \\
\midrule

U1
& 2.00 & 2.98 & 3.48
& 5.84 & 5.06 & 5.48
& 5.37 & 4.95 & 5.42
& 4.85 & 4.87 & 7.11
& 5.28 & 6.27 & 7.56
& 5.49 & 6.05 & 7.58
& 1.21 & 3.48 & 5.94
& 5.82 & 5.57 & 6.09 \\

U2
& 3.04 & 3.59 & 4.13
& 7.90 & 7.16 & 7.07
& 2.67 & 1.53 & 2.23
& 3.73 & 4.21 & 4.16
& 3.60 & 5.51 & 4.85
& 4.90 & 6.60 & 7.37
& -0.24 & 0.07 & 0.32
& 4.12 & 4.30 & 3.55 \\

U3
& 1.36 & 2.38 & 1.74
& 5.61 & 5.29 & 5.79
& 3.86 & 3.55 & 3.84
& 4.96 & 4.73 & 6.22
& 4.24 & 5.44 & 7.62
& 5.41 & 6.89 & 9.22
& 0.66 & 1.66 & 3.04
& 4.04 & 3.57 & 4.32 \\

U4
& 1.58 & 6.58 & 20.21
& 8.86 & 13.28 & 18.00
& 5.92 & 14.69 & 23.91
& 17.64 & 14.86 & 30.31
& 6.23 & 9.00 & 22.38
& 13.18 & 17.84 & 26.44
& 9.60 & 15.41 & 16.79
& 5.85 & 12.23 & 21.12 \\

\midrule
E1
& 0.86 & 6.03 & 13.10
& 11.05 & 26.70 & 35.94
& 5.29 & 19.63 & 35.10
& 10.36 & 13.68 & 32.07
& 9.83 & 22.26 & 39.83
& 8.94 & 33.22 & 40.86
& 7.18 & 19.89 & 28.66
& 8.35 & 18.08 & 26.90 \\

E2
& 1.46 & 1.72 & 0.69
& 2.22 & 3.54 & 6.61
& 0.53 & 11.20 & 24.16
& -0.88 & 10.81 & 16.24
& 0.82 & 8.52 & 13.79
& 1.37 & 6.00 & 16.17
& 0.61 & 7.15 & 1.86
& 4.89 & 2.70 & 7.16 \\

E3
& 1.82 & 12.83 & 10.12
& 11.40 & 19.08 & 27.16
& 9.10 & 15.62 & 15.94
& 12.70 & 13.34 & 13.90
& 13.14 & 15.41 & 18.82
& 16.82 & 24.27 & 27.16
& 9.32 & 18.00 & 19.50
& 8.85 & 15.63 & 19.59 \\

E4
& 0.00 & 0.38 & 10.63
& 2.09 & 4.75 & 5.69
& 1.22 & 12.40 & 19.47
& 1.96 & 15.48 & 14.94
& 0.06 & 6.34 & 9.94
& -0.72 & 11.65 & 12.22
& 2.99 & 14.66 & 12.93
& 3.17 & 8.99 & 23.27 \\

\midrule
R1
& 51.36 & 60.94 & 64.41
& 39.65 & 67.03 & 73.00
& 40.86 & 69.96 & 73.71
& 33.25 & 72.13 & 73.04
& 37.59 & 69.90 & 72.11
& 54.11 & 75.77 & 75.29
& 38.76 & 65.89 & 68.72
& 38.13 & 70.04 & 70.11 \\

R2
& 75.12 & 81.31 & 89.64
& 52.62 & 70.71 & 82.14
& 49.01 & 60.67 & 69.84
& 25.56 & 42.82 & 43.89
& 22.78 & 26.94 & 41.94
& 42.54 & 58.97 & 57.30
& 23.33 & 20.83 & 15.83
& 3.13 & -2.08 & 13.54 \\

R3
& 46.22 & 73.82 & 77.95
& 31.44 & 75.31 & 76.43
& 35.00 & 65.99 & 77.03
& 25.95 & 68.76 & 76.90
& 28.66 & 72.40 & 73.22
& 31.87 & 80.02 & 81.44
& 28.68 & 62.26 & 40.85
& 28.70 & 56.99 & 68.47 \\

\bottomrule
\end{tabular}
}
\end{table}

\end{document}